\documentclass[final]{cvpr}

\usepackage{times}
\usepackage{epsfig}
\usepackage{graphicx}
\usepackage{amsmath}
\usepackage{amssymb}
\usepackage{pifont}
\usepackage{rotating}

\usepackage[table]{xcolor}
\usepackage{multirow}

\usepackage{subcaption}
\usepackage[font=small,labelfont=bf]{caption}

\usepackage[ruled,vlined]{algorithm2e}
\usepackage{etoolbox}
\AtBeginEnvironment{algorithm}{\SetArgSty{textrm}}
\SetKwFor{For}{for (}{) $\lbrace$}{$\rbrace$}
\makeatletter
\patchcmd{\@algocf@start}
  {-1.5em}
  {0pt}
  {}{}
\makeatother

\usepackage[export]{adjustbox}

\newcommand\blfootnote[1]{%
  \begingroup
  \renewcommand\thefootnote{}\footnote{#1}%
  \addtocounter{footnote}{-1}%
  \endgroup
}

\definecolor{col1}{RGB}{232, 161, 148}
\definecolor{col2}{RGB}{148, 187, 232}
\definecolor{lightblue}{RGB}{225, 225, 255}
\definecolor{darkgreen}{RGB}{0, 150, 0}
\definecolor{darkred}{RGB}{200, 0, 0}

\definecolor{dsectioncolor}{RGB}{245, 245, 245}
\definecolor{ssectioncolor}{RGB}{175, 238, 255}
\definecolor{hrsectioncolor}{RGB}{255, 228, 196}
\definecolor{hintssectioncolor}{RGB}{255, 155, 155}
\definecolor{mssectioncolor}{RGB}{255, 200, 233}

\newcommand{\cmark}{{\color{darkgreen} \ding{51}}}%

\newcommand{\done}{\textbf{\color{green}DONE}}

\renewcommand{\done}{}

\usepackage[pagebackref=true,breaklinks=true,colorlinks,bookmarks=false]{hyperref}

\def\cvprPaperID{10204} 
\def\confYear{CVPR 2021}

\begin{document}

\title{Single Image Depth Prediction with Wavelet Decomposition}

\author{Micha\"el Ramamonjisoa$^{1,*}$ $\>\>$
Michael Firman$^{2}$
$\>\>$
Jamie Watson$^{2}$\\
Vincent Lepetit$^{1}$
$\>\>$
Daniyar Turmukhambetov$^{2}$
\\
$^1$LIGM, IMAGINE, Ecole des Ponts, Univ Gustave Eiffel, CNRS $\>\>\>$ 
$^2$Niantic\\
\url{www.github.com/nianticlabs/wavelet-monodepth}

}

\maketitle


\begin{abstract}
    We present a novel method for predicting accurate depths from monocular images with high efficiency.
    This optimal efficiency is achieved by exploiting wavelet decomposition, which is integrated in a fully differentiable encoder-decoder architecture. We demonstrate that we can reconstruct high-fidelity depth maps by predicting sparse wavelet coefficients.
    
    In contrast with previous works, we show that wavelet coefficients can be learned without direct supervision on coefficients. Instead we supervise only the final depth image that is reconstructed through the inverse wavelet transform. We additionally show that wavelet coefficients can be learned in fully self-supervised scenarios, without access to ground-truth depth.
    Finally, we apply our method to different state-of-the-art monocular depth estimation models, in each case giving similar or better results compared to the original model, while requiring less than half the multiply-adds in the decoder network.
\end{abstract}

\newcommand{\bB}{\boldsymbol{B}}
\newcommand{\bD}{\boldsymbol{D}}
\newcommand{\bI}{\boldsymbol{I}}
\newcommand{\bN}{\boldsymbol{N}}
\newcommand{\bM}{\boldsymbol{M}}
\newcommand{\bX}{\boldsymbol{X}}
\newcommand{\bn}{\boldsymbol{n}}
\newcommand{\bu}{\boldsymbol{u}}
\newcommand{\bp}{\boldsymbol{p}}
\newcommand{\calL}{\mathcal{L}}

\newcommand{\normals}{\text{n}}
\newcommand{\depth}{\text{d}}
\newcommand{\boundaries}{\text{c}}

\newcommand{\IR}{\mathbb{R}}
\newcommand{\IDWT}{\textbf{\text{IDWT}}}
\newcommand{\DWT}{\textbf{\text{DWT}}}
\newcommand{\lli}[1]{$\mathsf{LL}_{#1}$}
\newcommand{\lhi}[1]{$\mathsf{LH}_{#1}$}
\newcommand{\hli}[1]{$\mathsf{HL}_{#1}$}
\newcommand{\hhi}[1]{$\mathsf{HH}_{#1}$}
\section{Introduction}
\blfootnote{$^*$Work done during an internship at Niantic.}
\begin{figure}[!t]
    \setlength\lineskip{1pt}
    \renewcommand\arraystretch{0.3}
    \setlength\tabcolsep{1pt} 
    \centering
    \begin{subfigure}[b]{0.98\linewidth}
         \centering
        \includegraphics[width=\linewidth]{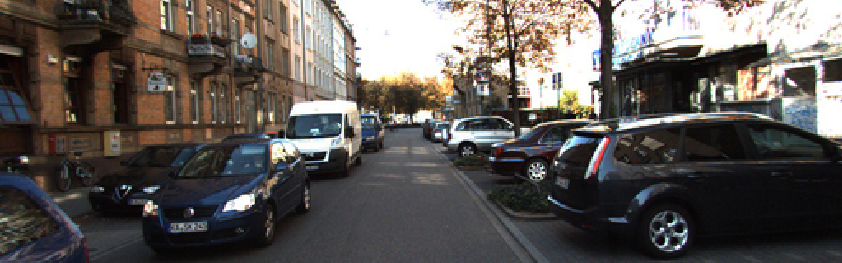}
        \vspace{-2pt}
         \caption{\textbf{Input color image} – (320$\times$1024)}
    \end{subfigure}\\
    \vspace{0.5em}
    \begin{subfigure}[b]{0.98\linewidth}
    \centering
            \begin{tabular}{
                >{\centering\arraybackslash}m{0.5\textwidth}
                >{\centering\arraybackslash}m{0.5\textwidth}
            }
               \includegraphics[width=\linewidth]{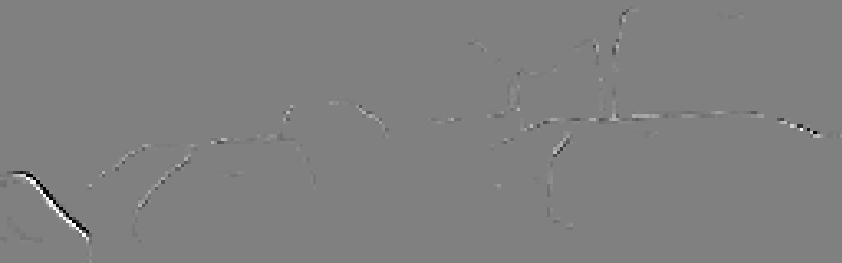} &
               \includegraphics[width=\linewidth]{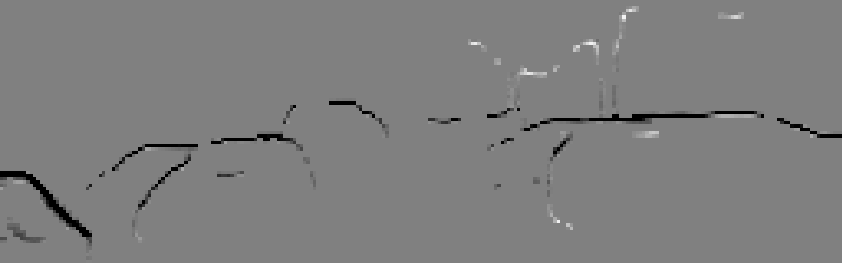}  \\
               \includegraphics[width=\linewidth]{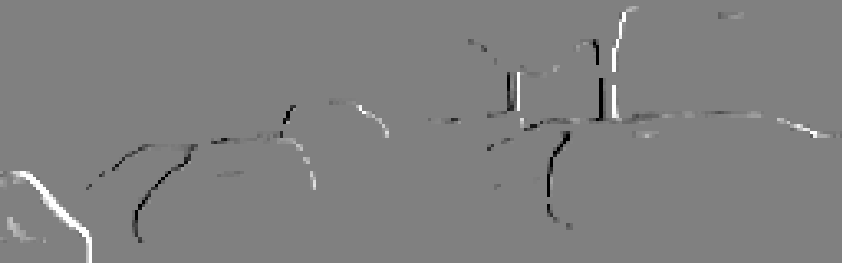} &
               \includegraphics[width=\linewidth]{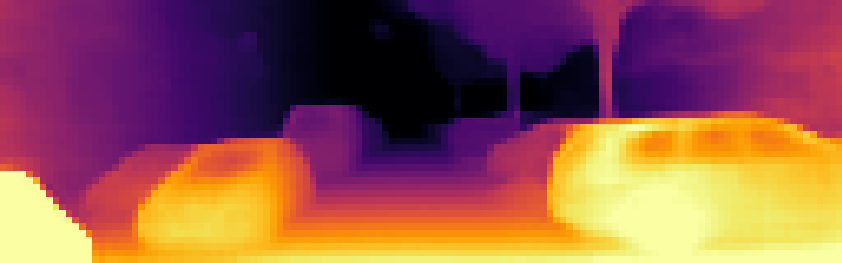} 
            \end{tabular}
            
            \vspace{-3pt}
    
         \caption{\textbf{Sparse estimation using wavelets.} Our network up-samples and refines a 1/16-resolution depth map (bottom-right), by estimating wavelet coefficients only in sparse regions.}
    \end{subfigure}
    \begin{subfigure}[b]{0.98\linewidth}
         \centering
         \vspace{7pt}
        \includegraphics[width=\linewidth]{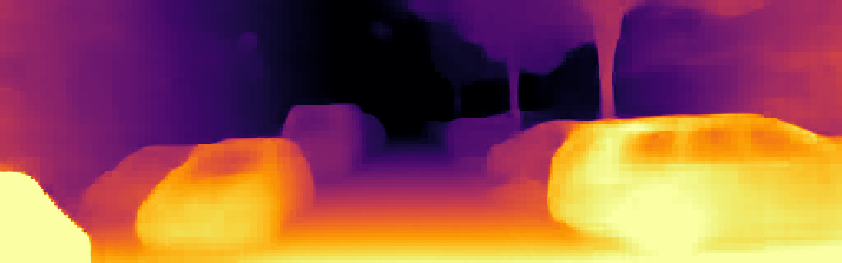}
         \vspace{-12pt}
         \caption{\textbf{Reconstruction of the output depth map using the inverse wavelet transform. }}
    \end{subfigure}
    \footnotesize
    \caption{\textbf{We can represent depth maps more efficiently with wavelets.} Here the network takes image (a) as input and outputs a low resolution depth map, together with sparse wavelet coefficients (b). We can reconstruct a high-resolution depth map (c) using the inverse wavelet transform. In our model we \emph{predict} multi-scale wavelet coefficients with an image-to-image network, and we exploit sparseness of the output to save computation.  \label{fig:figure1}}
    \vspace{-6pt}
\end{figure}

Single-image depth estimation methods are useful in many real-time applications, for example robotics, autonomous driving and augmented reality.
These areas are typically resource-constrained, so efficiency at prediction time is important.

Neural networks which estimate depth from a single image overwhelmingly use U-Net architectures, with skip connections between encoder and decoder layers \cite{Unet}.
Most work on single-image depth prediction has focused on improved depth accuracy, without focusing on efficiency.
Those that have cared about efficiency have typically borrowed tricks from the ``efficient network'' world \cite{howard2017mobilenets, sandler2018mobilenetv2} to make faster depth estimation, with the network using standard convolutions all the way through \cite{WofkFastDepth19, Poggipydnet18}.
All these approaches still use standard neural network components: convolutions, additions, summations and multiplications.

Inspired by sparse representations that can be achieved with wavelet decomposition, we propose an alternative network representation for more efficient depth estimation, using \emph{wavelet decomposition}.
We call this system \textbf{WaveletMonodepth}.
We make the observation that depth images of the man-made world are typically made up of many piece-wise flat regions, with a few `jumps' in depth between the flat regions.
This structure lends itself well to wavelets.
A low-frequency component can represent the overall scene structure, while the `jumps' can be well captured in high-frequency components.
Crucially, the high-frequency components are sparse, which means computation can be focused only in certain areas.
This has the effect of saving run-time computation, while still enabling high-quality depths to be estimated.

To the best of our knowledge, we are the first to train a single-image depth estimation network that reconstructs depth by predicting wavelet coefficients. Furthermore, we show that our models can be trained with self-supervised loss on the final depth signal, in contrast to other methods that directly supervise predicted wavelet coefficients.

We evaluate on NYU and KITTI datasets, where we train supervised and self-supervised, respectively.
We show that our approach allows us to effectively trade off depth accuracy against runtime computation.

\section{Related work}
We first give an overview of monocular depth estimation, before looking at works which have made depth estimation more efficient.
We  then discuss related works which have used wavelets for computer vision tasks, before finally looking at other forms of efficient neural networks.

\subsection{Monocular Depth Estimation}

Beyond early shape-from-shading methods, most works that estimate depth from a single image have been learning-based.
Early works used a Markov random field \cite{SaxenaMake3D}, but more recent works have used deep neural networks.
Supervised approaches use image-to-image networks to regress depth maps \eg~\cite{Eigen14, Eigen2015PredictingDS,kumar2018depthnet,fu2018deep}; however these require ground truth depth data at training time.
Self-supervision reduces the requirement of supervised data by using stereo frames \cite{garg2016unsupervised,godard2017unsupervised} or nearby video frames \cite{zhou2017unsupervised} as supervision, exploiting 3D geometry with image reconstruction losses to learn a depth estimator.
Focus in this area is typically around improving the depth accuracy scores, \eg by modelling moving objects at training time \cite{bian2019unsupervised,chen2019self,yin2018geonet,gordon2019depth, ranjan2018adversarial} or by modelling occlusion \cite{gordon2019depth, godard2019digging}.
While these improvements achieve higher scores with equivalently trained architectures, some works aim for improved depth accuracy \emph{at the expense of efficiency}.
For example, by using higher resolution images \cite{luo2019every}, larger networks \cite{guizilini20203d} or classification instead of regression at the output layer \cite{fu2018deep}.

\vspace{-12pt}

\paragraph{Efficient depth estimation.}
A relatively small number of works focus on efficiency specifically for depth.
Poggi \textit{et al.}~\cite{Poggipydnet18} introduce PyDNet, which uses an image pyramid to enable a high receptive field with a small number of parameters.
Wofk \textit{et al.}~\cite{WofkFastDepth19} introduce FastDepth, which uses depthwise separable layers and network pruning to achieve efficient depth estimation.
An alternative angle on efficient depth estimation is to focus on the training procedure.
Several works use knowledge distillation to enable a small depth estimation network to learn some of the knowledge from a larger network \eg~\cite{CreamIROS2018,StructKDPAMI20}.

In contrast to these works, our contribution is to change the internal representation of depth within the network itself. 
We note that our contributions could be used in conjunction with the above efficient architectures or distillation schemes.

\subsection{Wavelets in Computer Vision}

Wavelet decomposition is an extensively used technique in signal processing, image processing and computer vision. The discrete wavelet transform (DWT) allows a representation of a discrete signal which is more redundant and hence compressible. A notable example is compression of images with JPEG2000 format~\cite{UnserJPEG2000,JPEG2000Book}. Furthermore, wavelet decomposition is also a frequency transform, and can be used for denoising~\cite{DonohoSoftThreshold,DonohoWaveletShrinkage,Kang2018FrameletDenoising}.
Wavelet transforms have also recently been combined with Deep Learning to restore images affected by Moiré color artifacts, which occur when RGB sensors are unable to resolve high-frequency details~\cite{Luo2020CVPRW, liu2020waveletbased}. Li~\etal~\cite{Li2020CVPR} show that by substituting pooling operations in neural networks with discrete wavelet transforms it is possible to filter out high-frequency components of the input image during prediction and thus improve noise-robustness in image classification tasks. Super-resolution methods~\cite{Guo2017DWSR, huang2017wavelet, Deng2019ICCV} learn to estimate the high-frequency wavelet coefficients of an input low-resolution image to generate high-frequency image through inverse wavelet transform.

Closer to our work, Yang~\etal~\cite{Yang2020CVPR} use wavelets in a stereo matching network but require supervision of wavelet coefficients while we do not. Similarly, Luo~\etal~\cite{Luo2020CVPR} replaced the down-sampling and up-sampling operations of UNet-like architectures with DWTs and inverse DWT respectively, and replaced standard skip-connection with high-frequency coefficient skip-connections.
However, they do not directly predict wavelet coefficients of depth and as such are unable to exploit the sparse representation of wavelets for efficiency.
In contrast with both these works, we focus on efficient depth prediction \emph{from a single image}.

\subsection{Efficient Neural Networks}

Convolutional Neural Networks (CNNs)~\cite{Lecun1995convolutional} have revolutionized the field of computer vision as CNN based methods tend to outperform every other competing methods on regression or classification tasks, if they are provided enough training data. 
However, the best performing neural networks contain a large number of parameters and require a large number of floating point operations (FLOPs) at runtime, making deployment to lightweight platforms problematic. 
Many architectures have been developed to improve the accuracy/speed tradeoff in deep nets.
For example, depth-wise separable convolutions \cite{howard2017mobilenets}, inverted residual layers \cite{sandler2018mobilenetv2}, and pointwise group convolutions \cite{zhang2018shufflenet}.
An alternative approach though is to train a network before cutting down some of its unnecessary computations.

\vspace{-8pt}

\paragraph{Channel pruning.}
One line of research is \textit{network pruning}~\cite{Liu2017ICCV, He2017ICCV, yu2019slimmable}, which consists of removing some of the redundant filters in a trained neural network. 
While this helps reducing the network memory footprint as well as the number of FLOPs necessary for inference, sparsity is typically enforced through regularization terms~\cite{Wei16NIPS, he2017channel} to compress the network without losing performance. 
Using such regularisation, however, often requires careful tuning to achieve the desired result~\cite{ye2018rethinking}.
In contrast, our wavelet-based method intrinsically provides sparsity in outputs and intermediate activations, and the wavelet predictions coincide with \emph{edges} in the depth map, knowledge of which has direct applications \eg in augmented reality \cite{SharpNet2019, holynski2018fast}.

While most works focus on classification, channel pruning has also been successfully applied to depth estimation in the aforementioned FastDepth~\cite{WofkFastDepth19}, which uses Net\-Adapt~\cite{yang2018netadapt} to perform channel pruning. 

\vspace{-8pt}

\paragraph{Sparse inference.}
Another recent work considers spatially sparse inference in image-to-image translation tasks. PointRend~\cite{Kirillov2020CVPR} treats semantic segmentation as a rendering process, where a high-resolution estimate is obtained from a low-resolution one through a cascade of upsampling and sparse refinement operations. The location of these sparse \textit{rendering} operations is chosen based on an uncertainty measure of the classification method. However, while they demonstrate the efficiency and applicability of their method to classification tasks, their method cannot directly be applied to regression tasks because of the requirement to evaluate an uncertainty heuristic for all pixel locations. In contrast, our method can directly be applied to regression tasks, as \textit{rendering} locations are directly predicted by our model as non-zero-valued high-frequency wavelet coefficients.

\section{Method}
In this section, we first introduce the basics of 2D wavelet transforms. We chose Haar wavelets~\cite{haar1910theorie} due to their simplicity and provided efficiency. Next, we describe how to use the cascade nature of wavelet representations to build our efficient depth estimation architecture, which we call \textbf{WaveletMonodepth}. Finally, we discuss the computational benefits of sparse representations.

\subsection{Haar Wavelet Transform}

The Haar wavelet basis is the simplest basis of functions for wavelet decomposition.
A discrete wavelet transform (DWT) with Haar wavelets decomposes a 2D image into four coefficient maps: a low-frequency~(\textsf{L}) component \lli{} and three high-frequency~(\textsf{H}) components \lhi{}, \hli{}, \hhi{} at half the resolution of the input image. For the remainder of the paper, we refer to the coefficient maps as the output of the $\DWT$.
The $\DWT$ is an invertible operation,
where its inverse, $\IDWT$, converts four coefficient maps into a 2D signal at twice the resolution of the coefficient maps.

\begin{figure} 
	\centering
	\begin{subfigure}[b]{\linewidth}
		\centering
		\includegraphics[width=\linewidth]{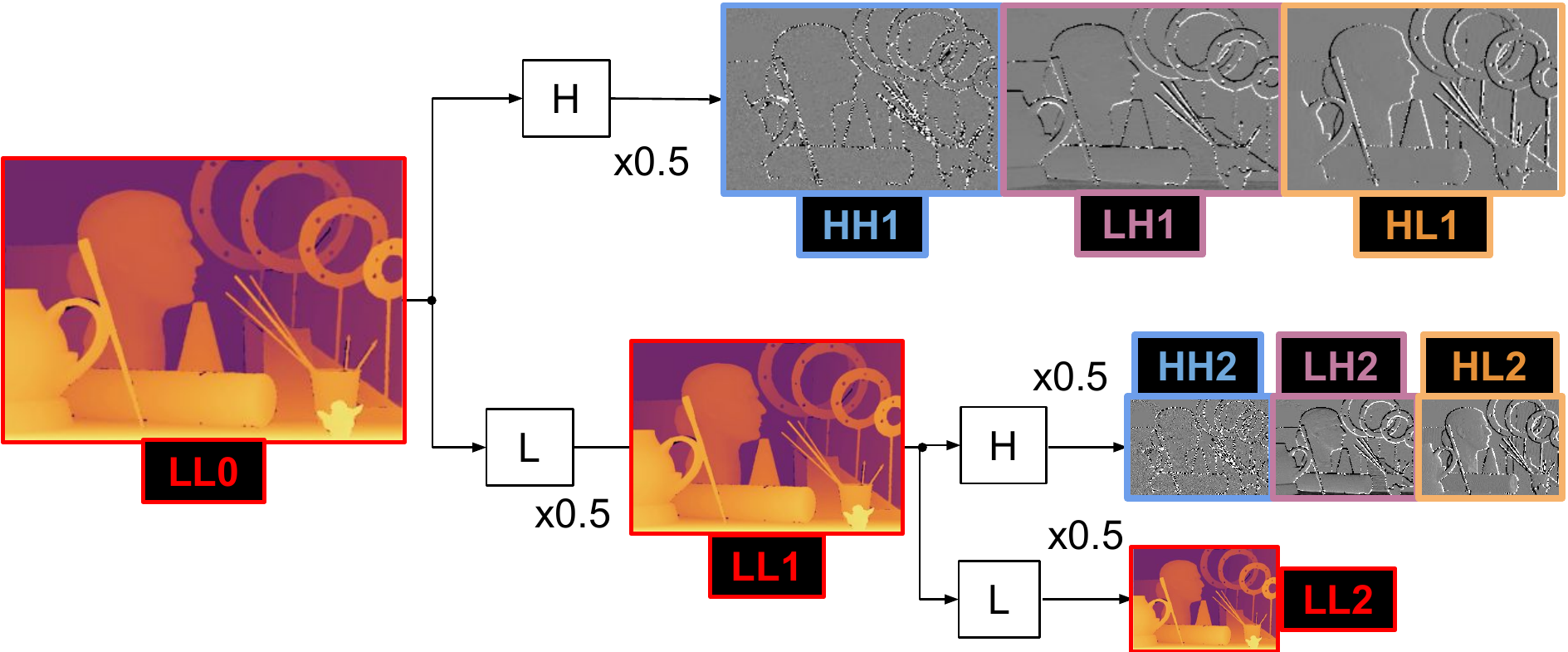}
		\caption{\textbf{Forward transform (DWT).} This decomposes the high resolution input (left) into multi-scale low resolution outputs.}
		\label{fig:background_analysis}
	\end{subfigure}\\
	\vspace{0.5em}
	\begin{subfigure}[b]{\linewidth}
		\centering
		\includegraphics[width=\linewidth]{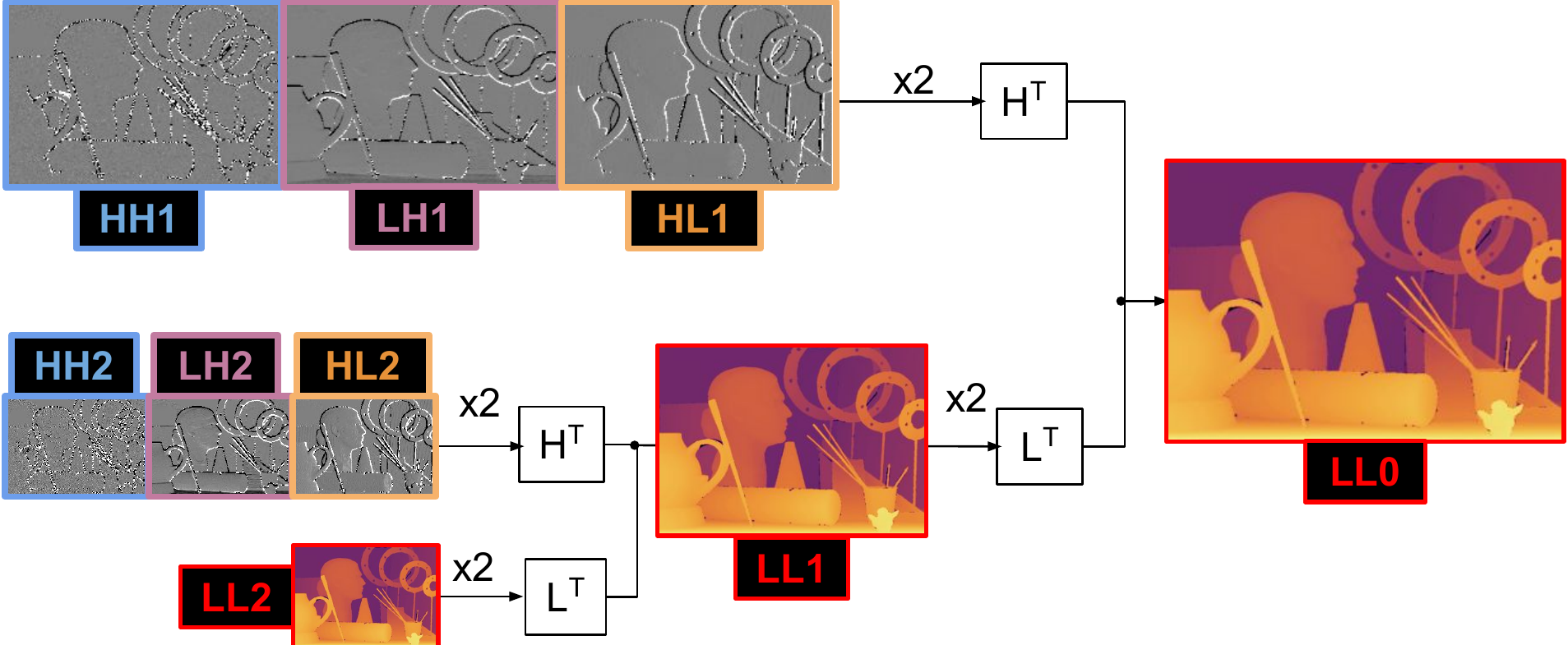}
		\caption{\textbf{Inverse transform (IDWT).} This reconstructs the original input from estimated low resolution inputs.}
		\label{fig:background_synthesis}
	\end{subfigure}
	\caption{Illustration of a two-level wavelet representation of a depth image. The input image $\mathsf{LL_0}$ is passed through a two-level wavelet decomposition (a), to produce a low frequency depth map together with associated wavelets for high frequency detail. The inverse wavelet transform (b) can reconstruct the original image from the wavelet decomposition.}
	\vspace{-10pt}
	\label{fig:filter_banks}
\end{figure}

The multi-scale and multi-frequency wavelet representation is build by recursively applying $\DWT$ to the low-frequency coefficient map \lli{}, starting from the input image---see Figure~\ref{fig:filter_banks}(a). Similarly, the multi-scale representation can be recursively inverted to reconstruct a full resolution image~(Figure~\ref{fig:filter_banks}(b)). This synthesis operation is the building block of our depth reconstruction method.

\label{ssec:filter_banks}

\begin{figure*} 
	\begin{center}
		\includegraphics[width=0.97\textwidth, height=9cm]{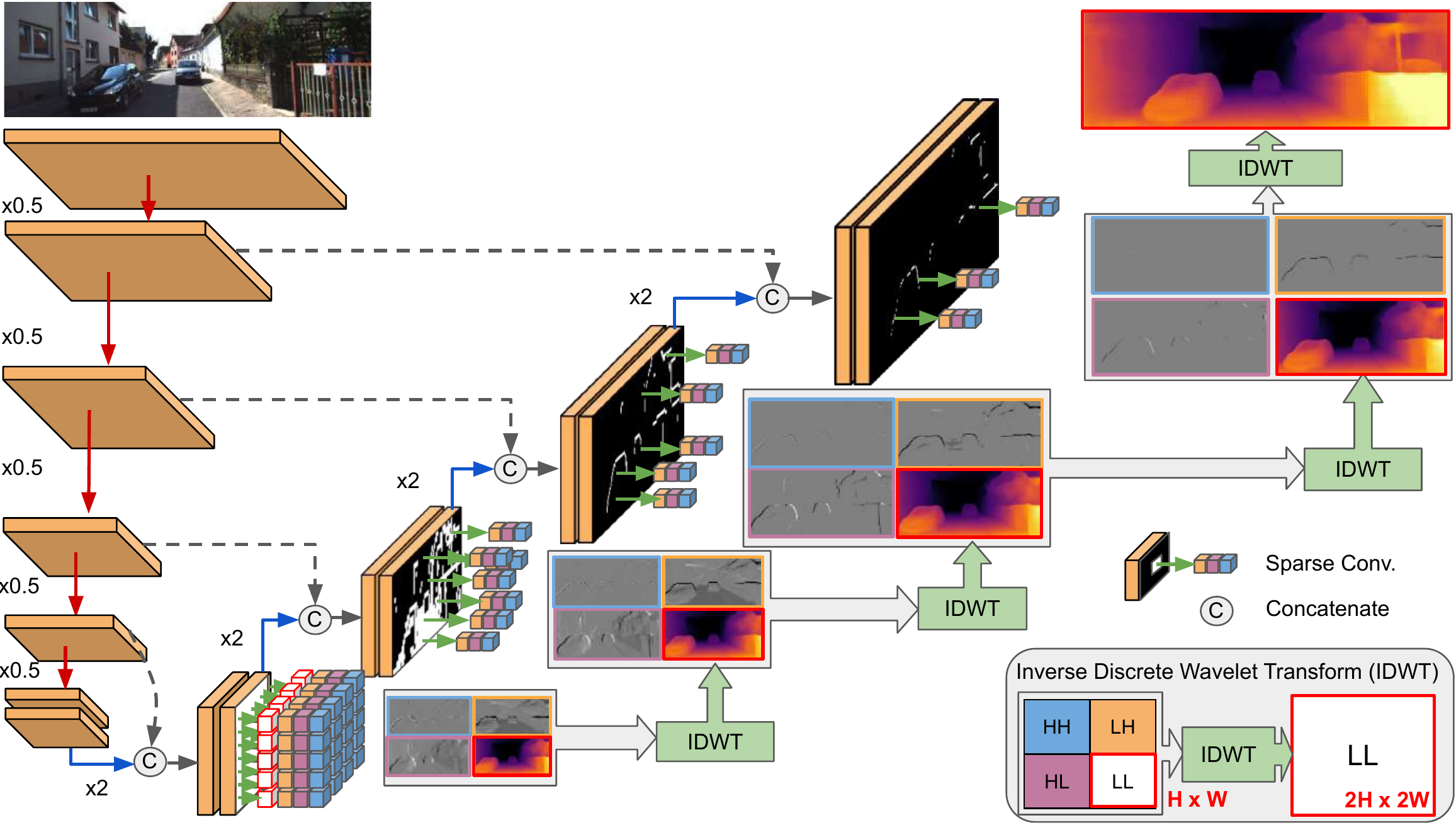}
	\end{center}
	\vspace{-4pt}
	\caption{\textbf{Our method WaveletMonoDepth predicts depth from a single image using wavelets.} 
		At each stage in our decoder, we predict sparse wavelet coefficients $\{\mathsf{LH}, \mathsf{HL}, \mathsf{HH}\}$. 
		These capture the high-frequency details of the depth map, \eg occlusion boundaries. These are combined with the low-frequency depth map $\mathsf{LL}$, taken from the previous level in the decoder, and passed through an inverse discrete wavelet transform~(IDWT). This generates a new depth map at twice the resolution of $\mathsf{LL}$. This process is continued through the decoder until the original input image resolution is reached. Because the wavelet coefficients are \emph{sparse}, we can save computations; we need only to evaluate each decoder layer at the \emph{non-zero} wavelet locations in the previous level. See Algorithm~\ref{algo:decoder} for more details.
		\label{fig:architecture}}
	\vspace{-8pt}
\end{figure*}

\subsection{WaveletMonoDepth}
\label{sec:waveletmonodepth}
Our method, which we call WaveletMonoDepth, is summarized in Figure~\ref{fig:architecture}. 
It builds on a recursive use of $\IDWT$ operation applied to predicted coefficient maps.
Thus, we reconstruct a depth map at the input scale by first predicting a coarse estimate at the bottleneck scale of a UNet-like architecture~\cite{Unet}, and iteratively upscale and refine this estimate by predicting high-frequency coefficient maps.

In our network architecture, the coarse depth estimate $\mathsf{LL_3}$ is estimated at 1/16 of the input scale. This depth map is then progressively upscaled and refined using Algorithm~\ref{algo:decoder}. A forward pass of our model generates a collection of 5 depth maps $\mathsf{LL_s}$ for scales [1/16, 1/8, 1/4, 1/2, 1]. We choose to supervise only the four last scales as in \cite{godard2019digging}. It is worth noting that the coefficient maps are predicted at scales [1/16, 1/8, 1/4, 1/2], thus removing the need for full-resolution computation.

\begin{algorithm}[!ht]
	\SetKwInOut{Input}{Input}
	\SetKwInOut{Init}{Initialize}
	\SetAlgoLined
	\SetKwFunction{densepredict}{\text{DensePredict}}\SetKwFunction{sparsepredict}{\text{SparsePredict}}
	\SetKwFunction{getsparsemask}{\text{GetSparseMask}}
	\SetKwProg{myproc}{procedure}{}{}
	\KwResult{Depth map at input scale}
	\Input{Pyramid of feature maps [F$_4$, F$_3$, F$_2$, F$_1$]\;}
	Current scale $s = 3$\; 
	\lli{3} $\gets$ \densepredict(F$_4$)\;
	Threshold $\eta$ \;
	Sparse computation mask $M =$ Initialize with 1\;
	\For{$s=3;\ s>=0;\ s=s-1$}{ $\gets$ 
		\lhi{s}, \hli{s}, \hhi{s} $\gets$ \sparsepredict(F$_{s+1}$; M)\;
		\lli{s-1} $\gets$ \IDWT(\lli{s}, [\lhi{s}, \hli{s}, \hhi{s}])\;
		$\eta_s$ $\gets$ $\eta \cdot (\max(\text{\lli{s-1}})-\min(\text{\lli{s-1}}))$\;
		$M$ $\gets$ \getsparsemask(\lhi{s}, \hli{s}, \hhi{s}, $\eta_s$)\;
	}
	\myproc{\sparsepredict{F, M}}{
		\Input{Feature map $F$, Sparse mask $M$}
		\For{\text{all} $p$ s.t. $M[p]==1$}{
			H[p] = SparseConv3x3(F[p])\;
		}
		\KwRet H\;}{}
	
	\myproc{\densepredict{F}}{
		\Input{Feature map F}
		Initialize $M$ with ones\;
		\KwRet \sparsepredict{F, M}\;}{}
	
	\myproc{\getsparsemask{H, $\eta$}}{
		\Input{High frequency coefficient maps $H$}
		$M = \max(|\mathsf{LH}|, |\mathsf{HL}|, |\mathsf{HH}|) > \eta$ \;
		$M = \texttt{upsample}_{\times 2}(M)$ \;
		\KwRet $M$\;}{}
	\caption{Computing depth with wavelets}
	\label{algo:decoder}
\end{algorithm}

\subsection{Sparse Computations in Decoder}\label{sec:sparse}

For piecewise flat depth maps, high-frequency coefficient maps have a small number of non-zero values; these are located around depth edges. Hence, for full-resolution depth reconstruction, only some pixel locations need to predict non-zero coefficient map values at each scale. At any scale, we assume that these pixel locations with non-zero values can be determined from high-frequency coefficient maps estimated at the previous scale defined by a mask $M$ described in $\texttt{GetSparseMask}$ of Algorithm~\ref{algo:decoder}.

The sparsity level achieved by using mask $M$ is
\begin{equation}
	\psi = \frac{\sum_{r,c=1,1}^{H,W}M_{r,c}}{HW},
\end{equation}
which allows us to remove redundant computation in the decoder layer.
Indeed, for a typical $K\times K$ convolution (with a bias term) on a feature tensor of size $H \times W$ that has $C_\text{in}$ input channels and $C_\text{out}$ output channels, the number of multiply-add operations is
\begin{equation}
	\text{MAC}_{\text{dense}} = HW(C_\text{in}K^2 + 1)C_\text{out} .
\end{equation}
With the sparsity level $\psi$, it would be
\begin{equation}
	\text{MAC}_{\text{sparse}} = \psi HW(C_\text{in}K^2 + 1)C_\text{out}.
	\label{eq:mac_sparse}
\end{equation}

Note that our sparsification strategy aims to reduce FLOPs by decreasing the number of pixel locations at which we need to compute an output. This approach is orthogonal and complements other approaches such as channel pruning, which instead reduces $C_\text{in}$ and $C_\text{out}$, or separable convolutions. We refer to supplementary material for further details on these. 

Considering a quite conservative threshold $\eta=0.05$ used on high-frequency coefficient maps, the sparse decoder computation is about 3$\times$ lower in FLOPs compared to standard convolutions at all pixel locations for an image of size $320\times 1024$.

\subsection{Self-supervised Training}
Our self-supervised losses are as described in \cite{godard2019digging}, which we briefly describe here for completeness. See supplemental material for further details.
Given a stereo pair of images ($I_L, I_R$), we train our network to predict a depth map $D_L$, pixel-aligned with the left image.
We also assume access to  the camera intrinsics $K$, and the relative camera transformation between the images in the stereo pair $T_{R \to L}$.
We use the network's current estimate of depth to synthesise an image $I_{R \to L}$, computed as
\begin{align}
	I_{R \to L} &= I_{R}\Big\langle \text{proj}(D_L, T_{R \to L}, K) \Big\rangle,
\end{align}
where $\text{proj}()$ are the 2D pixel coordinates obtained by projecting the depths $D_L$ into image $I_R$, and $\big\langle\big\rangle$ is the sampling operator.
We follow standard practice in training  with a photometric reconstruction error $pe$, so our loss becomes $L_p = pe(I_L, I_{R \to L})$.
Following \cite{godard2019digging, chen2019self} etc., we set $pe$ to a weighted sum of SSIM and $L_1$ losses.

We also include the depth smoothness loss from \cite{godard2019digging}.

For our experiments which train on monocular and stereo sequences (`MS'), we combine reprojection errors from the three different source images: one frame forward in time, one frame back in time, and the corresponding stereo pair. 
In this case, we create synthesized images from the monocular sequence using relative poses estimated from a pose network, as described in \cite{godard2019digging}.
In this setting, we use a per-pixel minimum reprojection loss, again following \cite{godard2019digging}.

\section{Experiments}
Our validation experiments explore the task of training a CNN to predict depth from
a single color image, using wavelets as an intermediate representation.
Depending on the experiment, we compare against known leading baselines that supplement,
and pre- and post-process the stereo pairs used for supervision, and the output depth maps.

\subsection{Implementation Details}

\paragraph{Datasets}
We conduct experiments on the KITTI and NYUv2 depth datasets.
KITTI \cite{Geiger2012CVPR} consists of 22,600 calibrated stereo video pairs captured by a car driving around a city in Germany. Models are evaluated using the Eigen split \cite{Eigen2015PredictingDS} using corresponding LiDAR point clouds; see \eg~\cite{godard2017unsupervised} for details.
NYUv2 \cite{silberman2012indoor} consists of RGBD frames captured with a Kinect sensor. There are 120K raw frames collected by scanning various indoor scenes. 
As in DenseDepth~\cite{Densedepth}, we use a 50K samples subset of the full dataset where depth is inpainted using Levin~\textit{et al.} inpainting method~\cite{Levin2004colorization}. 
The NYUv2 evaluation is run on the 654 test frames introduced by Eigen~\etal~\cite{Eigen14}.

\vspace{-9pt}
\paragraph{Metrics} 
On the KITTI dataset we compute depth estimation scores based on the standard metrics introduced by Eigen \textit{et al.}~\cite{Eigen14}: Abs Rel, Sq Rel, RMSE, RMSE$_{log}$, $\delta_1=\delta < 1.25$, $\delta_2=\delta < 1.25^2$, and $\delta_3=\delta < 1.25^3$.
We use the same metrics for NYUv2, but we follow standard practice (\eg~\cite{fu2018deep}) in reporting $log_{10}$ instead of RMSE$_{log}$. To evaluate the sharpness of depth maps on NYUv2, we use the metrics introduced by Koch \textit{et al.}~\cite{Koch2018EvaluationOC, koch2020comparison} and the NYU-OC++ dataset manually annotated by Ramamonjisoa \textit{et al.}~\cite{SharpNet2019, ramamonjisoa2020displacement}.

\vspace{-9pt}
\paragraph{Models} To demonstrate the efficiency of our method, we choose two models for experiments on the NYUv2 and KITTI datasets. Both models are implemented using Pytorch and use a compatible implementation of \IDWT~\cite{CotterPytorchWavelets}.

For KITTI, we choose the weakly-supervised Depth Hints~\cite{watson-2019-depth-hints} method, which adds Semi Global Matching~\cite{hirschmuller2005accurate, hirschmuller2007stereo} supervision to the self-supervised Monodepth2 \cite{godard2019digging}, without requiring Lidar depth supervision. At each scale $s$ of the Monodepth2 decoder there is a  layer which outputs a one-channel disparity.
We replace this layer at each scale with a 3-channel output layer to predict \{\lhi{s}, \hli{s}, \hhi{s}\}. While our baseline consumes decoder feature maps at scales [1/16, 1/8, 1/4, 1/2, 1], we only need to keep the four scales [1/16, 1/8, 1/4, 1/2], as the $\IDWT$ outputs disparity at $2\times$ resolution. Both our model and baseline are trained with an Adam optimizer using a learning rate of $10^{-4}$, with batch size 12 for 20 epochs. Unless otherwise specified, our experiments are done with Resnet50-based model trained with depth hints loss at $320 \times 1024$ resolution.

For NYUv2, we implement a UNet-like baseline similar to DenseDepth~\cite{Densedepth}, and detail its architecture in supplementary material.
Similar to our KITTI experiments, we discard the last layer of the decoder as it is not needed, and add one extra layer at each scale to predict the wavelet coefficients. Both our model and baseline are trained using an Adam optimizer with standard parameters, for 20 epochs with batch size 8 and with  learning rate $10^{-4}$. It is worth noting that DenseDepth predicts outputs at half the input resolution, but evaluates at full resolution after bilinearly upsampling.

\subsection{Efficiency vs.~Accuracy Trade-off Analysis}

In this section, we study the relation between accuracy, sparsity, and efficiency of WaveletMonoDepth. For each set of experiments, we compare our method to an equivalently trained model without wavelets. 
We first study how wavelets contribute to high-frequency details, then show that they are sparse. Finally, we discuss how we trade off accuracy against efficiency by varying the threshold $\eta$ used in Algorithm~\ref{algo:decoder} to filter out close-to-zero coefficients.

\begin{figure} 
	\centering
	\includegraphics[width=0.9\linewidth]{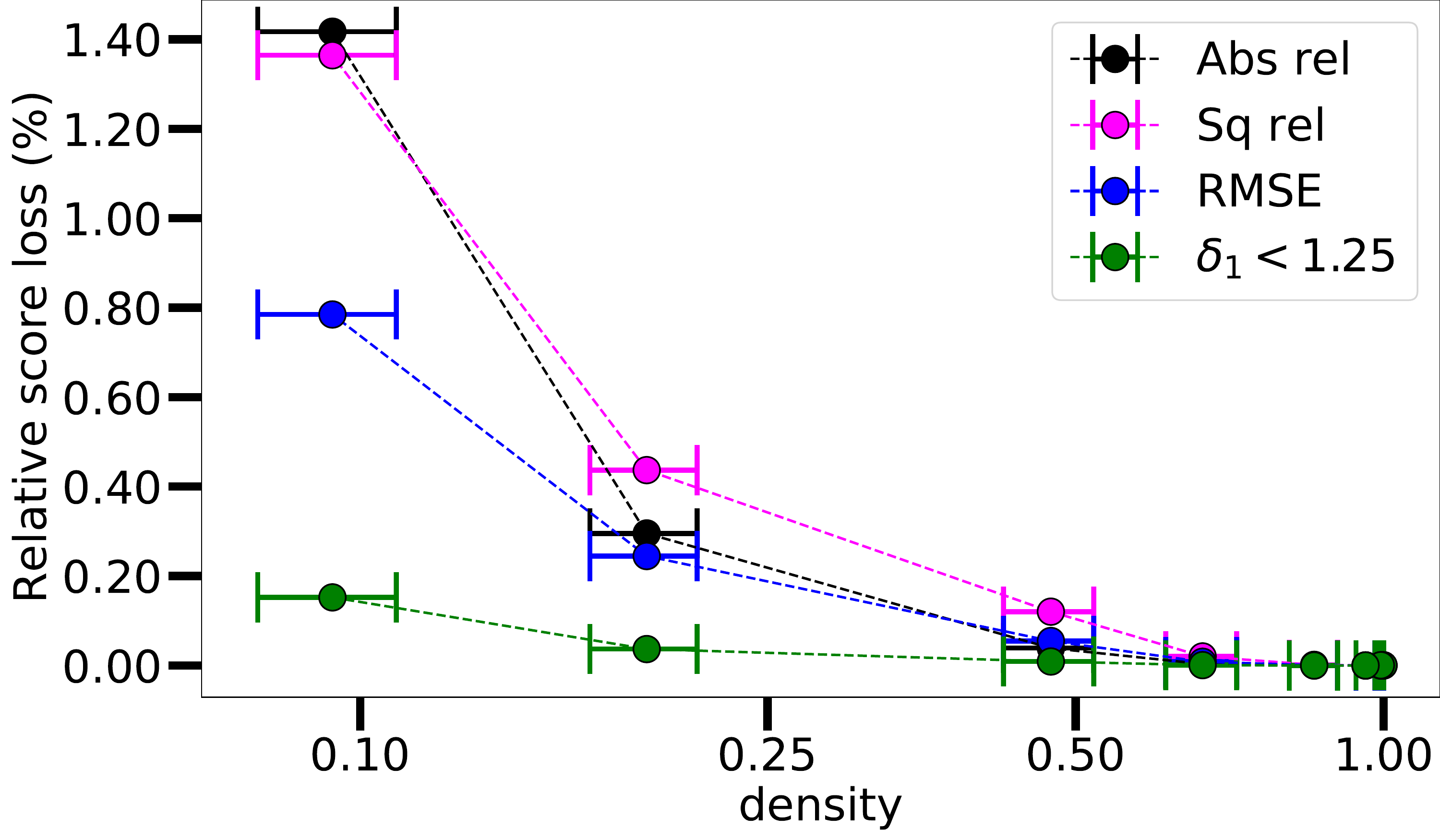}
	\vspace{-8pt}
	\caption{\textbf{Analysis of performance loss vs density on KITTI}.  Using our wavelet representation, we can drop up to 90\% of the wavelet coefficients while suffering a maximum relative performance loss of less than 1.4\%.}
	\vspace{-9pt}
	\label{fig:relative_scores_vs_sparsity_kitti}
\end{figure}

\begin{figure} 
	\centering
	\includegraphics[width=0.9\linewidth]{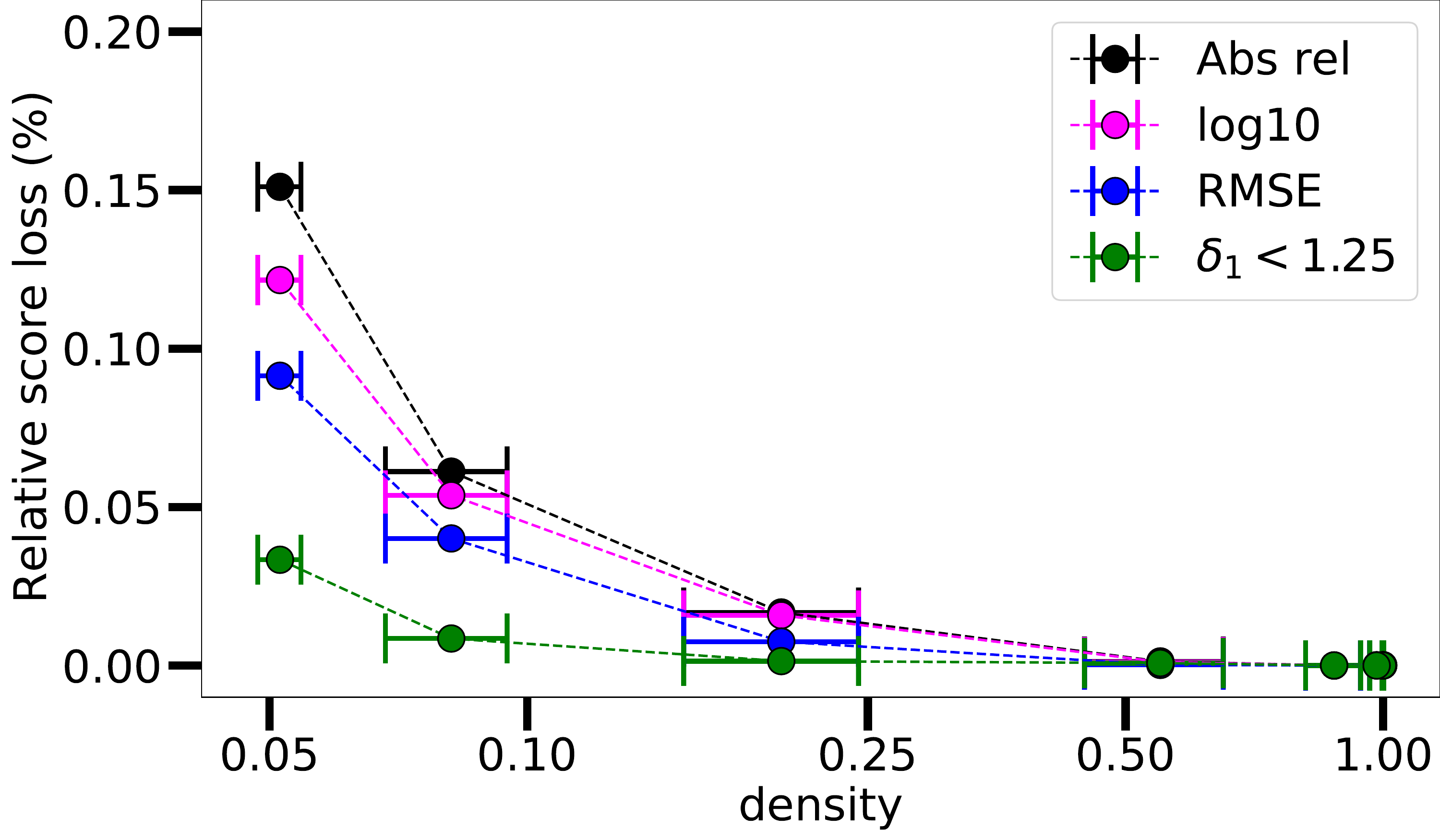}
	\vspace{-8pt}
	\caption{\textbf{Analysis of performance loss vs density on NYUv2}.  Using our wavelet representation, we can drop up to 95\% of the wavelet coefficients while suffering a maximum relative performance loss of less than 0.2\%.}
	\label{fig:relative_scores_vs_sparsity_nyu}
\end{figure}

\begin{figure}[!ht]
	\begin{center}
		\includegraphics[width=0.9\linewidth]{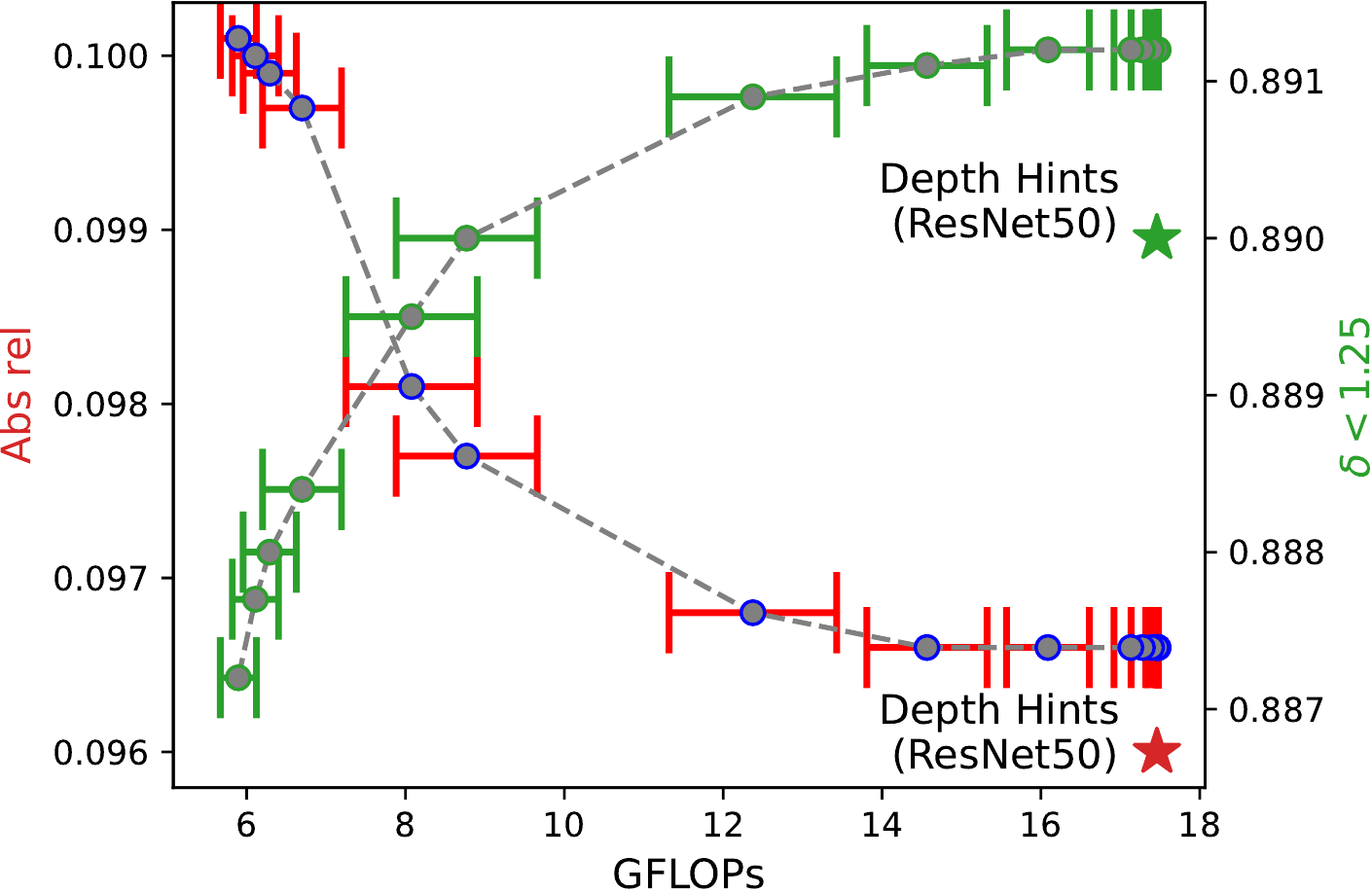}
	\end{center}
	\vspace{-17pt}
	\caption{\textbf{Analysis of performance vs decoder GFLOPs on KITTI.}
		By adjusting the parameter $\eta$, we trade off computation in GFLOPs (x-axis) against accuracy (y-axis; Abs Rel in red, and $\delta_1$ in green).
		We show here that we can reduce the computation by more than half while still remaining on par with our baseline.}
	\label{fig:scores_vs_flops}
	\vspace{-10pt}
\end{figure}

\begin{figure*}[!ht]
\centering
\renewcommand{\arraystretch}{0}
\setlength\tabcolsep{0pt} 
\begin{center}
\resizebox{1.0\textwidth}{!}{
\begin{tabular}{
>{\centering\arraybackslash}m{8em} 
>{\centering\arraybackslash}m{0.17\textwidth}
>{\centering\arraybackslash}m{0.17\textwidth}
>{\centering\arraybackslash}m{0.17\textwidth}@{\hskip 0.08in}
>{\centering\arraybackslash}m{0.11\textwidth}
>{\centering\arraybackslash}m{0.11\textwidth}
>{\centering\arraybackslash}m{0.11\textwidth}
}
~ & ~ & KITTI~\cite{Geiger2012CVPR} ~ &~ & ~ & NYUv2~\cite{Nyuv2} ~ &\\
RGB input & 
\begin{minipage}[b]{0.17\textwidth}
     \centering
     \includegraphics[width=\textwidth, height=1.3cm]{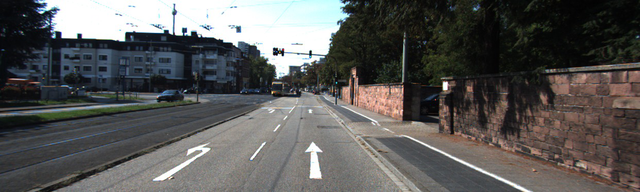}
\end{minipage} &
\begin{minipage}[b]{0.17\textwidth}
     \centering
     \includegraphics[width=\textwidth, height=1.3cm]{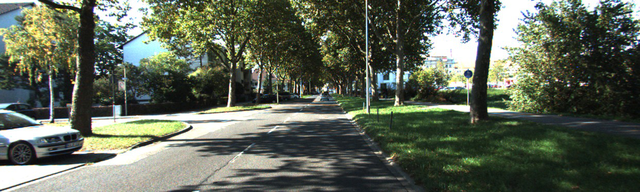}
\end{minipage} &
\begin{minipage}[b]{0.17\textwidth}
     \centering
     \includegraphics[width=\textwidth, height=1.3cm]{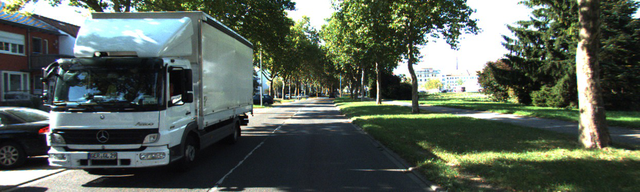}
\end{minipage} &
\begin{minipage}[b]{0.11\textwidth}
     \centering
     \includegraphics[width=\textwidth, height=1.3cm]{images/qualitative_nyu/299_rgb.png}
\end{minipage} &
\begin{minipage}[b]{0.11\textwidth}
     \centering
     \includegraphics[width=\textwidth, height=1.3cm]{images/qualitative_nyu/348_rgb.png}
\end{minipage} &
\begin{minipage}[b]{0.11\textwidth}
     \centering
     \includegraphics[width=\textwidth, height=1.3cm]{images/qualitative_nyu/333_rgb.png}
\end{minipage}\\
\cline{1-2}
\multirow{2}{*}[-1em]{\textsf{LL}$_3$ only} &
\begin{minipage}[b]{0.17\textwidth}
     \centering
     \includegraphics[width=\textwidth, height=1.3cm]{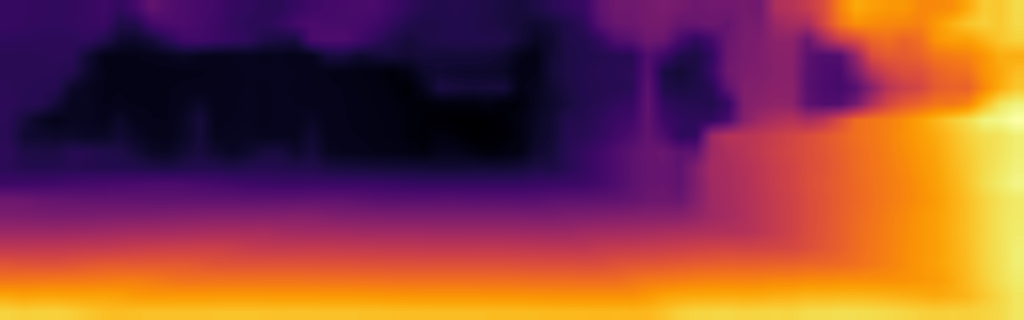}
\end{minipage} &
\begin{minipage}[b]{0.17\textwidth}
     \centering
     \includegraphics[width=\textwidth, height=1.3cm]{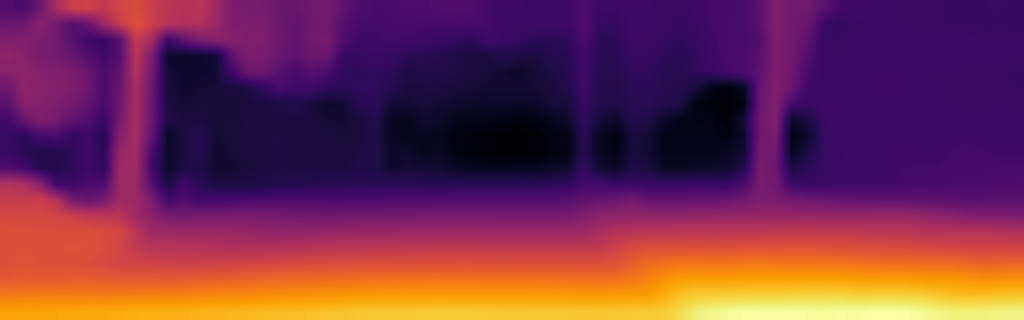}
\end{minipage} &
\begin{minipage}[b]{0.17\textwidth}
     \centering
     \includegraphics[width=\textwidth, height=1.3cm]{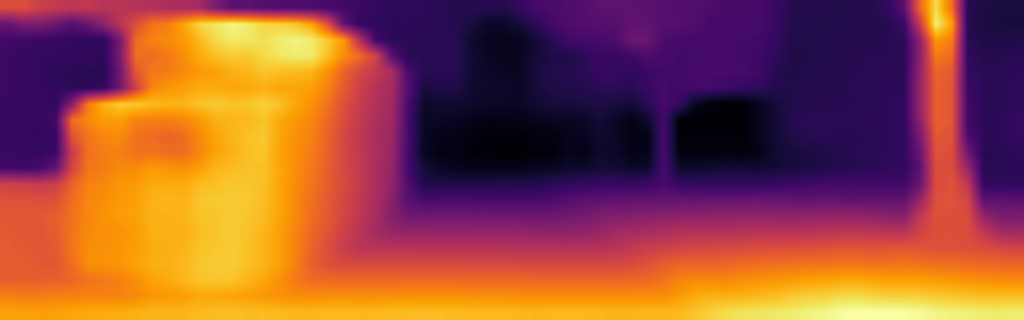}
\end{minipage} &
\begin{minipage}[b]{0.11\textwidth}
     \centering
     \includegraphics[width=\textwidth, height=1.3cm]{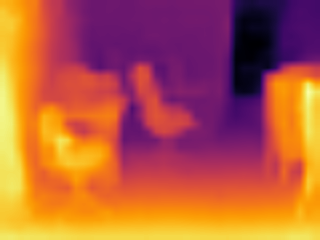}
\end{minipage} &
\begin{minipage}[b]{0.11\textwidth}
     \centering
     \includegraphics[width=\textwidth, height=1.3cm]{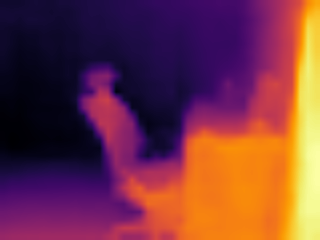}
\end{minipage} &
\begin{minipage}[b]{0.11\textwidth}
     \centering
     \includegraphics[width=\textwidth, height=1.3cm]{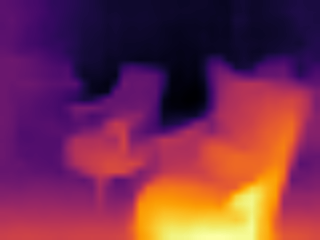}
\end{minipage}\\
~ & 
\begin{minipage}[b]{0.17\textwidth}
     \centering
     \includegraphics[width=\textwidth, height=1.3cm]{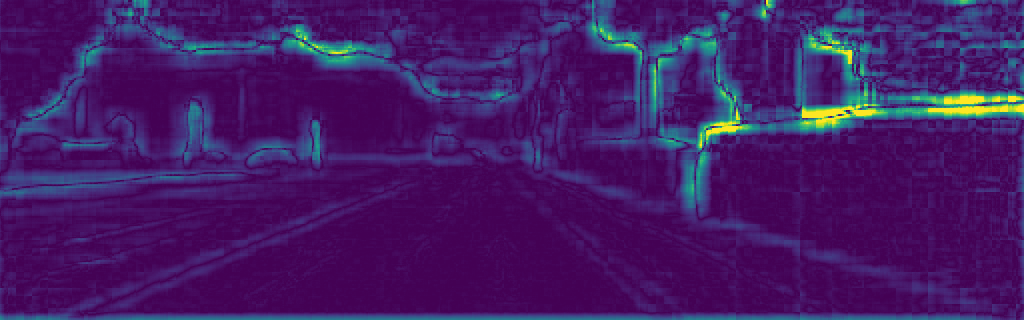}
\end{minipage} &
\begin{minipage}[b]{0.17\textwidth}
     \centering
     \includegraphics[width=\textwidth, height=1.3cm]{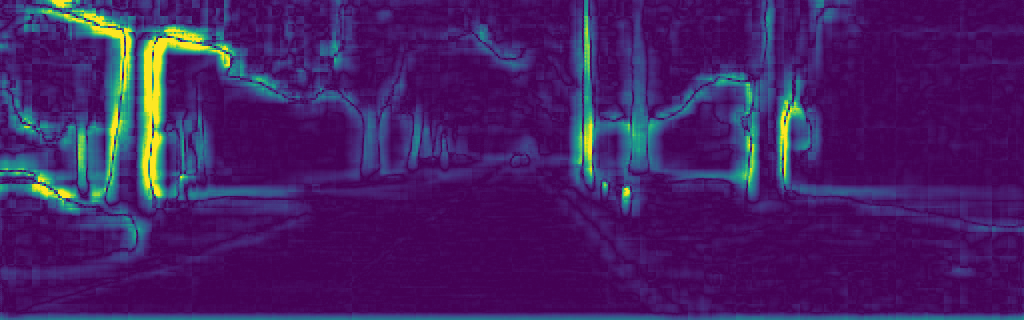}
\end{minipage} &
\begin{minipage}[b]{0.17\textwidth}
     \centering
     \includegraphics[width=\textwidth, height=1.3cm]{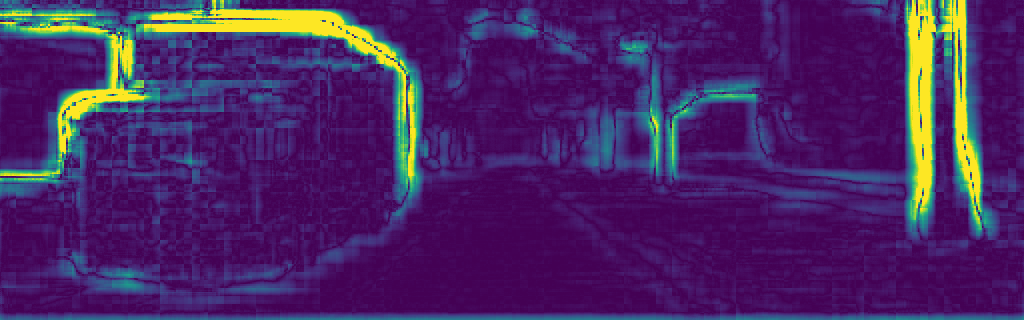}
\end{minipage} &
\begin{minipage}[b]{0.11\textwidth}
     \centering
     \includegraphics[width=\textwidth, height=1.3cm]{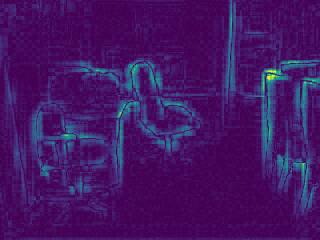}
\end{minipage} &
\begin{minipage}[b]{0.11\textwidth}
     \centering
     \includegraphics[width=\textwidth, height=1.3cm]{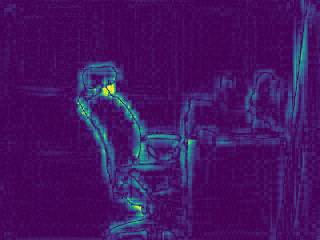}
\end{minipage} &
\begin{minipage}[b]{0.11\textwidth}
     \centering
     \includegraphics[width=\textwidth, height=1.3cm]{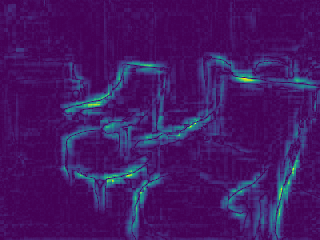}
\end{minipage}\\
\cline{1-2}
\multirow{2}{*}[-1em]{\shortstack[lb]{\textsf{LL}$_3$ \\ \{\textsf{LH}$_3$, \textsf{HL}$_3$, \textsf{HH}$_3$}\}} & 
\begin{minipage}[b]{0.17\textwidth}
     \centering
     \includegraphics[width=\textwidth, height=1.3cm]{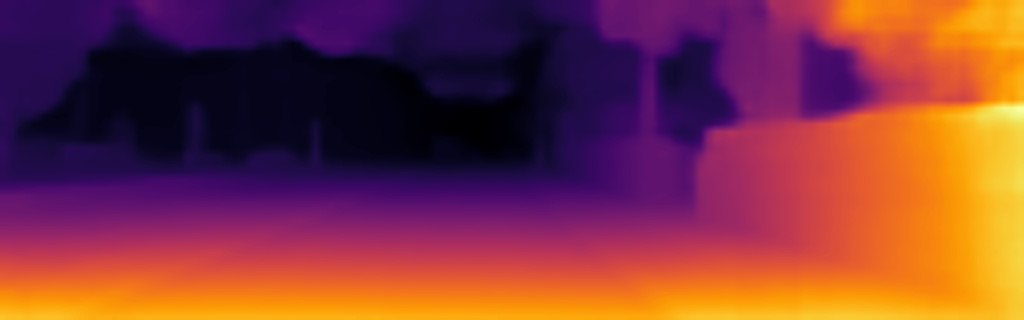}
\end{minipage} &
\begin{minipage}[b]{0.17\textwidth}
     \centering
     \includegraphics[width=\textwidth, height=1.3cm]{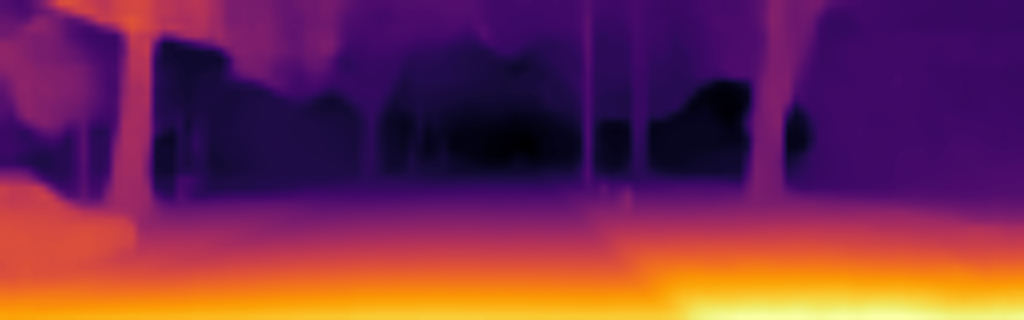}
\end{minipage} &
\begin{minipage}[b]{0.17\textwidth}
     \centering
     \includegraphics[width=\textwidth, height=1.3cm]{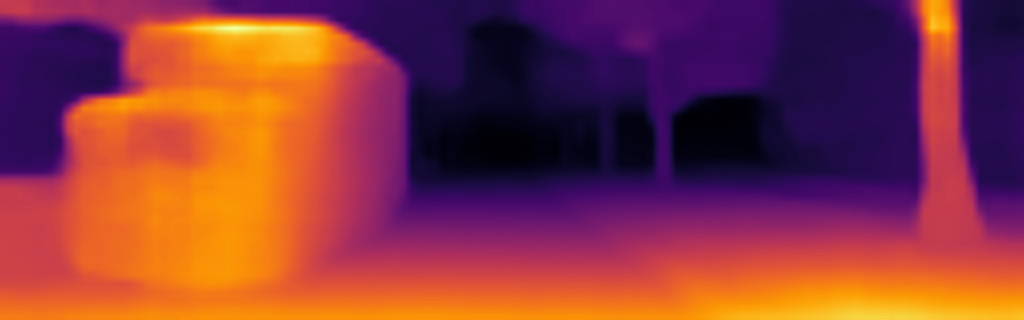}
\end{minipage} &
\begin{minipage}[b]{0.11\textwidth}
     \centering
     \includegraphics[width=\textwidth, height=1.3cm]{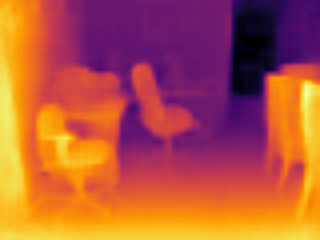}
\end{minipage} &
\begin{minipage}[b]{0.11\textwidth}
     \centering
     \includegraphics[width=\textwidth, height=1.3cm]{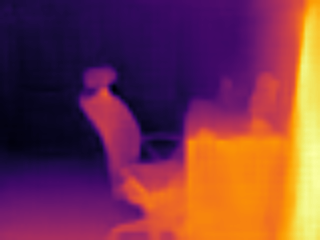}
\end{minipage} &
\begin{minipage}[b]{0.11\textwidth}
     \centering
     \includegraphics[width=\textwidth, height=1.3cm]{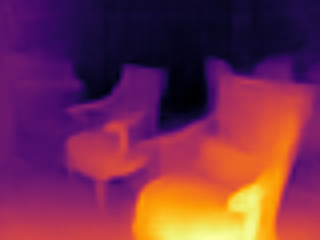}
\end{minipage}\\
~ &
\begin{minipage}[b]{0.17\textwidth}
     \centering
     \includegraphics[width=\textwidth, height=1.3cm]{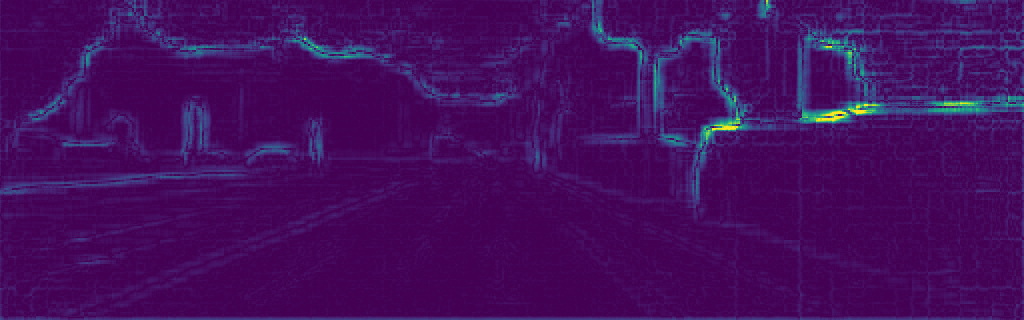}
\end{minipage} &
\begin{minipage}[b]{0.17\textwidth}
     \centering
     \includegraphics[width=\textwidth, height=1.3cm]{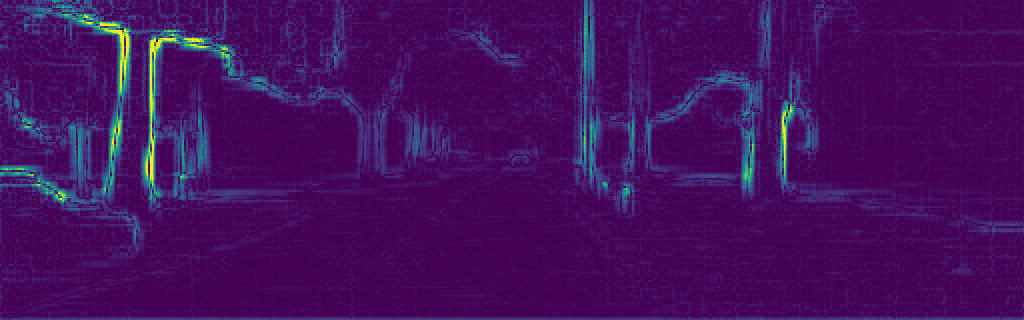}
\end{minipage} &
\begin{minipage}[b]{0.17\textwidth}
     \centering
     \includegraphics[width=\textwidth, height=1.3cm]{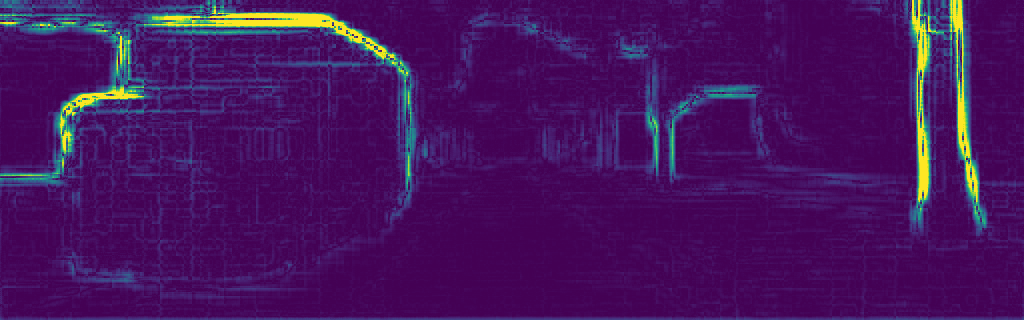}
\end{minipage} &
\begin{minipage}[b]{0.11\textwidth}
     \centering
     \includegraphics[width=\textwidth, height=1.3cm]{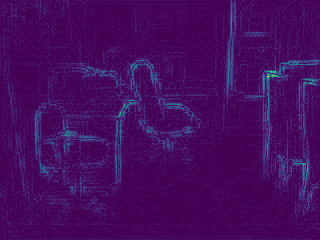}
\end{minipage} &
\begin{minipage}[b]{0.11\textwidth}
     \centering
     \includegraphics[width=\textwidth, height=1.3cm]{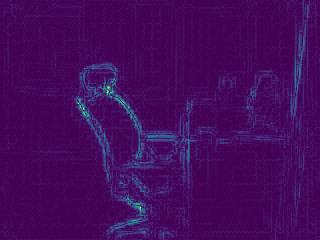}
\end{minipage} &
\begin{minipage}[b]{0.11\textwidth}
     \centering
     \includegraphics[width=\textwidth, height=1.3cm]{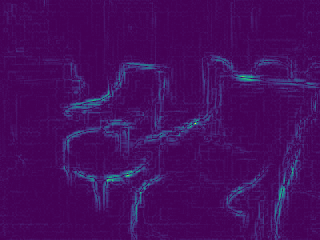}
\end{minipage}\\

\cline{1-2}
\multirow{2}{*}[-1em]{\shortstack[lb]{\textsf{LL}$_3$ + all \textsf{HF} \\ Sparse \\ ($\psi < 20\%$)}} &
\begin{minipage}[b]{0.17\textwidth}
     \centering
     \includegraphics[width=\textwidth, height=1.3cm]{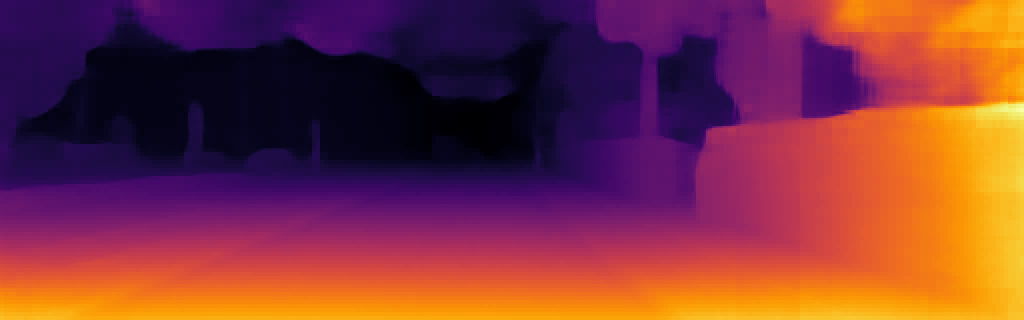}
\end{minipage} &
\begin{minipage}[b]{0.17\textwidth}
     \centering
     \includegraphics[width=\textwidth, height=1.3cm]{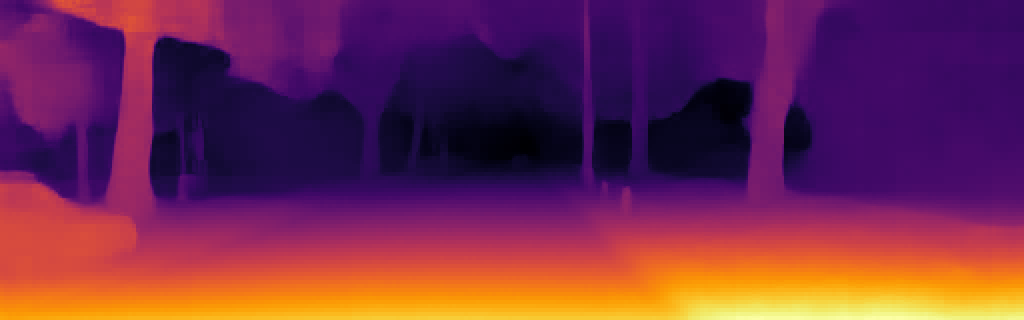}
\end{minipage} &
\begin{minipage}[b]{0.17\textwidth}
     \centering
     \includegraphics[width=\textwidth, height=1.3cm]{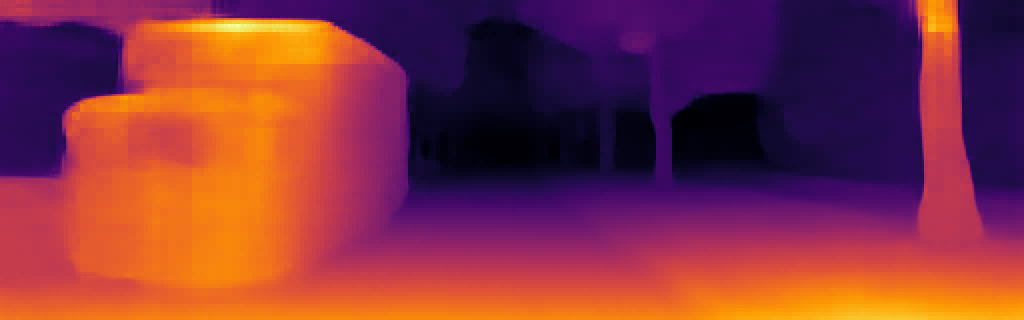}
\end{minipage} &
\begin{minipage}[b]{0.11\textwidth}
     \centering
     \includegraphics[width=\textwidth, height=1.3cm]{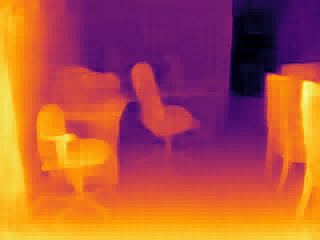}
\end{minipage} &
\begin{minipage}[b]{0.11\textwidth}
     \centering
     \includegraphics[width=\textwidth, height=1.3cm]{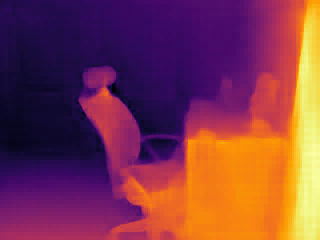}
\end{minipage} &
\begin{minipage}[b]{0.11\textwidth}
     \centering
     \includegraphics[width=\textwidth, height=1.3cm]{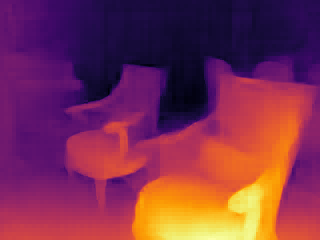}
\end{minipage}\\
~
&
\begin{minipage}[b]{0.17\textwidth}
     \centering
     \includegraphics[width=\textwidth, height=1.3cm]{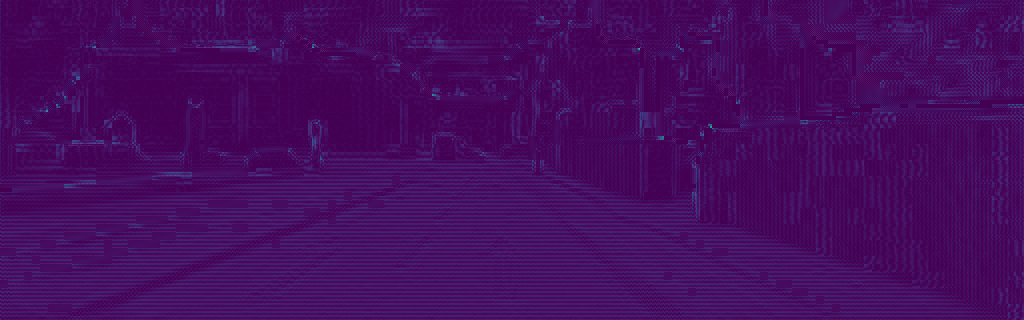}
\end{minipage} &
\begin{minipage}[b]{0.17\textwidth}
     \centering
     \includegraphics[width=\textwidth, height=1.3cm]{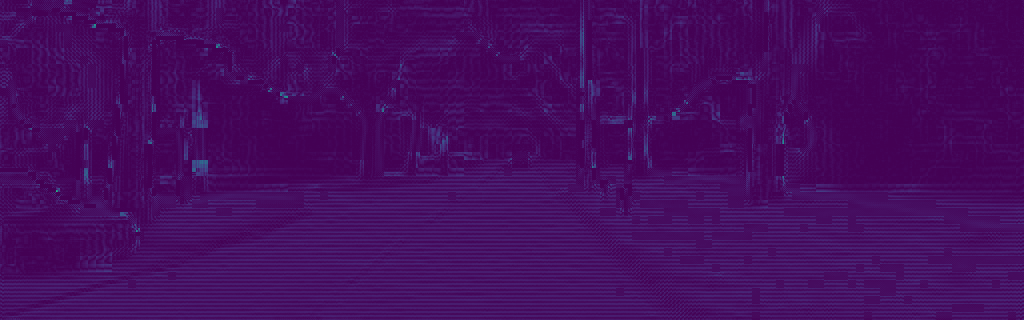}
\end{minipage} &
\begin{minipage}[b]{0.17\textwidth}
     \centering
     \includegraphics[width=\textwidth, height=1.3cm]{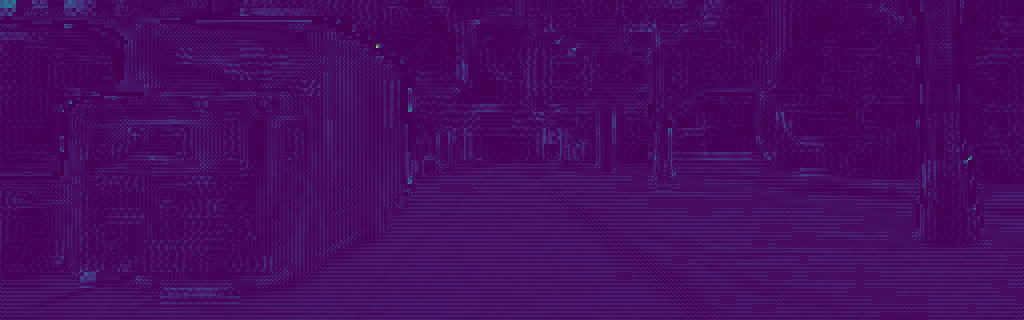}
\end{minipage} &
\begin{minipage}[b]{0.11\textwidth}
     \centering
     \includegraphics[width=\textwidth, height=1.3cm]{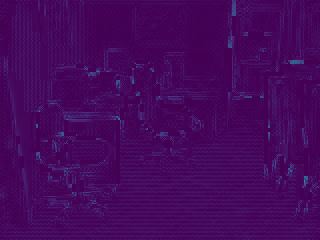}
\end{minipage} &
\begin{minipage}[b]{0.11\textwidth}
     \centering
     \includegraphics[width=\textwidth, height=1.3cm]{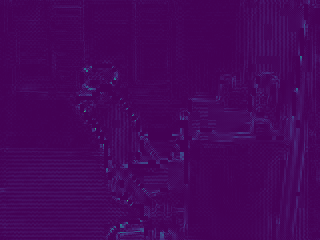}
\end{minipage} &
\begin{minipage}[b]{0.11\textwidth}
     \centering
     \includegraphics[width=\textwidth, height=1.3cm]{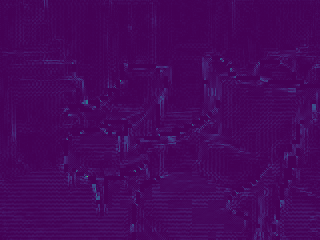}
\end{minipage}\\
\end{tabular}
}
\end{center}
    \vspace{-13pt}
    \caption{\textbf{Qualitative results on wavelet representation of depth maps}. When using only a subset of wavelet scales, we run the inverse wavelet transform up to the highest scale with those coefficients, then perform a bilinear upsampling up to the full resolution. For each experiment, the bottom line shows the $\ell_1$ error map between the considered depth maps and the depth map reconstructed with the complete (dense) set of predicted wavelet coefficients, which shows that wavelets contribute to refining details.}
    \label{fig:sparsity_qual}
    \vspace{-5pt}
\end{figure*}

\vspace{-5pt}
\paragraph{Wavelets enhance high-frequency details.} As mentioned in Section~\ref{sec:waveletmonodepth}, the wavelet representation of depth maps allows us to output depth at different resolutions, depending on how many levels of coefficients have been computed. 
Tables~\ref{tab:hf_ablation_bilinear_kitti} and~\ref{tab:hf_ablation_bilinear_nyu} demonstrate evaluation scores for depth maps produced at different levels of wavelet decomposition on the KITTI and NYUv2 datasets respectively. 
As can be seen, most of the signal is captured in low-frequency estimates of the depth map at the lowest resolution. This confirms previous works observations~\cite{Eigen2015PredictingDS, Chen2019SARPN} that a coarse estimate of depth is sufficient to capture the global geometry of the scene. Using more wavelet levels adds more high-frequency details to the depth map, yielding sharper results. Figure~\ref{fig:sparsity_qual} shows the sharpening effect of wavelets qualitatively on KITTI and NYUv2 images.

\begin{table}[!ht]
\begin{center}
\resizebox{\linewidth}{!}{
    \centering
    \scriptsize
    \begin{tabular}{|l||c|c|c|c|c|c|c|}
        \hline
        Activated HF & \cellcolor{col1}Abs$_{Rel}$ & \cellcolor{col1}Sq$_{Rel}$ & \cellcolor{col1}R & \cellcolor{col1}Rlog & \cellcolor{col2}$\delta_1$ & \cellcolor{col2}$\delta_2$ & \cellcolor{col2}$\delta_3$ \\ \hline\hline
        LL only &   0.104  &   0.668  &   4.415  &   0.179  &   0.878  &   0.962  &   0.985\\ \hline
        [3] &   0.097  &   0.659  &   4.321  &   0.177  &   0.887  &   0.964  &   0.984  
        \\ \hline
        [3, 2] &   0.096  &   0.679  &   4.333  &   0.179  &   0.890  &   0.963  &   0.983 \\ \hline
        [3, 2, 1] &   0.096  &   0.702  &   4.366  &   0.180  &   0.891  &   0.963  &   0.983  \\ \hline
        [3, 2, 1, 0] &   0.097  &   0.714  &   4.386  &   0.181  &   0.891  &   0.963  &   0.983  \\ \hline
    \end{tabular}
}
\end{center}
\vspace{-12pt}
\caption{\textbf{Ablation study on high frequency coefficients on KITTI.} While most of the relevant depth information is captured by the low-frequency estimate, predicting higher frequency coefficients increases accuracy. Results are evaluated without post-processing.
}
\vspace{-12pt}
\label{tab:hf_ablation_bilinear_kitti}
\end{table}
\begin{table*}[!ht]
\begin{center}
    \centering
    \footnotesize
    \begin{tabular}{|l||c|c|c|c|c|c|c|c|}
        \hline
        \multirow{2}{*}{Activated HF} & \multicolumn{6}{c|}{Depth Accuracy} & \multicolumn{2}{c|}{Occ. Boundaries}\\ \cline{2-9}
        ~ & \cellcolor{col1}Abs$_{Rel}$ & \cellcolor{col1}RMSE & \cellcolor{col1}log$_{10}$ & \cellcolor{col2}$\delta_1$ & \cellcolor{col2}$\delta_2$ & \cellcolor{col2}$\delta_3$ & \cellcolor{col1}$\epsilon_{acc}$ & \cellcolor{col1}$\epsilon_{comp}$ \\ \hline\hline
        \textsf{LL} only &  0.1281  &  0.5549  &  0.0548  &  0.8419  &  0.9674  &  0.9915  &  8.3672  &  9.8552  \\ \hline
        [3] &  0.1264  &  0.5517  &  0.0543  &  0.8446  &  0.9680  &  0.9917  &  3.3945  &  8.7933  \\ \hline
        [3, 2] &  0.1259  &  0.5512  &  0.0542  &  0.8451  &  0.9682  &  0.9917  &  2.1259  &  7.6702  \\ \hline
        [3, 2, 1] &  0.1258  &  0.5515  &  0.0542  &  0.8451  &  0.9681  &  0.9917  &  1.8070  &  7.1073  \\ \hline
    \end{tabular}
\end{center}
\vspace{-12pt}
\caption{\textbf{Ablation study on high frequency coefficients on NYU.} While most of the relevant depth information is captured by the low-frequency estimate, predicting higher frequency coefficients increases depth and occlusion boundaries accuracy. We evaluate occlusion boundary quality using metrics from Koch \textit{et al.}~\cite{Koch2018EvaluationOC, koch2020comparison} and the NYU-OC++ dataset manually annotated by Ramamonjisoa~\textit{et al.}~\cite{SharpNet2019, ramamonjisoa2020displacement}.
}
\vspace{-10pt}
\label{tab:hf_ablation_bilinear_nyu}
\end{table*}
\vspace{-10pt}
\paragraph{Wavelets are sparse.}
Next, we show that high-frequency coefficients are sparse.
As an example, Figure~\ref{fig:figure1}(b) shows one low-frequency and three high-frequency coefficient maps for a given depth map. We observe that the high-frequency maps have non-zero values near depth edges. More wavelet predictions can be found in supplementary. As depth edges are sparse, high-frequency coefficients at only a few pixel locations are necessary to produce high-accuracy depth maps. 

\vspace{-12pt}
\paragraph{Trading off accuracy against efficiency using sparsity.}
After training our network with standard convolutions, these are replaced with sparse ones as in Figure~\ref{fig:architecture} and Algorithm~\ref{algo:decoder}. Varying the threshold value $\eta$ allows us to vary the sparsity level $\psi$ in Equation~\eqref{eq:mac_sparse}, and consequently to trade off accuracy against complexity. 
Because wavelets are sparse, we can compute them only at a very small number of pixel locations and suffer a minimal loss in depth accuracy.
Figures~\ref{fig:relative_scores_vs_sparsity_kitti} and~\ref{fig:relative_scores_vs_sparsity_nyu} show relative score changes with varying sparsity threshold on KITTI and NYU datasets respectively. Note that a fixed value of $\eta$ produces different sparsity levels depending on the content of an image, so we also plot standard deviation of sparsity levels for each $\eta$ value. 
Figure~\ref{fig:relative_scores_vs_sparsity_kitti}  indicates that computing the wavelet coefficients at only 10 percent of pixel locations results in a relative loss in scores of less than $1.4\%$ for KITTI images.
Similarly, Figure~\ref{fig:relative_scores_vs_sparsity_nyu} shows that we can compute wavelet coefficients at only 5 percent of pixel locations while suffering a loss in scores of less than $0.20\%$ for NYU images.

Finally, we demonstrate how sparsity of high-frequency coefficient maps can be exploited for efficiency gains in the decoder. Figure~\ref{fig:scores_vs_flops} shows Abs~Rel and $\delta_1$ scores for varying $\eta$ used during prediction.
As can be seen, the score change is minimal when using half multiply-add operations in the decoder and the performance is comparable to SOTA methods using only a third of multiply-add operations. Note that biggest efficiency gains are obtained at higher resolution, as sparsity increases with resolution.

\begin{table*}[th]
  \centering
  \centering
  \footnotesize
  \resizebox{1.0\textwidth}{!}{
  \begin{tabular}{|c l|c|c|c||c|c|c|c|c|c|c|}
  \hline
  Cit. & Method & PP & Data & H $\times$ W & \cellcolor{col1}{\scriptsize Abs Rel} & \cellcolor{col1}{\scriptsize Sq Rel} & \cellcolor{col1}{\scriptsize RMSE}  & \cellcolor{col1}{\scriptsize RMSE log} & \cellcolor{col2}${\scriptstyle \delta < 1.25}$ & \cellcolor{col2}${\scriptstyle \delta < 1.25^{2}}$ & \cellcolor{col2}${\scriptstyle \delta < 1.25^{3}}$\\

\arrayrulecolor{black}
\hline
\hline

\cite{godard2019digging} & \cellcolor{ssectioncolor} Monodepth2 Resnet18 & \cmark & S & 192 $\times$ 640 &
0.108 & 0.842 & 4.891 & 0.207 & 0.866 & 0.949 & 0.976  \\  

& \cellcolor{ssectioncolor} \textbf{WaveletMonodepth} Resnet18\done & \cmark & S & 192 $\times$ 640 & 0.110  &   0.876  &   4.916  &   0.206  &   0.864  &   0.950  &   0.976\\

& \cellcolor{ssectioncolor} Monodepth2 Resnet50 & \cmark & S & 192 $\times$ 640 &
 0.108  &   0.802  &   4.577  &   0.185  &   0.886  &   0.963  &   0.983  \\

& \cellcolor{ssectioncolor} \textbf{WaveletMonodepth} Resnet50\done & \cmark & S & 192 $\times$ 640 &
0.106  &   0.824  &   4.824  &   0.205  &   0.870  &   0.949  &   0.975  \\  

\hline
\cite{watson-2019-depth-hints}& \cellcolor{ssectioncolor} Depth Hints & \cmark & S$_{SGM}$ & 192 $\times$ 640 &
0.106 & 0.780 & 4.695 & 0.193 & 0.875 & 0.958 & 0.980\\

& \cellcolor{ssectioncolor} \textbf{WaveletMonodepth} Resnet18\done & \cmark & S$_{SGM}$ & 192 $\times$ 640 &
0.106 & 0.813 & 4.693 & 0.193 & 0.876 & 0.957 & 0.980\\

& \cellcolor{ssectioncolor} Depth Hints Resnet50 & \cmark & S$_{SGM}$ & 192 $\times$ 640 &
0.102 & 0.762 & 4.602 & 0.189 & 0.880 & 0.960 & 0.981\\

& \cellcolor{ssectioncolor} \textbf{WaveletMonodepth} Resnet50\done & \cmark & S$_{SGM}$ & 192 $\times$ 640 &
 0.105  &   0.813  &   4.625  &   0.191  &   0.879  &   0.959  &   0.981  \\  
\hline

\cite{godard2019digging} & \cellcolor{mssectioncolor} Monodepth2 Resnet18 & \cmark & MS  & 192 $\times$ 640 &  0.104 & 0.786 & 4.687 & 0.194 & 0.876 & 0.958 & 0.980\\ 

& \cellcolor{mssectioncolor} \textbf{WaveletMonodepth} Resnet18\done & \cmark  & MS & 192 $\times$ 640 & 0.109  &   0.814  &   4.808  &   0.198  &   0.868  &   0.955  &   0.980 \\

\hline

\cite{watson-2019-depth-hints} & \cellcolor{mssectioncolor} Depth Hints & \cmark & MS + $S_{SGM}$ & 192 $\times$ 640 & 0.105 &    0.769 &   4.627 &    0.189 &   0.875 &   0.959 &   0.982 \\

& \cellcolor{mssectioncolor} \textbf{WaveletMonodepth} Resnet18\done & \cmark & MS + $S_{SGM}$ & 192 $\times$ 640 & 0.110  &   0.840  &   4.741  &   0.195  &   0.868  &   0.956  &   0.981 \\

\hline
\cite{godard2019digging} & \cellcolor{hrsectioncolor} Monodepth2 Resnet18& \cmark & S & 320 $\times$ 1024 &
0.105 & 0.822 & 4.692 & 0.199 & 0.876 & 0.954 & 0.977  \\  

& \cellcolor{hrsectioncolor} \textbf{WaveletMonodepth} Resnet18\done & \cmark & S & 320 $\times$ 1024 & 
0.105  &   0.797  &   4.732  &   0.203  &   0.869  &   0.952  &   0.977\\
\hline
\cite{watson-2019-depth-hints} & \cellcolor{hrsectioncolor} Depth Hints & \cmark & S$_{SGM}$ & 320 $\times$ 1024 &
0.099 &   0.723 &  4.445 &  0.187 & 0.886 & 0.961 & 0.982\\

& \cellcolor{hrsectioncolor} \textbf{WaveletMonodepth} Resnet18\done & \cmark & S$_{SGM}$ & 320 $\times$ 1024 &
0.102  &   0.739  &   4.452  &   0.188  &   0.883  &   0.960  &   0.981 \\

& \cellcolor{hrsectioncolor} Depth Hints Resnet50& \cmark & S$_{SGM}$ & 320 $\times$ 1024 &
\textbf{0.096} &   \textbf{0.710} & 4.393 &  0.185 & 0.890 & \textbf{0.962} & 0.981\\

& \cellcolor{hrsectioncolor} \textbf{WaveletMonodepth} Resnet50\done & \cmark & S$_{SGM}$ & 320 $\times$ 1024 &
0.097  &   0.718  &   \textbf{4.387}  &   \textbf{0.184}  &   \textbf{0.891}  &   \textbf{0.962}  &   \textbf{0.982}\\

& \cellcolor{hrsectioncolor} \textbf{WaveletMonodepth} Resnet50 ($\eta=0.05$) \done & \cmark & S$_{SGM}$ & 320 $\times$ 1024 &
0.100 &	0.726 &	4.444 &	0.186 &	0.888 &	\textbf{0.962} &	\textbf{0.982}\\
\arrayrulecolor{black}
\hline
  \end{tabular}}
    \vspace{-6pt}
  \caption{\textbf{Quantitative results on KITTI.} We compare our method to our baselines on KITTI~\cite{Geiger2012CVPR}, using the Eigen split. %
  The \textit{Data}  column indicates the training data modality: 
  S is for self-supervised training on stereo images, MS is for models trained with both monocular (forward and backward frames) and stereo data and S$_{SGM}$ refers to the extra stereo ground truth which was used in ~\cite{watson-2019-depth-hints}. 
  }
  \vspace{-3pt}
  \label{tab:kitti_eigen}
\end{table*}

\begin{table*}[th]
  \centering
  \footnotesize
  \begin{tabular}{|l|c||c|c|c|c|c|c|c|c|c|}
  \hline
  Method & H $\times$ W & \cellcolor{col1}{\scriptsize Abs Rel} &
  \cellcolor{col1}{\scriptsize RMSE}  & \cellcolor{col1}{\scriptsize $log_{10}$} & \cellcolor{col2}${\scriptstyle \delta < 1.25}$ & \cellcolor{col2}${\scriptstyle \delta < 1.25^{2}}$ & \cellcolor{col2}${\scriptstyle \delta < 1.25^{3}}$ & 
  \cellcolor{col1}$\epsilon_{acc}$ & \cellcolor{col1}$\epsilon_{comp}$\\
  \hline
\hline

DenseNet baseline
& 480 $\times$ 640 &
0.1277  &    \textbf{0.5479}  &    \textbf{0.0539}  &    0.8430  &    \textbf{0.9681}  &    \textbf{0.9917}  &    \textbf{1.7170}  &    \textbf{7.0638}\\
\textbf{WaveletMonodepth (last scale sup.) } & 480 $\times$ 640 &
0.1280  &    0.5589  &    0.0546  &    0.8436  &    0.9658  &    0.9908  &    1.7678  &    7.1433\\

\textbf{WaveletMonodepth}  & 480 $\times$ 640 &
\textbf{0.1258}  &    0.5515  &    0.0542  &    \textbf{0.8451}  &    \textbf{0.9681}  &    \textbf{0.9917}  &    1.8070  &    7.1073\\

\textbf{WaveletMonodepth}  ($\eta = 0.04$) & 480 $\times$ 640 &
0.1259  &    0.5517  &    0.0543  &    0.8450  &    \textbf{0.9681}  &    \textbf{0.9917}  &    1.8790  &    7.0746\\

\arrayrulecolor{black}\hline
  \end{tabular}
    \vspace{-5pt}
  \caption{\textbf{Quantitative results on NYUv2}~\cite{Nyuv2} 
  We compare our DenseDepth~\cite{Densedepth}-inspired baseline to our implementation with wavelets and with sparsity. All results are evaluated in the Eigen center crop, without post-processing. As in DenseDepth, our network outputs a $240\times320$ depth map which is then upsampled for evaluation.
  }
  \vspace{-12pt}
  \label{tab:nyu_eigen}
\end{table*}

\subsection{KITTI results}
We summarize our results on the KITTI dataset in Table~\ref{tab:kitti_eigen}. 
Here we show that our method, which simply replaces depth or disparity predictions with wavelet predictions, can be applied to a wide range of single image depth estimation models and losses.
In each section of the table, the off-the-shelf model numbers are reported, together with numbers from a model trained with our wavelet formulation.
For example, we demonstrate that wavelets can be used in self-supervised depth estimation frameworks such as Monodepth2~\cite{godard2019digging}, as well as its weakly-supervised extension Depth Hints~\cite{watson-2019-depth-hints}.
We note that we achieve our best results when using Depth Hints and high-resolution input images. 
This is not surprising, as supervision from SGM should give better scores, but more importantly using high resolution inputs and outputs allows for more sparsification, as edge pixels become sparser as resolution grows.
Importantly, we show overall that replacing fully convolutional layers with wavelets gives models with comparable performance to the off-the-shelf, non-wavelet baselines. 
We show qualitative results from KITTI in Figure~\ref{fig:sparsity_qual}~(left).

\subsection{NYUv2 results}
\vspace{-3pt}
Scores on NYUv2 are shown in Table~\ref{tab:nyu_eigen}. Our method performs on par with our baseline, which demonstrates that it is possible to estimate accurate depth and sparse wavelets without directly supervising the wavelet coefficients, in contrast with \cite{Yang2020CVPR}. 
In Table~\ref{tab:nyu_eigen}, we show that supervising depth only at the last scale performs on par with our network supervised at all scales, which shows that a full multi-scale wavelet reconstruction network can be trained end-to-end.
Qualitative results from NYUv2 are shown in Figure~\ref{fig:sparsity_qual}~(right).

\vspace{-5pt}
\section{Conclusion}
\vspace{-3pt}
In this work we combine wavelet representation with deep learning for a single-image depth prediction task. We demonstrate that a neural network can learn to predict wavelet coefficient maps through supervision of the reconstructed depth map with existing losses. Our experiments using KITTI and NYUv2 datasets show that we can achieve scores comparable to SOTA models using similar encoder-decoder neural network architectures to the baseline models, but with wavelet representations.

We also analyze sparsity of wavelet coefficients and show that sparsified wavelet coefficient maps can generate high-quality depth maps. Finally, we exploit this sparsity to reduce multiply-add operations in the decoder network by at least a factor of 2.

\vspace{-6pt}
\section*{Acknowledgements} 
\vspace{-6pt}
We would like to thank Aron Monszpart for helping us set up cloud experiments, and our reviewers for their useful suggestions.

{\small
\bibliographystyle{ieee_fullname}
\bibliography{biblio}

\begin{thebibliography}{10}\itemsep=-1pt

\bibitem{aleotti2021real}
Filippo Aleotti, Giulio Zaccaroni, Luca Bartolomei, Matteo Poggi, Fabio Tosi,
  and Stefano Mattoccia.
\newblock Real-time single image depth perception in the wild with handheld
  devices.
\newblock {\em Sensors}, 2021.

\bibitem{Densedepth}
Ibraheem Alhashim and Peter Wonka.
\newblock {High Quality Monocular Depth Estimation via Transfer Learning}.
\newblock {\em arXiv:1812.11941}, 2018.

\bibitem{bian2019unsupervised}
Jiawang Bian, Zhichao Li, Naiyan Wang, Huangying Zhan, Chunhua Shen, Ming-Ming
  Cheng, and Ian Reid.
\newblock Unsupervised scale-consistent depth and ego-motion learning from
  monocular video.
\newblock In {\em NeurIPS}, 2019.

\bibitem{Chen2019SARPN}
Xiaotian Chen, Xuejin Chen, and Zheng-Jun Zha.
\newblock Structure-aware residual pyramid network for monocular depth
  estimation.
\newblock 2019.

\bibitem{chen2019self}
Yuhua Chen, Cordelia Schmid, and Cristian Sminchisescu.
\newblock Self-supervised learning with geometric constraints in monocular
  video: Connecting flow, depth, and camera.
\newblock In {\em ICCV}, 2019.

\bibitem{clevert2015elus}
Djork-Arn{\'e} Clevert, Thomas Unterthiner, and Sepp Hochreiter.
\newblock Fast and accurate deep network learning by exponential linear units
  (elus).
\newblock {\em arXiv preprint arXiv:1511.07289}, 2015.

\bibitem{CotterPytorchWavelets}
Fergal Cotter.
\newblock Uses of complex wavelets in deep convolutional neural networks
  (doctoral thesis).
\newblock {\em https://doi.org/10.17863/CAM.53748}, Chapter 3, 2019.

\bibitem{Deng2019ICCV}
Xin Deng, Ren Yang, Mai Xu, and Pier~Luigi Dragotti.
\newblock Wavelet domain style transfer for an effective perception-distortion
  tradeoff in single image super-resolution.
\newblock In {\em ICCV}, 2019.

\bibitem{DonohoSoftThreshold}
David~L. {Donoho}.
\newblock De-noising by soft-thresholding.
\newblock {\em IEEE Transactions on Information Theory}, 41(3):613--627, 1995.

\bibitem{DonohoWaveletShrinkage}
David~L Donoho and Iain~M Johnstone.
\newblock {Ideal spatial adaptation by wavelet shrinkage}.
\newblock {\em Biometrika}, 81(3):425--455, 09 1994.

\bibitem{Eigen2015PredictingDS}
David Eigen and Rob Fergus.
\newblock {Predicting Depth, Surface Normals and Semantic Labels with a Common
  Multi-Scale Convolutional Architecture}.
\newblock In {\em ICCV}, 2015.

\bibitem{Eigen14}
David Eigen, Christian Puhrsch, and Rob Fergus.
\newblock {Depth Map Prediction from a Single Image Using a Multi-Scale Deep
  Network}.
\newblock In {\em NeurIPS}, pages 2366--2374, 2014.

\bibitem{fu2018deep}
Huan Fu, Mingming Gong, Chaohui Wang, Kayhan Batmanghelich, and Dacheng Tao.
\newblock Deep ordinal regression network for monocular depth estimation.
\newblock In {\em CVPR}, 2018.

\bibitem{garg2016unsupervised}
Ravi Garg, Vijay Kumar~BG, and Ian Reid.
\newblock Unsupervised {CNN} for single view depth estimation: Geometry to the
  rescue.
\newblock In {\em ECCV}, 2016.

\bibitem{Geiger2012CVPR}
Andreas Geiger, Philip Lenz, and Raquel Urtasun.
\newblock Are we ready for autonomous driving? the kitti vision benchmark
  suite.
\newblock In {\em CVPR}, 2012.

\bibitem{godard2017unsupervised}
Cl{\'e}ment Godard, Oisin Mac~Aodha, and Gabriel~J Brostow.
\newblock Unsupervised monocular depth estimation with left-right consistency.
\newblock In {\em CVPR}, 2017.

\bibitem{Monodepth17}
C. Godard, O. {Mac Aodha}, and G.~J. Brostow.
\newblock {Unsupervised Monocular Depth Estimation with Left-Right
  Consistency}.
\newblock In {\em CVPR}, 2017.

\bibitem{godard2019digging}
Cl{\'e}ment Godard, Oisin Mac~Aodha, Michael Firman, and Gabriel~J. Brostow.
\newblock Digging into self-supervised monocular depth estimation.
\newblock In {\em ICCV}, 2019.

\bibitem{gordon2019depth}
Ariel Gordon, Hanhan Li, Rico Jonschkowski, and Anelia Angelova.
\newblock Depth from videos in the wild: Unsupervised monocular depth learning
  from unknown cameras.
\newblock In {\em ICCV}, 2019.

\bibitem{guizilini20203d}
Vitor Guizilini, Rares Ambrus, Sudeep Pillai, Allan Raventos, and Adrien
  Gaidon.
\newblock {3D} packing for self-supervised monocular depth estimation.
\newblock In {\em CVPR}, 2020.

\bibitem{Guo2017DWSR}
Tiantong Guo, Hojjat~Seyed Mousavi, Tiep~Huu Vu, and Vishal Monga.
\newblock Deep wavelet prediction for image super-resolution.
\newblock In {\em CVPR Workshops}, 2017.

\bibitem{haar1910theorie}
Alfréd Haar.
\newblock Zur theorie der orthogonalen funktionensysteme.
\newblock {\em Mathematische Annalen}, 69:331--371, 1910.

\bibitem{He2016ResNet}
K. He, X. Zhang, S. Ren, and J. Sun.
\newblock {Deep Residual Learning for Image Recognition}.
\newblock In {\em CVPR}, 2016.

\bibitem{He2017ICCV}
Yihui He, Xiangyu Zhang, and Jian Sun.
\newblock Channel pruning for accelerating very deep neural networks.
\newblock In {\em ICCV}, Oct 2017.

\bibitem{he2017channel}
Yihui He, Xiangyu Zhang, and Jian Sun.
\newblock Channel pruning for accelerating very deep neural networks.
\newblock In {\em CVPR}, 2017.

\bibitem{hinton2015distilling}
Geoffrey Hinton, Oriol Vinyals, and Jeff Dean.
\newblock Distilling the knowledge in a neural network.
\newblock {\em arXiv:1503.02531}, 2015.

\bibitem{hirschmuller2005accurate}
Heiko Hirschmuller.
\newblock Accurate and efficient stereo processing by semi-global matching and
  mutual information.
\newblock In {\em CVPR}, 2005.

\bibitem{hirschmuller2007stereo}
Heiko Hirschmuller.
\newblock Stereo processing by semiglobal matching and mutual information.
\newblock {\em PAMI}, 2007.

\bibitem{holynski2018fast}
Aleksander Holynski and Johannes Kopf.
\newblock Fast depth densification for occlusion-aware augmented reality.
\newblock {\em ACM Transactions on Graphics (TOG)}, 2018.

\bibitem{howard2017mobilenets}
Andrew~G Howard, Menglong Zhu, Bo Chen, Dmitry Kalenichenko, Weijun Wang,
  Tobias Weyand, Marco Andreetto, and Hartwig Adam.
\newblock Mobilenets: Efficient convolutional neural networks for mobile vision
  applications.
\newblock {\em arXiv:1704.04861}, 2017.

\bibitem{huang2019densenet}
Gao Huang, Zhuang Liu, Geoff Pleiss, Laurens Van Der~Maaten, and Kilian
  Weinberger.
\newblock Convolutional networks with dense connectivity.
\newblock {\em IEEE Transactions on Pattern Analysis and Machine Intelligence},
  2019.

\bibitem{huang2017densenet}
Gao Huang, Zhuang Liu, Laurens van~der Maaten, and Kilian~Q Weinberger.
\newblock Densely connected convolutional networks.
\newblock In {\em Proceedings of the IEEE Conference on Computer Vision and
  Pattern Recognition}, 2017.

\bibitem{huang2017wavelet}
Huaibo Huang, Ran He, Zhenan Sun, and Tieniu Tan.
\newblock Wavelet-srnet: A wavelet-based cnn for multi-scale face super
  resolution.
\newblock In {\em ICCV}, 2017.

\bibitem{Kang2018FrameletDenoising}
Eunhee Kang, Won Chang, Jaejun Yoo, and Jong~Chul Ye.
\newblock Deep convolutional framelet denosing for low-dose {CT} via wavelet
  residual network.
\newblock {\em IEEE Transactions on Medical Imaging}, 37(6):1358--1369, 2018.

\bibitem{Kirillov2020CVPR}
Alexander Kirillov, Yuxin Wu, Kaiming He, and Ross Girshick.
\newblock Pointrend: Image segmentation as rendering.
\newblock In {\em CVPR}, 2020.

\bibitem{Koch2018EvaluationOC}
Tobias Koch, Lukas Liebel, Friedrich Fraundorfer, and Marco Körner.
\newblock {Evaluation of {CNN}-Based Single-Image Depth Estimation Methods}.
\newblock In {\em ECCV}, 2018.

\bibitem{koch2020comparison}
Tobias Koch, Lukas Liebel, Marco K{\"o}rner, and Friedrich Fraundorfer.
\newblock Comparison of monocular depth estimation methods using geometrically
  relevant metrics on the ibims-1 dataset.
\newblock {\em CVIU}, 2020.

\bibitem{kumar2018depthnet}
Arun~CS Kumar, Suchendra~M Bhandarkar, and Mukta Prasad.
\newblock {DepthNet}: A recurrent neural network architecture for monocular
  depth prediction.
\newblock In {\em CVPR Workshops}, 2018.

\bibitem{Laina2016DeeperDP}
I. Laina, C. Rupprecht, V. Belagiannis, F. Tombari, and N. Navab.
\newblock {Deeper Depth Prediction with Fully Convolutional Residual Networks}.
\newblock In {\em {3DV}}, 2016.

\bibitem{Lecun1995convolutional}
Yann LeCun and Yoshua Bengio.
\newblock {Convolutional Networks for Images, Speech, and Time Series}.
\newblock {\em The handbook of brain theory and neural networks}, 3361(10),
  1995.

\bibitem{Levin2004colorization}
Anat Levin, Dani Lischinski, and Yair Weiss.
\newblock {Colorization Using Optimization}.
\newblock In {\em ACM Transactions on Graphics}, 2004.

\bibitem{Li2020CVPR}
Qiufu Li, Linlin Shen, Sheng Guo, and Zhihui Lai.
\newblock Wavelet integrated cnns for noise-robust image classification.
\newblock In {\em CVPR}, June 2020.

\bibitem{liu2020waveletbased}
Lin Liu, Jianzhuang Liu, Shanxin Yuan, Gregory Slabaugh, Ales Leonardis,
  Wengang Zhou, and Qi Tian.
\newblock Wavelet-based dual-branch network for image demoireing.
\newblock In {\em ECCV}, 2020.

\bibitem{StructKDPAMI20}
Yifan Liu, Changyong Shun, Jingdong Wang, and Chunhua Shen.
\newblock Structured knowledge distillation for dense prediction.
\newblock {\em PAMI}, 2020.

\bibitem{Liu2017ICCV}
Zhuang Liu, Jianguo Li, Zhiqiang Shen, Gao Huang, Shoumeng Yan, and Changshui
  Zhang.
\newblock Learning efficient convolutional networks through network slimming.
\newblock In {\em ICCV}, 2017.

\bibitem{Luo2020CVPR}
Chenchi Luo, Yingmao Li, Kaimo Lin, George Chen, Seok-Jun Lee, Jihwan Choi,
  Youngjun~Francis Yoo, and Michael~O. Polley.
\newblock Wavelet synthesis net for disparity estimation to synthesize dslr
  calibre bokeh effect on smartphones.
\newblock In {\em CVPR}, 2020.

\bibitem{luo2019every}
Chenxu Luo, Zhenheng Yang, Peng Wang, Yang Wang, Wei Xu, Ram Nevatia, and Alan
  Yuille.
\newblock Every pixel counts++: Joint learning of geometry and motion with {3D}
  holistic understanding.
\newblock {\em PAMI}, 2019.

\bibitem{Luo2020CVPRW}
Xiaotong Luo, Jiangtao Zhang, Ming Hong, Yanyun Qu, Yuan Xie, and Cuihua Li.
\newblock Deep wavelet network with domain adaptation for single image
  demoireing.
\newblock In {\em CVPR Workshops}, 2020.

\bibitem{mayer2016Sceneflow}
Nikolaus Mayer, Eddy Ilg, Philip Hausser, Philipp Fischer, Daniel Cremers,
  Alexey Dosovitskiy, and Thomas Brox.
\newblock A large dataset to train convolutional networks for disparity,
  optical flow, and scene flow estimation.
\newblock In {\em Proceedings of the IEEE conference on computer vision and
  pattern recognition}, pages 4040--4048, 2016.

\bibitem{silberman2012indoor}
Pushmeet~Kohli Nathan~Silberman, Derek~Hoiem and Rob Fergus.
\newblock Indoor segmentation and support inference from rgbd images.
\newblock In {\em ECCV}, 2012.

\bibitem{Poggipydnet18}
Matteo Poggi, Filippo Aleotti, Fabio Tosi, and Stefano Mattoccia.
\newblock Towards real-time unsupervised monocular depth estimation on cpu.
\newblock In {\em IROS}, 2018.

\bibitem{ramamonjisoa2020displacement}
Michael Ramamonjisoa, Yuming Du, and Vincent Lepetit.
\newblock Predicting sharp and accurate occlusion boundaries in monocular depth
  estimation using displacement fields.
\newblock In {\em CVPR}, 2020.

\bibitem{SharpNet2019}
Michael Ramamonjisoa and Vincent Lepetit.
\newblock {SharpNet: Fast and Accurate Recovery of Occluding Contours in
  Monocular Depth Estimation}.
\newblock In {\em {ICCV Workshop}}, 2019.

\bibitem{ranjan2018adversarial}
Anurag Ranjan, Varun Jampani, Kihwan Kim, Deqing Sun, Jonas Wulff, and
  Michael~J Black.
\newblock Competitive collaboration: Joint unsupervised learning of depth,
  camera motion, optical flow and motion segmentation.
\newblock In {\em CVPR}, 2019.

\bibitem{Unet}
Olaf Ronneberger, Philipp Fischer, and Thomas Brox.
\newblock {U-Net: Convolutional Networks for Biomedical Image Segmentation}.
\newblock In {\em MICCAI}, 2015.

\bibitem{sandler2018mobilenetv2}
Mark Sandler, Andrew Howard, Menglong Zhu, Andrey Zhmoginov, and Liang-Chieh
  Chen.
\newblock {MobileNetV2}: Inverted residuals and linear bottlenecks.
\newblock In {\em CVPR}, 2018.

\bibitem{SaxenaMake3D}
Ashutosh Saxena, Min Sun, and Andrew~Y. Ng.
\newblock {Make3D: Learning {3D} Scene Structure from a Single Still Image}.
\newblock {\em IEEE TPAMI}, 2009.

\bibitem{Nyuv2}
Nathan Silberman, Derek Hoiem, Pushmeet Kohli, and Rob Fergus.
\newblock {Indoor Segmentation and Support Inference from RGBD Images}.
\newblock In {\em ECCV}, 2012.

\bibitem{CreamIROS2018}
Andrew Spek, Thanuja Dharmasiri, and Tom Drummond.
\newblock {CReaM}: Condensed real-time models for depth prediction using
  convolutional neural networks.
\newblock In {\em IROS}, 2018.

\bibitem{JPEG2000Book}
David Taubman and Michael Marcellin.
\newblock {\em JPEG2000 Image Compression Fundamentals, Standards and
  Practice}.
\newblock Springer Publishing Company, Incorporated, 2013.

\bibitem{uhrig2017sparse}
Jonas Uhrig, Nick Schneider, Lukas Schneider, Uwe Franke, Thomas Brox, and
  Andreas Geiger.
\newblock Sparsity invariant {CNNs}.
\newblock In {\em 3DV}, 2017.

\bibitem{UnserJPEG2000}
Michael Unser and Thierry Blu.
\newblock Mathematical properties of the jpeg2000 wavelet filters.
\newblock {\em IEEE Transactions on Image Processing}, 12(9):1080--1090, 2003.

\bibitem{watson-2019-depth-hints}
Jamie Watson, Michael Firman, Gabriel~J. Brostow, and Daniyar Turmukhambetov.
\newblock Self-supervised monocular depth hints.
\newblock In {\em ICCV}, 2019.

\bibitem{Wei16NIPS}
Wei Wen, Chunpeng Wu, Yandan Wang, Yiran Chen, and Hai Li.
\newblock Learning structured sparsity in deep neural networks.
\newblock In {\em NeurIPS}, 2016.

\bibitem{WofkFastDepth19}
{Wofk, Diana and Ma, Fangchang and Yang, Tien-Ju and Karaman, Sertac and Sze,
  Vivienne}.
\newblock {FastDepth: Fast Monocular Depth Estimation on Embedded Systems}.
\newblock In {\em ICRA}, {2019}.

\bibitem{xu2015empirical}
Bing Xu, Naiyan Wang, Tianqi Chen, and Mu Li.
\newblock Empirical evaluation of rectified activations in convolutional
  network.
\newblock {\em arXiv preprint arXiv:1505.00853}, 2015.

\bibitem{Yang2020CVPR}
Menglong Yang, Fangrui Wu, and Wei Li.
\newblock Waveletstereo: Learning wavelet coefficients of disparity map in
  stereo matching.
\newblock In {\em CVPR}, June 2020.

\bibitem{yang2018netadapt}
Tien-Ju Yang, Andrew Howard, Bo Chen, Xiao Zhang, Alec Go, Mark Sandler,
  Vivienne Sze, and Hartwig Adam.
\newblock Netadapt: Platform-aware neural network adaptation for mobile
  applications.
\newblock In {\em ECCV}, 2018.

\bibitem{ye2018rethinking}
Jianbo Ye, Xin Lu, Zhe Lin, and James~Z Wang.
\newblock Rethinking the smaller-norm-less-informative assumption in channel
  pruning of convolution layers.
\newblock {\em arXiv:1802.00124}, 2018.

\bibitem{yin2018geonet}
Zhichao Yin and Jianping Shi.
\newblock {GeoNet}: Unsupervised learning of dense depth, optical flow and
  camera pose.
\newblock In {\em CVPR}, 2018.

\bibitem{yu2019slimmable}
Jiahui Yu, Linjie Yang, Ning Xu, Jianchao Yang, and Thomas Huang.
\newblock Slimmable neural networks.
\newblock In {\em ICLR}, 2019.

\bibitem{zhang2018shufflenet}
Xiangyu Zhang, Xinyu Zhou, Mengxiao Lin, and Jian Sun.
\newblock {ShuffleNet}: An extremely efficient convolutional neural network for
  mobile devices.
\newblock In {\em CVPR}, 2018.

\bibitem{zhou2017unsupervised}
Tinghui Zhou, Matthew Brown, Noah Snavely, and David Lowe.
\newblock Unsupervised learning of depth and ego-motion from video.
\newblock In {\em CVPR}, 2017.

\end{thebibliography}
}

\section*{}
\newpage
~
\newpage

\textbf{\huge{Supplementary Material}}

\section{Network Architectures and Losses}
\subsection{On direct supervision of wavelet coefficients}

The previous work WaveletStereo~\cite{Yang2020CVPR} supervises its wavelet based stereo matching method with ground truth wavelet coefficients at the different levels of the decomposition. However, wavelets can only reliably be supervised when ground truth depth -or disparity- is provided and when it does not contain missing values or high-frequency noise, as they show on the synthetic SceneFlow~\cite{mayer2016Sceneflow} dataset. The sparsity of ground truth data in the KITTI dataset especially around edges makes it impossible to estimate reliably ground truth wavelet coefficients. On NYUv2, the noise in depth maps is also an issue for direct supervision of wavelets, e.g. with creases in the layout or inaccurate depth edges. This noise also prohibits the use of Semi Global Matching ground truth for wavelet coefficient supervision.

As we show in our work, supervising the network on wavelet reconstructions allows us to ignore missing values and be robust to noisy labels.
\blfootnote{$^*$Work done during an internship at Niantic}

\subsection{Experiments on KITTI}

\paragraph{Architecture}
The architecture we use for our experiments is a modification of the architecture used in \cite{godard2019digging}, as described in the main paper.
In Table~\ref{tab:architecture_KITTI_resnet}, we set out our decoder architecture in detail.

\paragraph{Self-supervised losses}

Our self-supervised losses are as described in \cite{godard2019digging}, which we repeat here for completeness.
Given a stereo pair of images ($I_L, I_R$), we train our network to predict a depth map $D_L$, pixel-aligned with the left image.
We also assume access to  the camera intrinsics $K$, and the relative camera transformation between the images in the stereo pair $T_{R \to L}$.
We use the network's current estimate of depth to synthesise an image $I_{R \to L}$, computed as
\begin{align}
    I_{R \to L} &= I_{R}\Big\langle \text{proj}(D_L, T_{R \to L}, K) \Big\rangle,
\end{align}
where $\text{proj}()$ are the 2D pixel coordinates obtained by projecting the depths $D_L$ into image $I_R$, and $\big\langle\big\rangle$ is the sampling operator.
We follow standard practice in training the model under a photometric reconstruction error $pe$, so our loss becomes
\begin{align}
    L_p &= pe(I_L, I_{R \to L}).
\end{align}
Following \cite{godard2019digging, chen2019self} etc.~we use a weighted sum of SSIM and L1 losses
\begin{align}
    pe(I_a, I_b) &= \alpha \frac{1 - \mathrm{SSIM}(I_a, I_b)}{2} + (1 - \alpha) \|I_a - I_b\|_1 \notag,
\end{align}
where $\alpha = 0.85$.
We additionally follow \cite{godard2019digging} in using the smoothness loss:
\begin{align}
L_s &= \left | \partial_x d^*_L   \right | e^{-\left | \partial_x I_L \right |} + \left | \partial_y d^*_L   \right | e^{-\left | \partial_y I_L \right |},
\end{align} 
where $d^*_L = d_L / \overline{d_L}$ is the mean-normalized inverse depth for image $I_L$.

When we train on  monocular and stereo sequences~(`MS'), we again follow \cite{godard2019digging} --- see our main paper for an overview, and \cite{godard2019digging} for full details.

\paragraph{Depth Hints loss}

When we train with depth hints, we use the proxy loss from \cite{watson-2019-depth-hints}, which we recap here.
For stereo training pairs, we compute a proxy depth map $\tilde{D}_L$ using semi-global matching \cite{hirschmuller2005accurate}, an off-the-shelf stereo matching algorithm.
We use this to create a second synthesized image
\begin{align}
    \tilde{I}_{R \to L} &= I_{R}\Big\langle \text{proj}(\tilde{D}_L, T_{R \to L}, K) \Big\rangle,
\end{align}
We decide whether or not to apply a supervised loss using $\tilde{D}_L$ as ground truth on a \emph{per-pixel} basis.
We only add this supervised loss for pixels where $pe(I_L, \tilde{I}_{R \to L})$ is lower $pe(I_L, I_{R \to L})$.
The supervised loss term we use is $\log L_1$, following \cite{watson-2019-depth-hints}.
For experiments where Depth Hints are used for training, we disable the smoothness loss term.

\begin{table}[t]
  \centering
  \resizebox{1.0\columnwidth}{!}{
          
        \begin{tabular}[t]{|l|l|l|l|l|l|l|}
        \hline
        \multicolumn{7}{|l|}{\textbf{Depth Decoder}} \\
        \hline
        \textbf{layer} & \textbf{k} & \textbf{s} & \textbf{chns} & \textbf{res} & \textbf{input}   & \textbf{activation}    \\ \hline
        upconv5       & 3      & 1      & 256      & 32    & econv5                    & ELU \cite{clevert2015elus} \\ \hline
        \multicolumn{7}{|l|}{Level 3 coefficients predictions} \\ \hline
        iconv4        & 3      & 1      & 256      & 16    & $\uparrow$upconv5, econv4 & ELU \\ 
        
        \textbf{disp4}         & 3      & 1      & 1        & 16    & iconv4                    & Sigmoid \\
        \textbf{wave4}         & 3      & 1      & 3        & 16    & iconv4                    & Sigmoid \\
        upconv4       & 3      & 1      & 128    & 16    & iconv4                    & ELU \\ 
        $\IDWT_3$     & -      & -      & 1      & 8     & \textbf{disp4}, \textbf{wave4} & -\\
        \hline
        
        \multicolumn{7}{|l|}{Level 2 coefficients predictions} \\ \hline
        
        iconv3        & 3      & 1      & 128      & 8     & $\uparrow$upconv4, econv3 & ELU \\ 
        
        
        \textbf{wave3}         & 3      & 1      & 3        & 8     & iconv3                    & Sigmoid  \\ 
        upconv3       & 3      & 1      & 64       & 8     & iconv3                    & ELU \\
        $\IDWT_2$ & - & - & 1 & 8 & $\IDWT_3$, \textbf{wave3} & -\\
        \hline
        
        \multicolumn{7}{|l|}{Level 1 coefficients predictions} \\ \hline
        
        iconv2        & 3      & 1      & 64       & 4     & $\uparrow$upconv3, econv2 & ELU \\ 
        
        \textbf{wave2}         & 3      & 1      & 3        & 4     & iconv2                    & Sigmoid \\ 
        upconv2       & 3      & 1      & 32       & 4     & iconv2                    & ELU \\ 
        $\IDWT_1$ & - & - & 1 & 8 & $\IDWT_2$, \textbf{wave2} & -\\
        \hline
        
        \multicolumn{7}{|l|}{Level 0 coefficients predictions} \\ \hline
        
        iconv1        & 3      & 1      & 32       & 2     & $\uparrow$upconv2, econv1 & ELU \\
        
        \textbf{wave1}         & 3      & 1      & 3        & 2     & iconv1                    & Sigmoid \\ 
        $\IDWT_0$ & - & - & - & 1 & $\IDWT_1$, \textbf{wave1} & -\\
        \hline
        \end{tabular}  

 }
  \caption{\textbf{Our decoder network architecture for experiments on the KITTI~\cite{Geiger2012CVPR} dataset using ResNet backbone} Here \textbf{k} is the kernel size, \textbf{s} the stride, \textbf{chns} the number of output channels for each layer, \textbf{res} is the downscaling factor for each layer relative to the input image, and \textbf{input} corresponds to the input of each layer where $\uparrow$ is a $2\times$ nearest-neighbor upsampling of the layer. \textbf{disp4} is used produce the low-resolution estimate \lli{3}, while \textbf{waveJ} is used to decode $\{\text{\lhi{\textbf{J}}}, \text{\hli{\textbf{J}}}, \text{\hhi{\textbf{J}}}\}$ at level \textbf{J}. \textbf{disp4} and \textbf{waveJ} are convolution blocks detailed in Table~\ref{tab:waveconv_KITTI}. 
  }
\label{tab:architecture_KITTI_resnet}
\end{table}

\begin{table}[t]
  \centering
  \resizebox{1.0\columnwidth}{!}{
    \begin{tabular}[t]{|l|l|l|l|l|l|l|}
        \hline
        \multicolumn{7}{|l|}{\textbf{disp4 Layer}} \\
        \hline
        \textbf{layer} & \textbf{k} & \textbf{s} & \textbf{chns} & \textbf{res} & \textbf{input}   & \textbf{activation}    \\ \hline
        disp4(1)       & 1      & 1     & \textbf{chns}(iconv5) / 4  & 16 & iconv5 & LeakyReLU(0.1)~\cite{xu2015empirical} \\
        disp4(2)       & 3      & 1  & 1    & 16 & disp4-1 & Sigmoid \\
        \hline
        \multicolumn{7}{l}{~} \\
        \hline
        \multicolumn{7}{|l|}{\textbf{Wavelet Decoding Layer - waveJ}} \\
        \hline
        \textbf{layer} & \textbf{k} & \textbf{s} & \textbf{chns} & \textbf{res} & \textbf{input}   & \textbf{activation}    \\ \hline
        wave\textbf{J}(1+)       & 1      & 1      & \textbf{chns}(iconv[\textbf{J}+1])      & $2^{\boldsymbol{J}}$   & iconv[\textbf{J}+1]                     & LeakyReLU(0.1) \\
        wave\textbf{J}(2+)       & 3      & 1      & 3  & $2^{\boldsymbol{J}}$   & wave\textbf{J}(1+) & Sigmoid \\ \hline
        wave\textbf{J}(1-)       & 1      & 1      & \textbf{chns}(iconv[\textbf{J}+1])      & $2^{\boldsymbol{J}}$   & iconv[\textbf{J}+1]                     & LeakyReLU(0.1) \\
        wave\textbf{J}(2-)       & 3      & 1      & 3  & $2^{\boldsymbol{J}}$   & wave\textbf{J}(1-) & Sigmoid \\ \hline
        \multirow{2}{*}{substract} & \multirow{2}{*}{1} & \multirow{2}{*}{1} & \multirow{2}{*}{3} & \multirow{2}{*}{$2^{\boldsymbol{J}}$} & wave\textbf{J}(2+), & \multirow{2}{*}{Linear} \\
        ~ & ~ & ~ & ~ & ~ & wave\textbf{J}(2-) & ~\\
        \hline
    \end{tabular}
    }
  \caption{\textbf{Architecture of our wavelet decoding layer used for KITTI experiments} J denotes the level of the decoder. \textbf{disp4} is used produce the low-resolution estimate \lli{3}, while \textbf{waveJ} is used to decode $\{\text{\lhi{\textbf{J}}}, \text{\hli{\textbf{J}}}, \text{\hhi{\textbf{J}}}\}$.
  } 
\label{tab:waveconv_KITTI}
\end{table}

\paragraph{Additional experiments} We additionally tried training using edge-aware sparsity constraints that penalize non-zero coefficients at non-edge regions, by replacing depth gradients with wavelets coefficients in Monodepth's~\cite{Monodepth17} \emph{disparity smoothness loss}, which unfortunately made training unstable. We also tried to supervise wavelet coefficients using distillation~\cite{hinton2015distilling, aleotti2021real} from a teacher depth network, which resulted in lower performances.

\subsection{Experiments on NYUv2}
\paragraph{Architecture}

We adapted our architecture from the PyTorch implementation of DenseDepth~\cite{Densedepth}. Our implementation uses a DenseNet161 encoder instead of a DenseNet169, and a standard decoder with up-convolutions. We first design a baseline that does not use wavelets, using the architecture detailed in Table~\ref{tab:architecture_NYU_densenet_baseline}. Our wavelet adaptation of that baseline is then detailed in Table~\ref{tab:architecture_NYU_densenet}. For experiments reported in the main paper, we follow the DenseDepth strategy and predict outputs at half the input resolution. Hence, the last level of the depth decoder in Table~\ref{tab:architecture_NYU_densenet} is discarded. For experiments using a light-weight decoder discussed later in Section~\ref{ssec:low_res}, which predicts $224\times224$ depth maps given a $224\times224$ input image, we keep all four levels of wavelet decomposition.

\begin{table}[t]
  \centering
  \resizebox{1.0\columnwidth}{!}{
          
        \begin{tabular}[t]{|l|l|l|l|l|l|l|}
        \hline
        \multicolumn{7}{|l|}{\textbf{Depth Decoder}} \\
        \hline
        \textbf{layer} & \textbf{k} & \textbf{s} & \textbf{chns} & \textbf{res} & \textbf{input}   & \textbf{activation}    \\ \hline
        upconv5       & 3      & 1      & 1104      & 32    & econv5                    & Linear \\
        iconv4        & 3      & 1      & 552      & 16    & $\uparrow$upconv5, econv4 & LeakyReLU(0.2) \\
        
        
        iconv3        & 3      & 1      & 276      & 8     & $\uparrow$iconv4, econv3 & LeakyReLU(0.2) \\ 
        
        
        
        
        iconv2        & 3      & 1      & 138       & 4     & $\uparrow$iconv3, econv2 & LeakyReLU(0.2) \\
        
        
        
        iconv1        & 3      & 1      & 69       & 2     & $\uparrow$iconv2, econv1 & LeakyReLU(0.2) \\
        outconv0        & 1      & 1      & 1       & 2     & iconv1 & Linear \\ \hline
        \end{tabular}  

 }
  \caption{\textbf{Architecture of our DenseNet baseline decoder for experiments on the NYUv2~\cite{Nyuv2} dataset} Note that as in DenseDepth~\cite{Densedepth} we produce a depth map at half-resolution. Table adapted from \cite{godard2019digging}.
  }
\label{tab:architecture_NYU_densenet_baseline}
\end{table}

\begin{table}[t]
  \centering
  \resizebox{1.0\columnwidth}{!}{
          
        \begin{tabular}[t]{|l|l|l|l|l|l|l|}
        \hline
        \multicolumn{7}{|l|}{\textbf{Depth Decoder}} \\
        \hline
        \textbf{layer} & \textbf{k} & \textbf{s} & \textbf{chns} & \textbf{res} & \textbf{input}   & \textbf{activation}    \\ \hline
        upconv5       & 3      & 1      & 1104      & 32    & econv5                    & Linear\\ \hline
        \multicolumn{7}{|l|}{Level 3 coefficients predictions} \\ \hline
        iconv4        & 3      & 1      & 552      & 16    & $\uparrow$upconv5, econv4 & LeakyReLU(0.2) \\ 
        
        disp4         & 1      & 1      & 1        & 16    & upconv5                    & Linear \\
        wave4         & 3      & 1      & 3        & 16    & upconv5                    & Linear \\
        $\IDWT_3$ & - & - & 1 & 8 & disp4, wave4 & - \\ \hline
        
        
        \multicolumn{7}{|l|}{Level 2 coefficients predictions} \\ \hline
        iconv3        & 3      & 1      & 276      & 8     & $\uparrow$iconv4, econv3 & LeakyReLU(0.2) \\ 
        
        
        wave3         & 3      & 1      & 3        & 8     & iconv3                    & Linear  \\
        $\IDWT_2$ & - & - & 1 & 4 & $\IDWT_3$, wave3 & - \\ \hline
        
        \multicolumn{7}{|l|}{Level 1 coefficients predictions} \\ \hline
        
        iconv2        & 3      & 1      & 138       & 4     & $\uparrow$iconv2, econv2 & LeakyReLU(0.2) \\ 
        
        wave2         & 3      & 1      & 3        & 4     & iconv2                    & Linear \\
        $\IDWT_1$ & - & - & 1 & 2 & $\IDWT_2$, wave2 & - \\ \hline
        
        
        
        \end{tabular}  

 }
  \caption{\textbf{Our decoder network architecture for experiments on the NYUv2~\cite{Nyuv2} dataset} Note that since like in DenseDepth~\cite{Densedepth} we produce a depth map at half-resolution, we only need to predict wavelet coefficients until quarter-resolution. Table adapted from \cite{godard2019digging}.
  }
\label{tab:architecture_NYU_densenet}
\end{table}

\paragraph{Supervised losses}
For our NYU results in the main paper, we supervise depth using an L1 loss and SSIM: 
\begin{equation}
    L_D(y, y^*) = \lambda_1 \ell_1(y, y^*)
\label{eq:loss_nyu}
\end{equation}
where $y$ and $y^*$ are respectively predicted and ground truth depth and $\lambda_1=0.1$. Similar to \cite{SharpNet2019,ramamonjisoa2020displacement}, we clamp depth between 0.4 and 10 meters.

\section{Scores on Improved KITTI Ground Truth}

We report results on the improved KITTI ground truth~\cite{uhrig2017sparse} in Table~\ref{tab:kitti_eigen_improved}. 
As we saw in the main paper, our method is competitive on scores with non-wavelets baselines, but as we have shown our wavelet decomposition enables more efficient predictions.

\section{Qualitative Results}

In this section, we show qualitative results of our method.

In Figures~\ref{fig:qualitative_wavelets_nyu1}-\ref{fig:qualitative_wavelets_nyu2}-\ref{fig:qualitative_wavelets_nyu3} and Figures~\ref{fig:qualitative_wavelets_kitti1}-\ref{fig:qualitative_wavelets_kitti2}-\ref{fig:qualitative_wavelets_kitti3} we first show our sparse prediction process with corresponding sparse wavelets and masks, on the NYUv2 and KITTI datasets respectively. While we only need to compute wavelet coefficients in less than 10\% of pixel locations in the decoding process, we show that our wavelets efficiently retain relevant information. Furthermore, we show that wavelets efficiently detect depth edges and their orientation. Therefore, future work could make efficient use of our wavelet based depth estimation method for occlusion boundary detection.

\begin{figure*}[ht]
	\centering
	\renewcommand{\arraystretch}{2}
	
	\setlength\tabcolsep{3pt} 
	\begin{tabular}{>{\centering\arraybackslash}m{0.99\textwidth}}
		\centering
		\begin{tabular}{>{\centering\arraybackslash}m{0.32\textwidth}
				>{\centering\arraybackslash}m{0.32\textwidth}
				>{\centering\arraybackslash}m{0.32\textwidth}}
			Input image &  Baseline model & Wavelets prediction
		\end{tabular}\\
		\includegraphics[width=0.99\linewidth, valign=b]{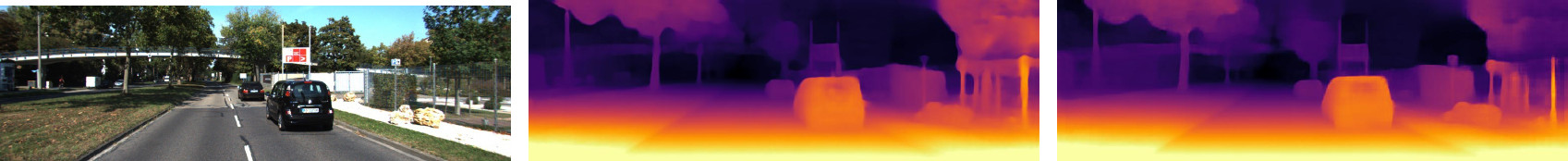} \\
		\includegraphics[width=0.99\linewidth, valign=b]{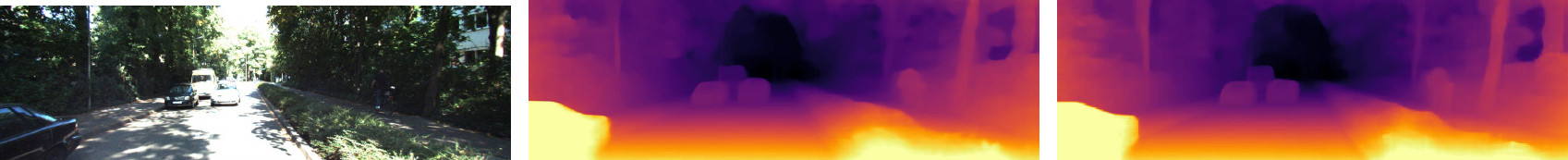} \\
		\includegraphics[width=0.99\linewidth, valign=b]{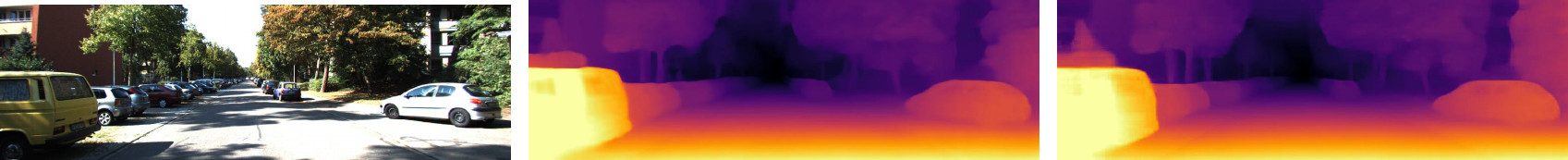} \\
		\includegraphics[width=0.99\linewidth, valign=b]{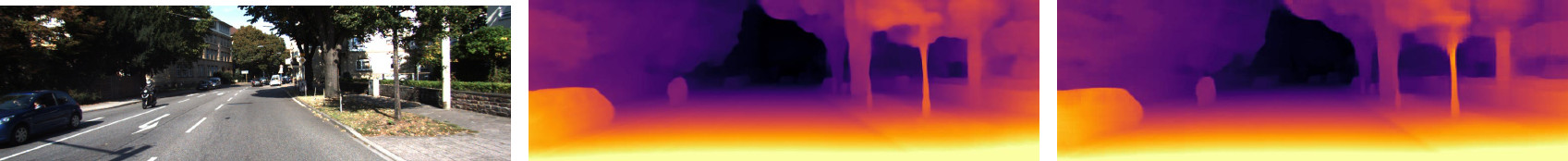} \\
		\includegraphics[width=0.99\linewidth, valign=b]{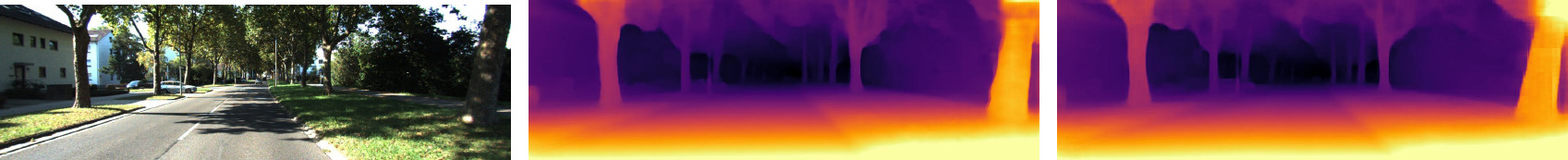} \\
	\end{tabular} 
	\caption{\textbf{Comparing wavelet predictions to a baseline model on the KITTI dataset}. On the left we show the input image, and in the middle column we show the prediction from an off-the-shelf Depth Hints ResNet 50 model \cite{watson-2019-depth-hints}. On the right we show an equivalently trained ResNet 50 model, but with our wavelets in the decoder. We see that our predictions retain the high quality of the baseline predictions, but are more efficient to predict.}
	\label{fig:qualitative_results_compare_depthhints}
	\vspace{10pt}
\end{figure*}

In Figure~\ref{fig:qualitative_results_compare_depthhints}, we show comparative results between our baseline Depth Hints~\cite{watson-2019-depth-hints} and our wavelet based method. 

\begin{table*}[!ht]
  \centering
  \centering
  \footnotesize
  \resizebox{1.0\textwidth}{!}{
  \begin{tabular}{|c l|c|c|c||c|c|c|c|c|c|c|}
  \hline
  Cit. & Method & PP & Data & H $\times$ W & \cellcolor{col1}{\scriptsize Abs Rel} & \cellcolor{col1}{\scriptsize Sq Rel} & \cellcolor{col1}{\scriptsize RMSE}  & \cellcolor{col1}{\scriptsize RMSE log} & \cellcolor{col2}${\scriptstyle \delta < 1.25}$ & \cellcolor{col2}${\scriptstyle \delta < 1.25^{2}}$ & \cellcolor{col2}${\scriptstyle \delta < 1.25^{3}}$\\
  \hline

\arrayrulecolor{black}
\hline
\hline

\cite{godard2019digging} & \cellcolor{ssectioncolor} Monodepth2 Resnet18 & \cmark & S & 192 $\times$ 640 &
  0.079  &   0.512  &   3.721  &   0.131  &   0.924  &   0.982  &   0.994  \\

& \cellcolor{ssectioncolor} \textbf{WaveletMonodepth Resnet18} & \cmark & S & 192 $\times$ 640 &   0.084  &   0.523  &   3.807  &   0.137  &   0.914  &   0.980  &   0.994\\

& \cellcolor{ssectioncolor} \textbf{WaveletMonodepth Resnet50} & \cmark & S & 192 $\times$ 640 &
0.081  &   0.477  &   3.658  &   0.133  &   0.920  &   0.981  &   0.994  \\  

\hline
\cite{watson-2019-depth-hints}& \cellcolor{ssectioncolor} Depth Hints & \cmark & S$_{SGM}$ & 192 $\times$ 640 &
0.085 & 0.487 & 3.670 & 0.131 & 0.917 & 0.983 & 0.996 \\

& \cellcolor{ssectioncolor} \textbf{WaveletMonodepth Resnet18} & \cmark & S$_{SGM}$ & 192 $\times$ 640 &
0.083 & 0.476 & 3.635 & 0.129 & 0.920 & 0.983 & 0.995\\

& \cellcolor{ssectioncolor} Depth Hints Resnet50 & \cmark & S$_{SGM}$ & 192 $\times$ 640 &
  0.081  &   0.432  &   3.510  &   0.124  &   0.924  &   0.985  &   0.996  \\

& \cellcolor{ssectioncolor} \textbf{WaveletMonodepth Resnet50} & \cmark & S$_{SGM}$ & 192 $\times$ 640 &
0.081  &   0.449  &   3.509  &   0.125  &   0.923  &   0.986  &   0.996   \\  
\hline

\cite{godard2019digging} & \cellcolor{mssectioncolor} Monodepth2 Resnet18 & \cmark & MS  & 192 $\times$ 640 &   0.084  &   0.494  &   3.739  &   0.132  &   0.918  &   0.983  &   0.995  \\

& \cellcolor{mssectioncolor} \textbf{WaveletMonodepth Resnet18} & \cmark  & MS & 192 $\times$ 640 & 0.085  &   0.497  &   3.804  &   0.134  &   0.912  &   0.982  &   0.995 \\

\hline

\cite{watson-2019-depth-hints} & \cellcolor{mssectioncolor} Depth Hints & \cmark & MS + $S_{SGM}$ & 192 $\times$ 640 & 
  0.087  &   0.526  &   3.776  &   0.133  &   0.915  &   0.982  &   0.995  \\
  
& \cellcolor{mssectioncolor} \textbf{WaveletMonodepth Resnet18} & \cmark & MS + $S_{SGM}$ & 192 $\times$ 640 & 0.086  &   0.497  &   3.699  &   0.131  &   0.914  &   0.983  &   0.996 \\

\hline
\cite{godard2019digging} & \cellcolor{hrsectioncolor} Monodepth2 Resnet18& \cmark & S & 320 $\times$ 1024 &
 0.082  &   0.497  &   3.637  &   0.132  &   0.924  &   0.982  &   0.994  \\

& \cellcolor{hrsectioncolor} \textbf{WaveletMonodepth Resnet18} & \cmark & S & 320 $\times$ 1024 & 
0.080  &   0.443  &   3.544  &   0.130  &   0.919  &   0.983  &   0.995\\

& \cellcolor{hrsectioncolor} \textbf{WaveletMonodepth Resnet50} & \cmark & S & 320 $\times$ 1024 &
0.076  &   0.413  &   3.434  &   0.126  &   0.926  &   0.984  &   0.995 \\

\hline
\cite{watson-2019-depth-hints} & \cellcolor{hrsectioncolor} Depth Hints & \cmark & S$_{SGM}$ & 320 $\times$ 1024 &
0.077 & 0.404 & 3.345 & 0.119 & 0.930 &  0.988 & 0.997 \\

& \cellcolor{hrsectioncolor} \textbf{WaveletMonodepth Resnet18} & \cmark & S$_{SGM}$ & 320 $\times$ 1024 &
0.078  &   0.397  &   3.316  &   0.121  &   0.928  &   0.987  &   0.997 \\

& \cellcolor{hrsectioncolor} Depth Hints Resnet50& \cmark & S$_{SGM}$ & 320 $\times$ 1024 &
 \textbf{0.074}  &   0.363  &   3.198  &   \textbf{0.114}  &   \textbf{0.936}  &   \textbf{0.989}  &   \textbf{0.997}  \\

& \cellcolor{hrsectioncolor} \textbf{WaveletMonodepth Resnet50} & \cmark & S$_{SGM}$ & 320 $\times$ 1024 &
\textbf{0.074}  &   \textbf{0.357}  &   \textbf{3.170}  &   \textbf{0.114}  &   \textbf{0.936}  &   \textbf{0.989}  &   \textbf{0.997}\\

\arrayrulecolor{black}
\hline

\arrayrulecolor{black}\hline
  \end{tabular}
  }
    \vspace{3pt}
  \caption{\textbf{Quantitative results on the improved KITTI benchmark.} We compare our method to our baselines on the KITTI~\cite{Geiger2012CVPR} improved dataset introduced by~\cite{uhrig2017sparse}, using the Eigen split. 
  \textit{Data column} (data source used for training): 
  S is for self-supervised training on stereo images, MS is for models trained with both M (forward and backward frames) and S data and $S_{SGM}$ refers to the extra stereo ground truth which was used in ~\cite{watson-2019-depth-hints}.
  }
  \vspace{6pt}
  \label{tab:kitti_eigen_improved}
\end{table*}

\section{Exploring Other Efficiency Tracks}\label{sec:mobilenets}

Our paper mainly explores computation reduction in the decoder of a UNet-like architecture. However, this direction is orthogonal and complementary with all other complexity reduction lines of research.

Our approach is for example complementary with the FastDepth approach, which consists in reducing the overall complexity of a depth estimation network by compressing it in many dimensions such as (1) the encoder, (2) the decoder (3) the input resolution. They argue that the deep network introduced by Laina \textit{et al.}~\cite{Laina2016DeeperDP} suffers from high complexity, while it could largely be reduced. Here we present a set of experiments we conducted to explore these different aspects of complexity reduction.

\begin{table*}[!ht]
  \centering
  \centering
  \footnotesize
  \resizebox{1.0\textwidth}{!}{
  \begin{tabular}{|c l|c|c|c||c|c|c|c|c|c|c|}
  \hline
  Cit. & Method & PP & Data & H $\times$ W & \cellcolor{col1}{\scriptsize Abs Rel} & \cellcolor{col1}{\scriptsize Sq Rel} & \cellcolor{col1}{\scriptsize RMSE}  & \cellcolor{col1}{\scriptsize RMSE log} & \cellcolor{col2}${\scriptstyle \delta < 1.25}$ & \cellcolor{col2}${\scriptstyle \delta < 1.25^{2}}$ & \cellcolor{col2}${\scriptstyle \delta < 1.25^{3}}$\\
  \hline

\arrayrulecolor{black}
\hline
\hline

\hline
\cite{watson-2019-depth-hints}& \cellcolor{ssectioncolor} Depth Hints & \cmark & S$_{SGM}$ & 192 $\times$ 640 &
0.106 & 0.780 & 4.695 & 0.193 & 0.875 & 0.958 & 0.980 \\

& \cellcolor{ssectioncolor} \textbf{WaveletMonodepth MobileNetv2} & \cmark & S$_{SGM}$ & 192 $\times$ 640  &
0.109  &   0.851  &   4.754  &   0.194  &   0.870  &   0.957  &   0.980\\

& \cellcolor{ssectioncolor} \textbf{WaveletMonodepth Resnet18} & \cmark & S$_{SGM}$ & 192 $\times$ 640 &
0.107  &   0.829  &   4.693  &   0.193  &   0.873  &   0.957  &   0.980  \\

& \cellcolor{ssectioncolor} \textbf{WaveletMonodepth Resnet50} & \cmark & S$_{SGM}$ & 192 $\times$ 640 &
0.105  &   0.813  &   4.625  &   0.191  &   0.879  &   0.959  &   0.981   \\  
\hline

\cite{watson-2019-depth-hints} & \cellcolor{hrsectioncolor} Depth Hints & \cmark & S$_{SGM}$ & 320 $\times$ 1024 &
0.099 &   0.723 &  4.445 &  0.187 & 0.886 & 0.961 & 0.982 \\ 

& \cellcolor{hrsectioncolor} \textbf{WaveletMonodepth MobileNetv2} & \cmark & S$_{SGM}$ & 320 $\times$ 1024 &
0.104  &   0.772  &   4.545  &   0.188  &   0.880  &   0.960  &   0.982\\

& \cellcolor{hrsectioncolor} \textbf{WaveletMonodepth Resnet18} & \cmark & S$_{SGM}$ & 320 $\times$ 1024 &
0.102  &   0.739  &   4.452  &   0.188  &   0.883  &   0.960  &   0.981 \\

& \cellcolor{hrsectioncolor} Depth Hints Resnet50& \cmark & S$_{SGM}$ & 320 $\times$ 1024 &
\textbf{0.096} &   \textbf{0.710} & 4.393 &  0.185 & 0.890 & \textbf{0.962} & 0.981 \\  

& \cellcolor{hrsectioncolor} \textbf{WaveletMonodepth Resnet50} & \cmark & S$_{SGM}$ & 320 $\times$ 1024 &
0.097  &   0.718  &   \textbf{4.387}  &   \textbf{0.184}  &   \textbf{0.891}  &   \textbf{0.962}  &   \textbf{0.982}\\

\arrayrulecolor{black}
\hline

\arrayrulecolor{black}\hline
  \end{tabular}}
    \vspace{3pt}
  \caption{\textbf{Quantitative results on the KITTI dataset using MobileNetv2 encoder.} We evaluate results of our method using a lighter encoder on  KITTI~\cite{Geiger2012CVPR}, using the Eigen split. 
  \textit{Data column} (data source used for training): 
  S is for self-supervised training on stereo images, MS is for models trained with both M (forward and backward frames) and S data and $S_{SGM}$ refers to the extra stereo ground truth which was used in ~\cite{watson-2019-depth-hints}.
  }
  \vspace{-6pt}
  \label{tab:kitti_eigen_mobilenet}
\end{table*}

\begin{table*}[!ht]
  \centering
  \resizebox{1\textwidth}{!}{
  \begin{tabular}{|l|c|c|c||c|c|c|c|c|c|c|c|c|}
  \hline
  Method & Encoder & Depthwise &H $\times$ W & \cellcolor{col1}{\scriptsize Abs Rel} &
  \cellcolor{col1}{\scriptsize RMSE}  & \cellcolor{col1}{\scriptsize $log_{10}$} & \cellcolor{col2}$\delta_1$ & \cellcolor{col2}$\delta_2$ & \cellcolor{col2}$\delta_3$ & \cellcolor{col1}$\epsilon_{acc}$ & \cellcolor{col1}$\epsilon_{comp}$\\
  \hline

\hline

Dense baseline  & DenseNet161 & - & 480 $\times$ 640 &
0.1277  &    \textbf{0.5479}  &    \textbf{0.0539}  &    0.8430  &    \textbf{0.9681}  &    \textbf{0.9917}  &    \textbf{1.7170}  &    \textbf{7.0638}\\ 
\textbf{Ours} & DenseNet161 & - & 480 $\times$ 640 &
\textbf{0.1258}  &    0.5515  &    0.0542  &    \textbf{0.8451}  &    \textbf{0.9681}  &    \textbf{0.9917}  &    1.8070  &    7.1073 \\

\textbf{Ours} & DenseNet161 & \cmark & 480 $\times$ 640 & 
0.1275  &    0.5771  &    0.0557  &    0.8364  &    0.9635  &    0.9897  &    2.0133  &    7.1903\\

\hline
Dense baseline & MobileNetv2 & - & 480 $\times$ 640 & 0.1772  &    0.6638  &    0.0731  &    0.7419  &    0.9341  &    0.9835  &    1.8911  &    7.7960\\ 
\textbf{Ours} & MobileNetv2 & - & 480 $\times$ 640 & 
0.1727  &    0.6776  &    0.0732  &    0.7380  &    0.9362  &    0.9844  &    1.9732  &    7.9004\\
\textbf{Ours} & MobileNetv2 & \cmark & 480 $\times$ 640 & 0.1734  &    0.6700  &    0.0731  &    0.7391  &    0.9347  &    0.9844  &    2.3036  &    8.0538 \\

\hline

\arrayrulecolor{black}\hline
  \end{tabular}
  }
    \vspace{0pt}
  \caption{\textbf{Quantitative results on NYUv2~\cite{Nyuv2} using depth-wise convolutions and light-weight encoder} We show that our method is compatible with other efficiency seeking approaches such as depth-wise separable convolutions and lower complexity encoders.
  \vspace{-6pt}
  }
  \label{tab:nyu_mobilenet}
\end{table*}

\subsection{Experiment with a light-weight MobileNetv2 encoder}

First, we replace the costly ResNet~\cite{He2016ResNet} or DenseNet~\cite{huang2017densenet,huang2019densenet} backbone encoders with the efficient MobileNetv2~\cite{sandler2018mobilenetv2}.
Indeed, in contrast with FastDepth, in the main paper, we report results using large encoder models~(Resnet18/50 or Densenet161). Although this helps achieving better scores, we show in Table~\ref{tab:kitti_eigen_mobilenet} and Table~\ref{tab:nyu_mobilenet} that we can reach close to state-of-the-art results even with a small encoder such as MobileNetv2.

\subsection{Separable convolutions}
Secondly, FastDepth also shows that separable convolutions in their "NNConv" decoder provides the best score-efficiency trade-off. Since this approach is orthogonal to our sparsification method, it therefore complements our method and can be used to improve efficiency. Interestingly, we show in Table~\ref{tab:nyu_mobilenet} that replacing sparse convolutions with sparse-depthwise separable convolutions works on par with standard convolutions. This can be explained by the fact that {\IDWT} is also a separable operation, and therefore efficiently combines with depthwise separable convolutions. 

\begin{table*}[!ht]
  \centering
  \begin{tabular}{|l|c|c|c||c|c|c|c|c|c|c|}
  \hline
  Method & Encoder & Depthwise &H $\times$ W & \cellcolor{col1}{\scriptsize Abs Rel} &
  \cellcolor{col1}{\scriptsize RMSE}  & \cellcolor{col1}{\scriptsize $log_{10}$} & \cellcolor{col2}$\delta_1$ & \cellcolor{col2}$\delta_2$ & \cellcolor{col2}$\delta_3$\\
  \hline

\hline
Dense baseline  & DenseNet161 & - & 224 $\times$ 224 & 0.1278  &    0.5715  &    0.0557  &    0.8368  &    0.9620  &    0.9901\\ 
\textbf{Ours} & DenseNet161 & - & 224 $\times$ 224 & 0.1279  &    0.5651  &    0.0549  &    0.8399  &    0.9652  &    0.9899\\
\textbf{Ours} & DenseNet161 & \cmark & 224 $\times$ 224 &
0.1304  &    0.5775  &    0.0564  &    0.8329  &    0.9613  &    0.9892\\

\hline

Dense baseline & MobileNetv2 & - & 224 $\times$ 224 & 0.1505  &    0.6221  &    0.0632  &    0.7984  &    0.9526  &    0.9878\\ 
\textbf{Ours} & MobileNetv2 & - & 224 $\times$ 224 & 0.1530  &    0.6409  &    0.0655  &    0.7844  &    0.9500  &    0.9864\\ 
\textbf{Ours} & MobileNetv2 & \cmark & 224 $\times$ 224 &
0.1491  &    0.6463  &    0.0646  &    0.7880  &    0.9506  &    0.9871\\

\arrayrulecolor{black}\hline
  \end{tabular}
    \vspace{0pt}
  \caption{\textbf{Quantitative results on NYUv2~\cite{Nyuv2} using depth-wise convolutions and light-weight encoder} We show that our method is compatible with other efficiency seeking approaches such as depth-wise separable convolutions and lower complexity encoders.
  \vspace{-6pt}
  }
  \label{tab:nyu_224}
\end{table*}

\subsection{Channel pruning}
\begin{figure}[!ht]
	\centering
	\resizebox{0.5\textwidth}{!}{
		\begin{subfigure}[b]{0.39\linewidth}
			\centering
			\includegraphics[width=0.59\linewidth]{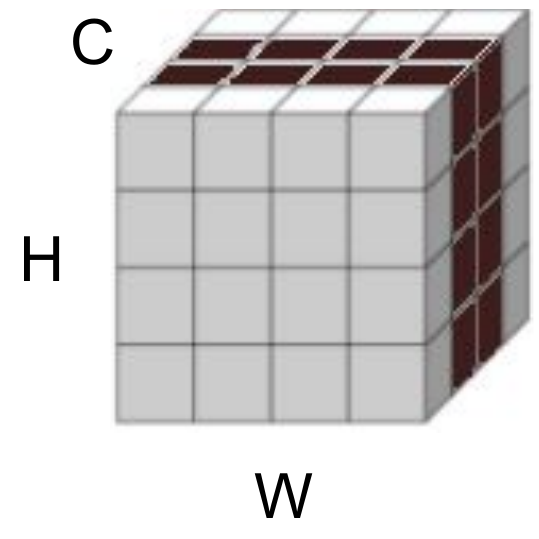}
			\caption{}
			\label{fig:dense_outconv}
		\end{subfigure} 
		\begin{subfigure}[b]{0.39\linewidth}
			\centering
			\includegraphics[width=0.59\linewidth]{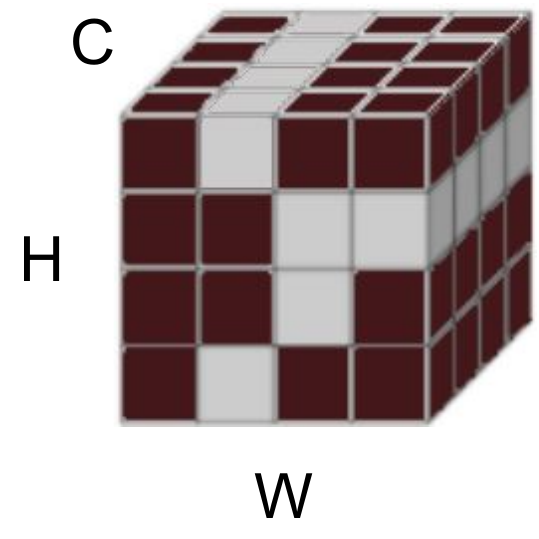}
			\caption{}
			\label{fig:sparse_outconv}
		\end{subfigure}
	}
	\caption{Channel pruning (a) vs our sparse computation (b). Our sparse computation enabled by wavelets is complimentary with the channel pruning strategy to reduce the amount of computation, as both computation reduction methods operate in orthogonal dimensions.}
	\label{fig:dense_vs_sparse_outconv}
\end{figure}

\begin{figure*}[!ht]
	\centering
	\setlength\lineskip{0pt}
	\setlength\tabcolsep{0pt} 
	\def\arraystretch{0}
	\begin{center}
		\begin{tabular}{>{\centering\arraybackslash}m{0.99\textwidth}
			}
			
			\begin{tabular}{>{\centering\arraybackslash}m{0.165\textwidth}
					>{\centering\arraybackslash}m{0.165\textwidth}
					>{\centering\arraybackslash}m{0.165\textwidth}
					>{\centering\arraybackslash}m{0.165\textwidth}
					>{\centering\arraybackslash}m{0.165\textwidth}
					>{\centering\arraybackslash}m{0.165\textwidth}}
				RGB / Mask$[\mathbf{J}]$ & Depth$_\mathbf{J}$ & \lhi{\textbf{J}} & \hli{\textbf{J}} & \hhi{\textbf{J}} & Error$[\mathbf{J}]$\\
			\end{tabular}\\

			\begin{subfigure}[b]{0.99\textwidth}	
				\centering
				\includegraphics[width=\textwidth]{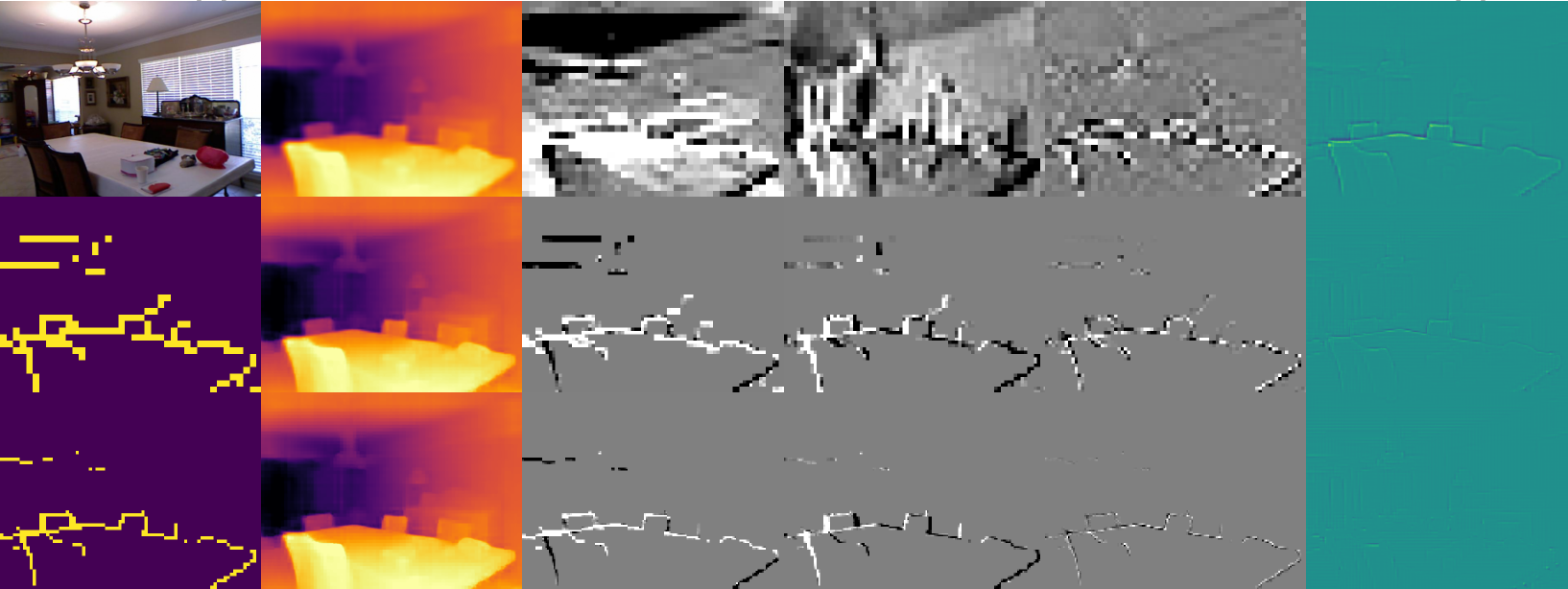}
			\end{subfigure}\\
			
			\vspace{3pt}
			\vspace{3pt}\\

			\begin{subfigure}[b]{0.99\textwidth}	
				\centering
				\includegraphics[width=\textwidth]{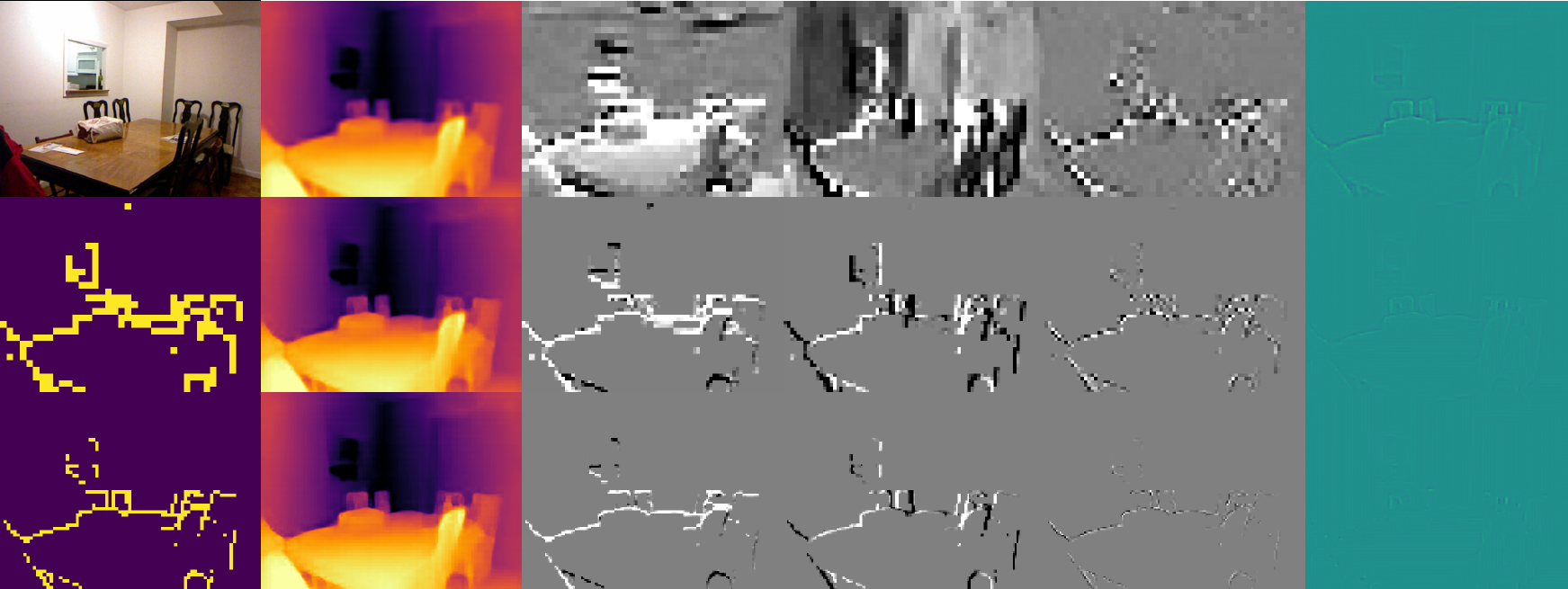}
			\end{subfigure}\\
			
			\vspace{3pt}
			\vspace{3pt}\\

			\begin{subfigure}[b]{0.99\textwidth}	
				\centering
				\includegraphics[width=\textwidth]{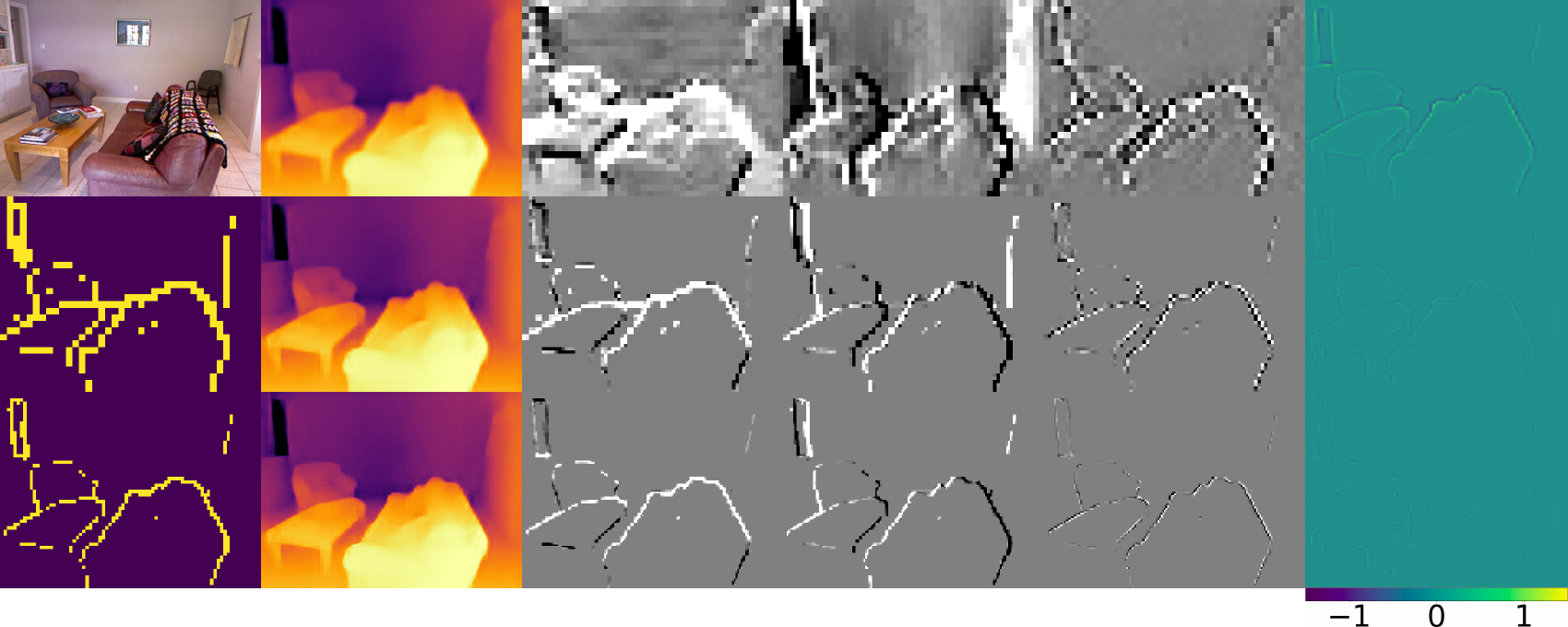}
			\end{subfigure}
		\end{tabular}
	\end{center}
	\vspace{-10pt}
	\caption{\textbf{Qualitative results of predicted wavelets coefficients of depth maps on the NYU dataset (1/3)}. Our predicted wavelets have two desirable properties: they are sparse, allowing for efficient computation, and they are accurately located around depth edges without needing to supervise them. Results are obtained with $\eta=0.04$. For each block of results, each row shows coefficients and depth maps obtained at scale \textbf{J} in the decoder from lowest to highest scale (decreasing \textbf{J}), as well as the (signed-)error between Depth$_\mathbf{J}$ and the Depth map obtained with dense wavelet coefficients at all scales. Error is displayed within range $[-1.5m, 1.5m]$.}
	\label{fig:qualitative_wavelets_nyu1}
\end{figure*}

\begin{figure*}[!ht]
	\centering
	\setlength\lineskip{0pt}
	\setlength\tabcolsep{0pt} 
	\def\arraystretch{0}
	\begin{center}
		\begin{tabular}{>{\centering\arraybackslash}m{0.99\textwidth}
			}
			
			\begin{tabular}{>{\centering\arraybackslash}m{0.165\textwidth}
							>{\centering\arraybackslash}m{0.165\textwidth}
							>{\centering\arraybackslash}m{0.165\textwidth}
							>{\centering\arraybackslash}m{0.165\textwidth}
							>{\centering\arraybackslash}m{0.165\textwidth}
							>{\centering\arraybackslash}m{0.165\textwidth}}
					RGB / Mask$[\mathbf{J}]$ & Depth$_\mathbf{J}$ & \lhi{\textbf{J}} & \hli{\textbf{J}} & \hhi{\textbf{J}} & Error$[\mathbf{J}]$\\
			\end{tabular}\\

			\begin{subfigure}[b]{0.99\textwidth}	
				\centering
				\includegraphics[width=\textwidth]{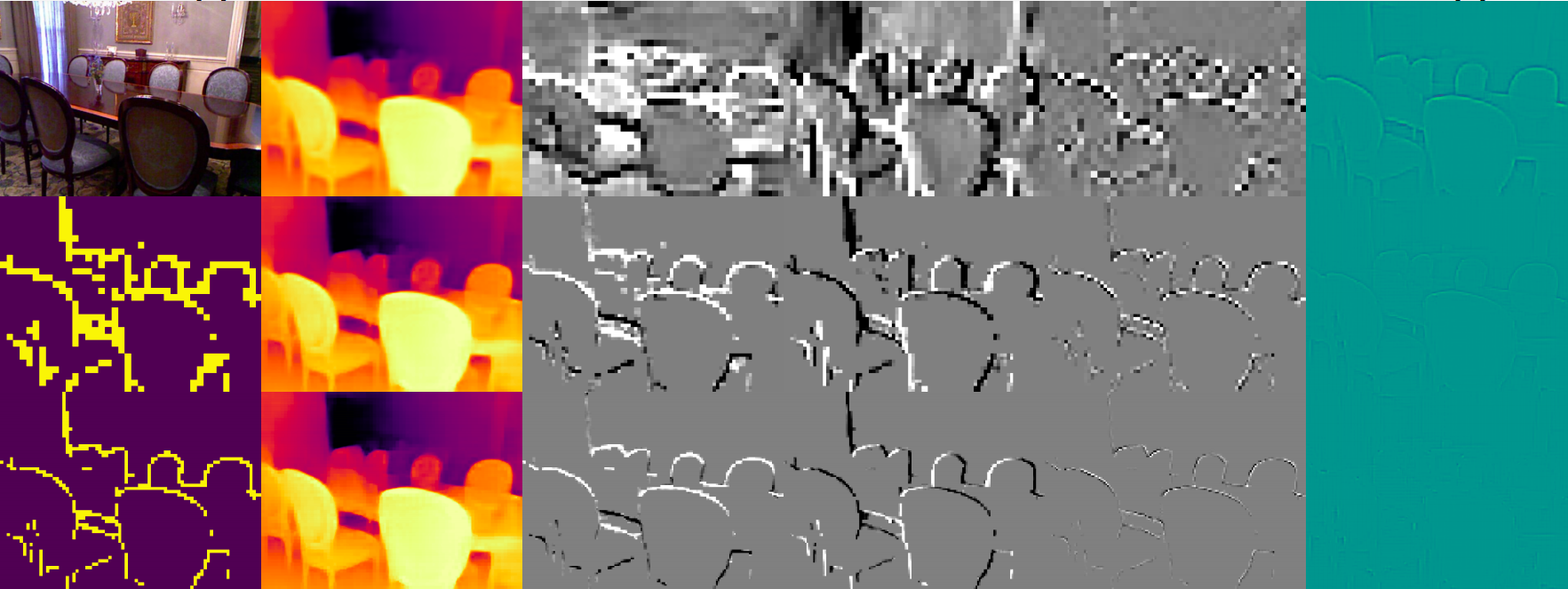}
			\end{subfigure}\\
			
			\vspace{3pt}
			\vspace{3pt}\\

			\begin{subfigure}[b]{0.99\textwidth}	
				\centering
				\includegraphics[width=\textwidth]{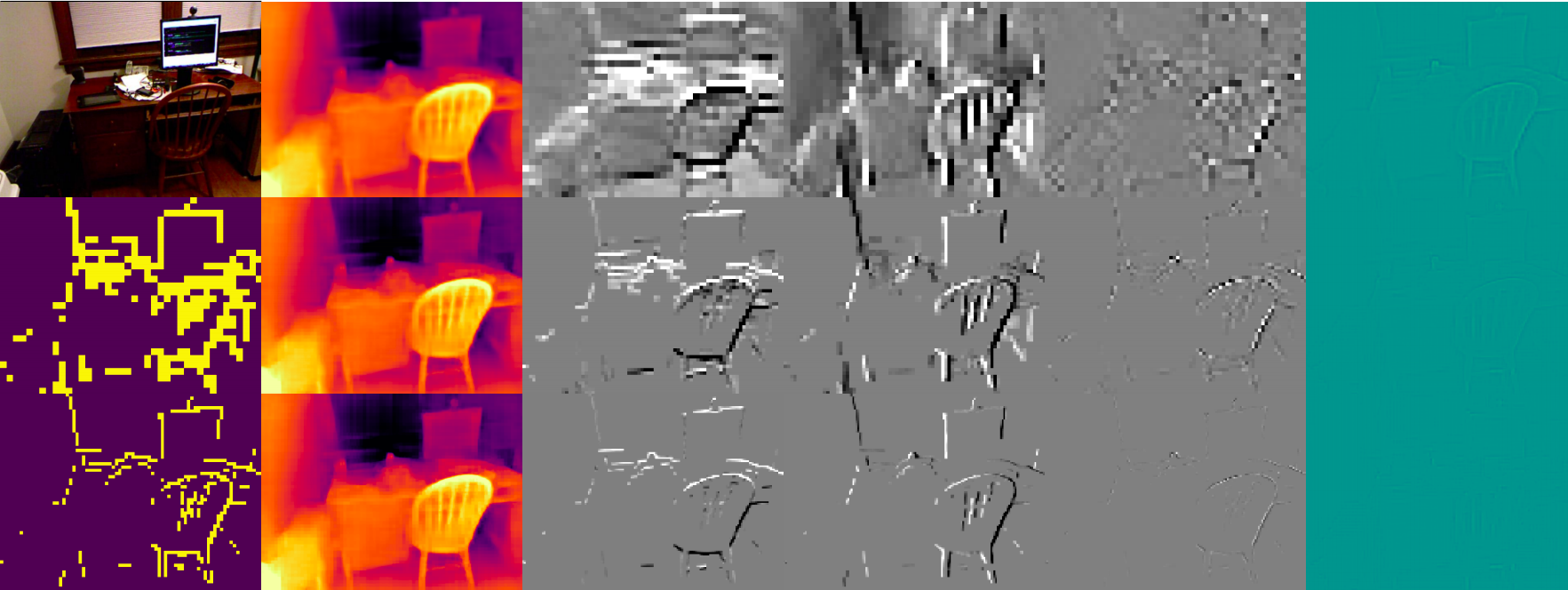}
			\end{subfigure}\\
			
			\vspace{3pt}
			\vspace{3pt}\\

			\begin{subfigure}[b]{0.99\textwidth}	
				\centering
				\includegraphics[width=\textwidth]{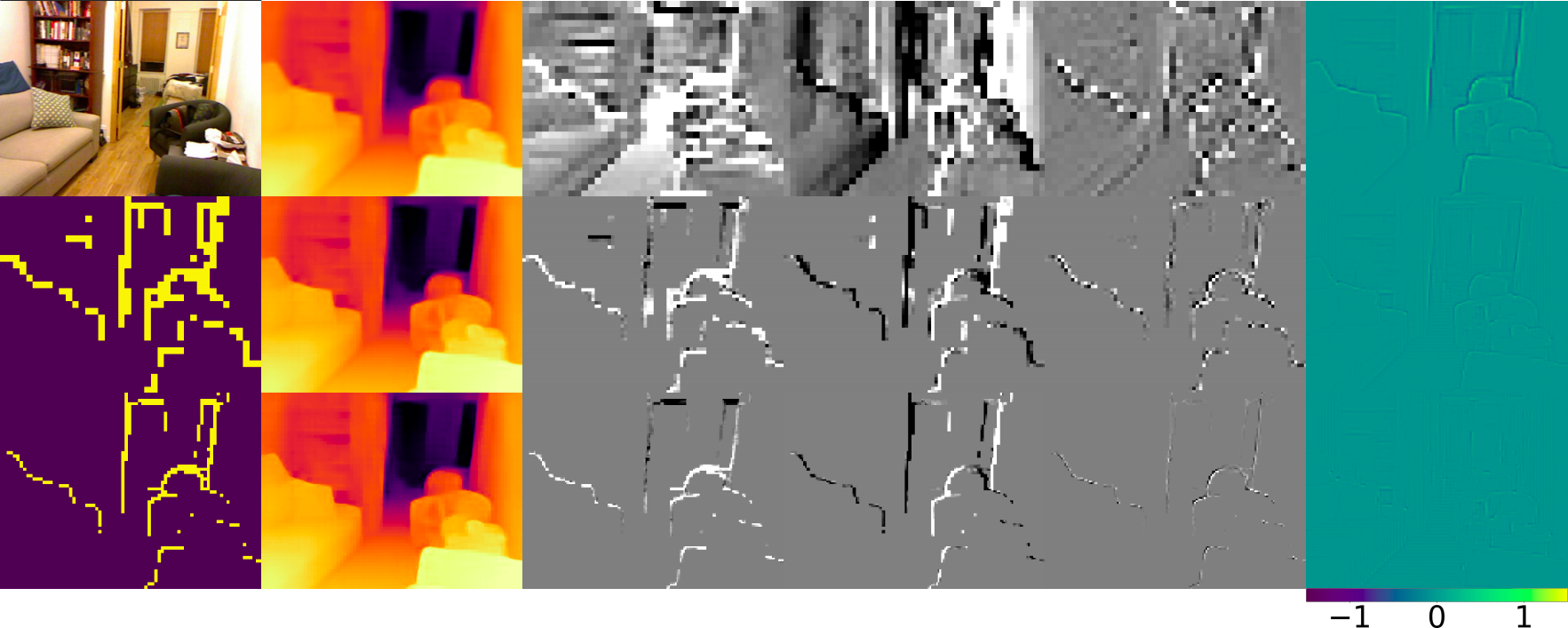}
			\end{subfigure}
		\end{tabular}
	\end{center}
	\vspace{-10pt}
	\caption{\textbf{Qualitative results of predicted wavelets coefficients of depth maps on the NYU dataset (2/3)}. Our predicted wavelets have two desirable properties: they are sparse, allowing for efficient computation, and they are accurately located around depth edges without needing to supervise them. Results are obtained with $\eta=0.04$. For each block of results, each row shows coefficients and depth maps obtained at scale \textbf{J} in the decoder from lowest to highest scale (decreasing \textbf{J}), as well as the (signed-)error between Depth$_\mathbf{J}$ and the Depth map obtained with dense wavelet coefficients at all scales. Error is displayed within range $[-1.5m, 1.5m]$.}
	\label{fig:qualitative_wavelets_nyu2}
\end{figure*}

\begin{figure*}[!ht]
	\centering
	\setlength\lineskip{0pt}
	\setlength\tabcolsep{0pt} 
	\def\arraystretch{0}
	\begin{center}
		\begin{tabular}{>{\centering\arraybackslash}m{0.99\textwidth}
			}
			
			\begin{tabular}{>{\centering\arraybackslash}m{0.165\textwidth}
					>{\centering\arraybackslash}m{0.165\textwidth}
					>{\centering\arraybackslash}m{0.165\textwidth}
					>{\centering\arraybackslash}m{0.165\textwidth}
					>{\centering\arraybackslash}m{0.165\textwidth}
					>{\centering\arraybackslash}m{0.165\textwidth}}
				RGB / Mask$[\mathbf{J}]$ & Depth$_\mathbf{J}$ & \lhi{\textbf{J}} & \hli{\textbf{J}} & \hhi{\textbf{J}} & Error$[\mathbf{J}]$\\
			\end{tabular}\\

			\begin{subfigure}[b]{0.99\textwidth}	
				\centering
				\includegraphics[width=\textwidth]{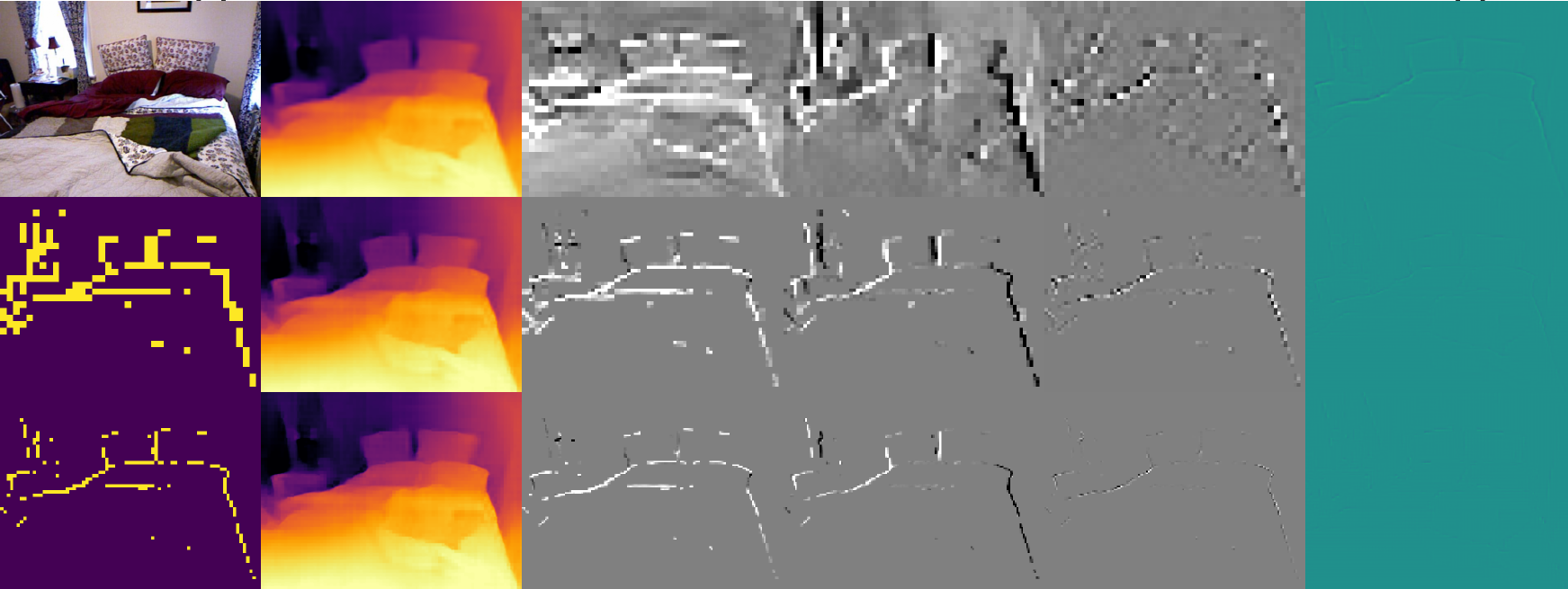}
			\end{subfigure}\\
			
			\vspace{3pt}
			\vspace{3pt}\\

			\begin{subfigure}[b]{0.99\textwidth}	
				\centering
				\includegraphics[width=\textwidth]{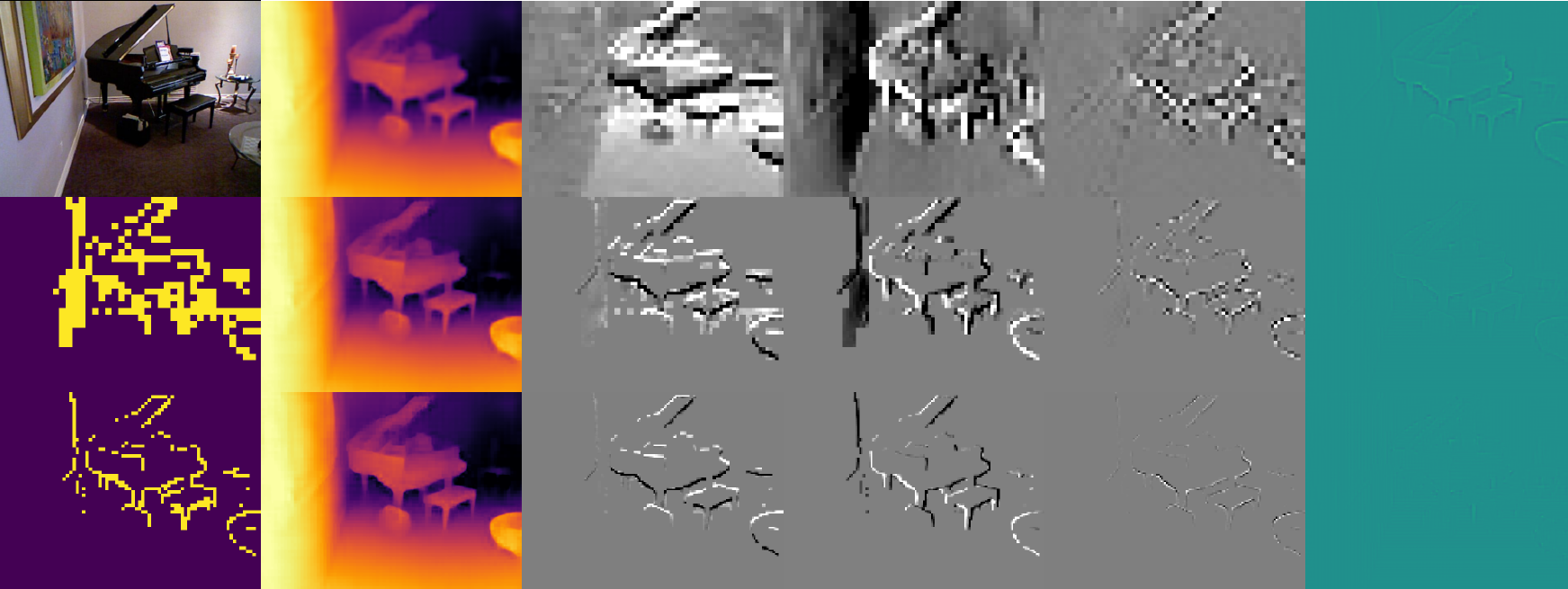}
			\end{subfigure}\\
			
			\vspace{3pt}
			\vspace{3pt}\\

			\begin{subfigure}[b]{0.99\textwidth}	
				\centering
				\includegraphics[width=\textwidth]{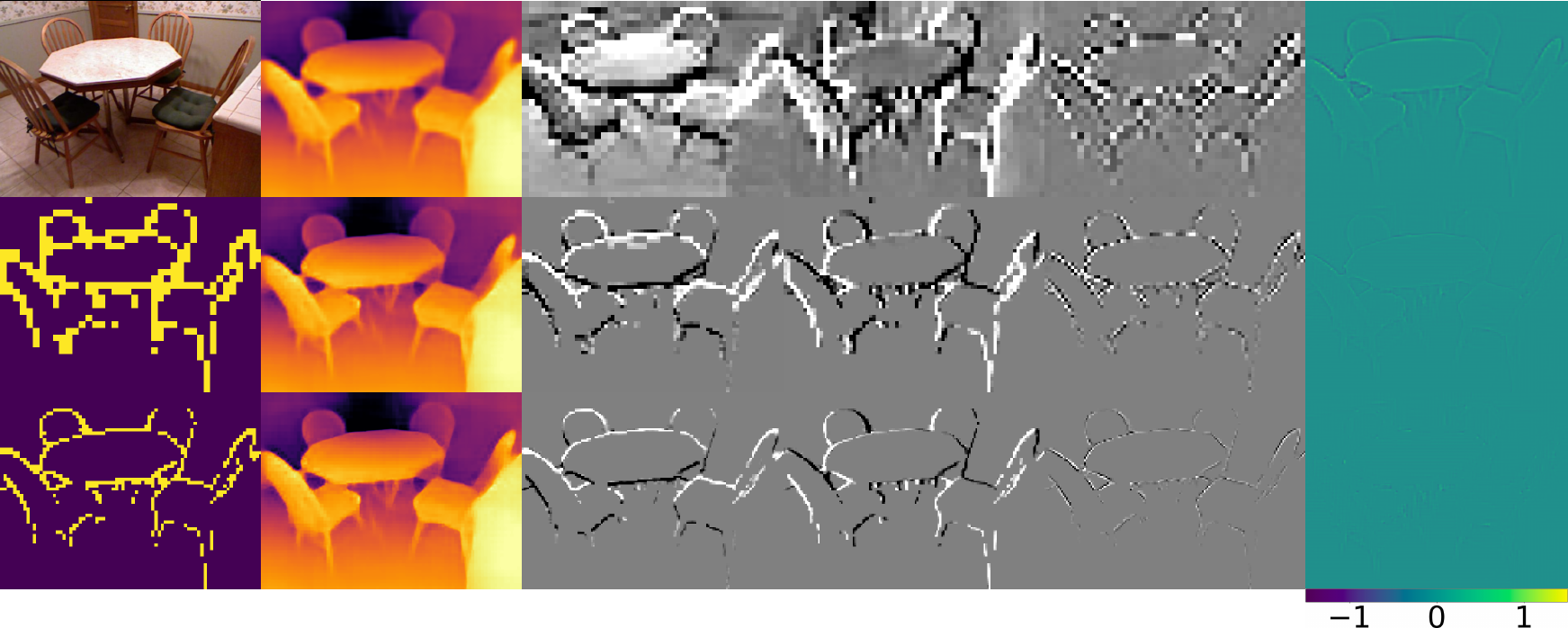}
			\end{subfigure}
		\end{tabular}
	\end{center}
	\vspace{-10pt}
	\caption{\textbf{Qualitative results of predicted wavelets coefficients of depth maps on the NYU dataset (3/3)}. Our predicted wavelets have two desirable properties: they are sparse, allowing for efficient computation, and they are accurately located around depth edges without needing to supervise them. Results are obtained with $\eta=0.04$. For each block of results, each row shows coefficients and depth maps obtained at scale \textbf{J} in the decoder from lowest to highest scale (decreasing \textbf{J}), as well as the (signed-)error between Depth$_\mathbf{J}$ and the Depth map obtained with dense wavelet coefficients at all scales. Error is displayed within range $[-1.5m, 1.5m]$.}
	\label{fig:qualitative_wavelets_nyu3}
\end{figure*}

\begin{figure*}[!ht]
	\centering
	\setlength\lineskip{0pt}
	\setlength\tabcolsep{0pt} 
	\def\arraystretch{0}
	\begin{center}
		\begin{tabular}{>{\centering\arraybackslash}m{0.99\textwidth}
			}
			
			\begin{tabular}{>{\centering\arraybackslash}m{0.165\textwidth}
					>{\centering\arraybackslash}m{0.165\textwidth}
					>{\centering\arraybackslash}m{0.165\textwidth}
					>{\centering\arraybackslash}m{0.165\textwidth}
					>{\centering\arraybackslash}m{0.165\textwidth}
					>{\centering\arraybackslash}m{0.165\textwidth}}
				RGB / Mask$[\mathbf{J}]$ & Depth$_\mathbf{J}$ & \lhi{\textbf{J}} & \hli{\textbf{J}} & \hhi{\textbf{J}} & Error$[\mathbf{J}]$\\
			\end{tabular}\\

			\begin{subfigure}[b]{0.99\textwidth}	
				\centering
				\includegraphics[width=\textwidth]{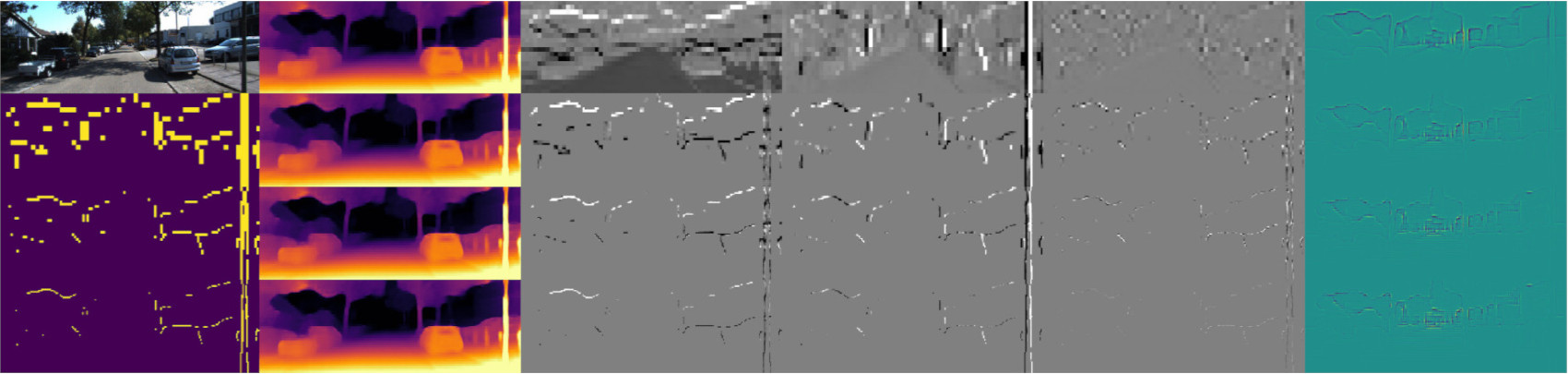}
			\end{subfigure}\\
			
			\vspace{3pt}
			\vspace{3pt}\\

			\begin{subfigure}[b]{0.99\textwidth}	
				\centering
				\includegraphics[width=\textwidth]{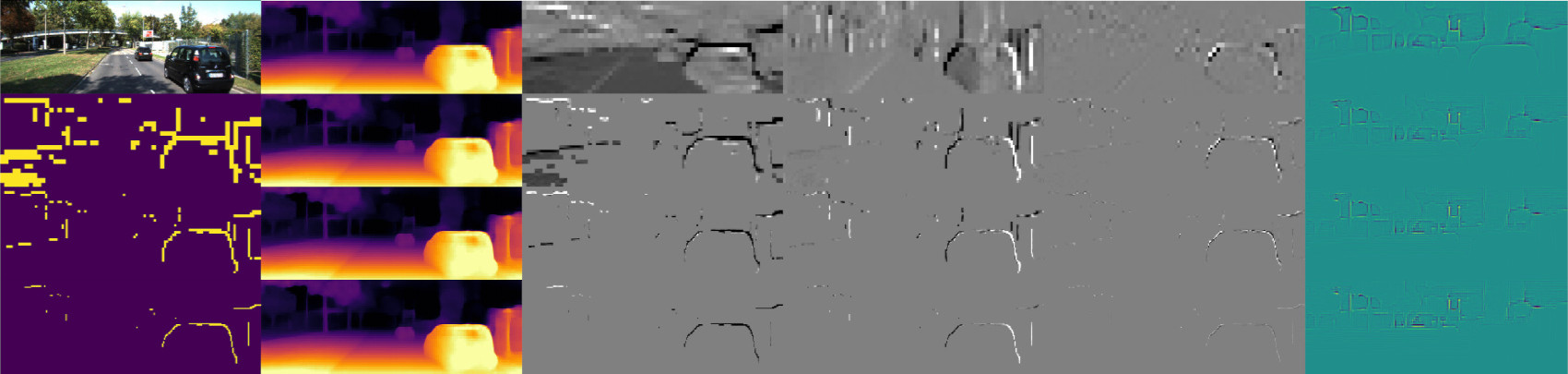}
			\end{subfigure}\\
			
			\vspace{3pt}
			\vspace{3pt}\\

			\begin{subfigure}[b]{0.99\textwidth}	
				\centering
				\includegraphics[width=\textwidth]{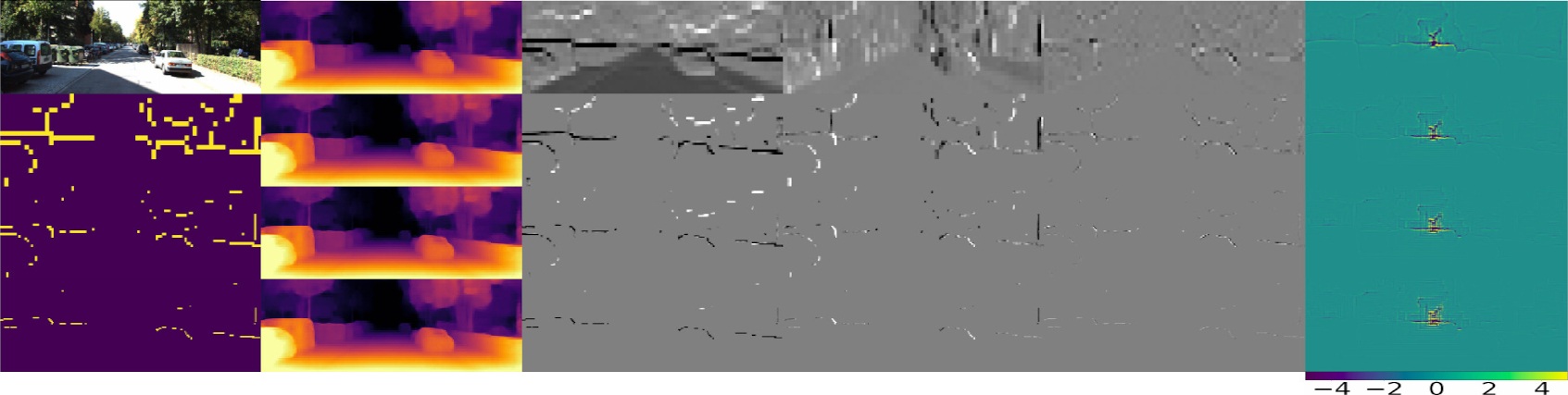}
			\end{subfigure}
		\end{tabular}
	\end{center}
	\vspace{-10pt}
	\caption{\textbf{Qualitative results of predicted wavelets coefficients of depth maps on the KITTI dataset (1/3)}. Our predicted wavelets have two desirable properties: they are sparse, allowing for efficient computation, and they are accurately located around depth edges without needing to supervise them. Results are obtained with $\eta=0.05$. For each block of results, each row shows coefficients and depth maps obtained at scale \textbf{J} in the decoder from lowest to highest scale (decreasing \textbf{J}), as well as the (signed-)error between Depth$_\mathbf{J}$ and the Depth map obtained with dense wavelet coefficients at all scales. Error is displayed within range $[-5m, 5m]$.}
	\label{fig:qualitative_wavelets_kitti1}
\end{figure*}

\begin{figure*}[!ht]
	\centering
	\setlength\lineskip{0pt}
	\setlength\tabcolsep{0pt} 
	\def\arraystretch{0}
	\begin{center}
		\begin{tabular}{>{\centering\arraybackslash}m{0.99\textwidth}
			}
			
			\begin{tabular}{>{\centering\arraybackslash}m{0.165\textwidth}
					>{\centering\arraybackslash}m{0.165\textwidth}
					>{\centering\arraybackslash}m{0.165\textwidth}
					>{\centering\arraybackslash}m{0.165\textwidth}
					>{\centering\arraybackslash}m{0.165\textwidth}
					>{\centering\arraybackslash}m{0.165\textwidth}}
				RGB / Mask$[\mathbf{J}]$ & Depth$_\mathbf{J}$ & \lhi{\textbf{J}} & \hli{\textbf{J}} & \hhi{\textbf{J}} & Error$[\mathbf{J}]$\\
			\end{tabular}\\

			\begin{subfigure}[b]{0.99\textwidth}	
				\centering
				\includegraphics[width=\textwidth]{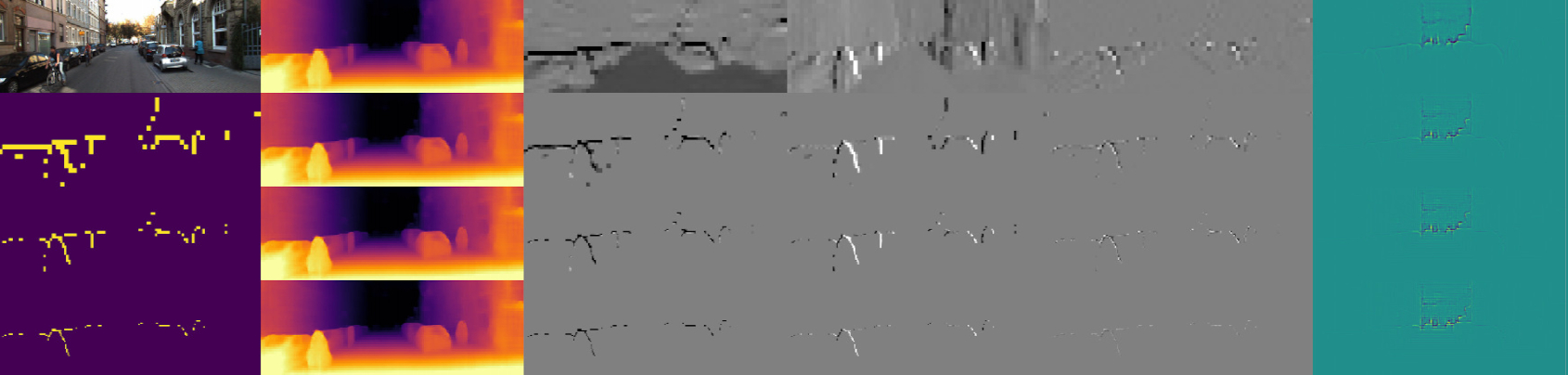}
			\end{subfigure}\\
			
			\vspace{3pt}
			\vspace{3pt}\\

			\begin{subfigure}[b]{0.99\textwidth}	
				\centering
				\includegraphics[width=\textwidth]{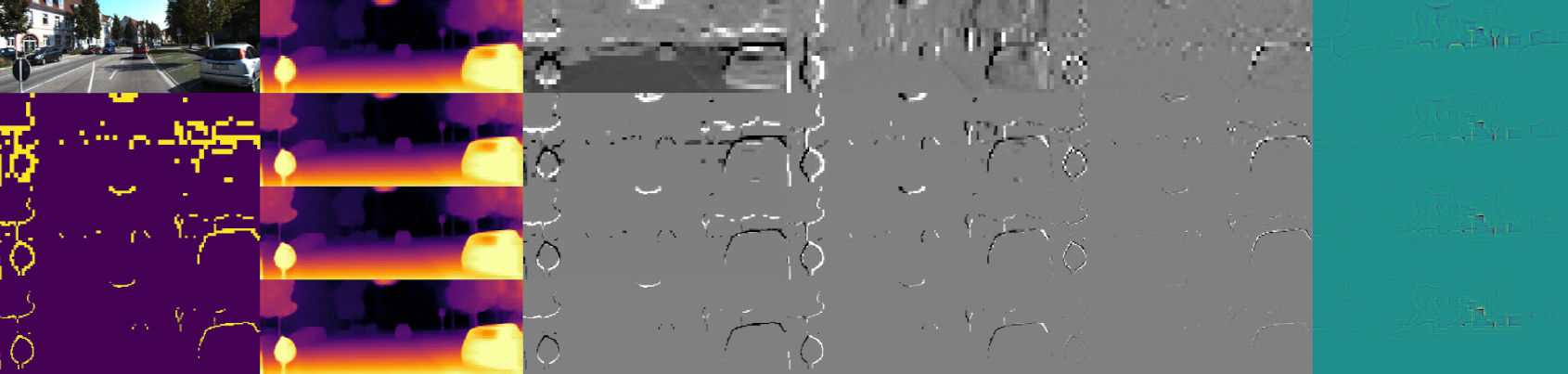}
			\end{subfigure}\\
			
			\vspace{3pt}
			\vspace{3pt}\\

			\begin{subfigure}[b]{0.99\textwidth}	
				\centering
				\includegraphics[width=\textwidth]{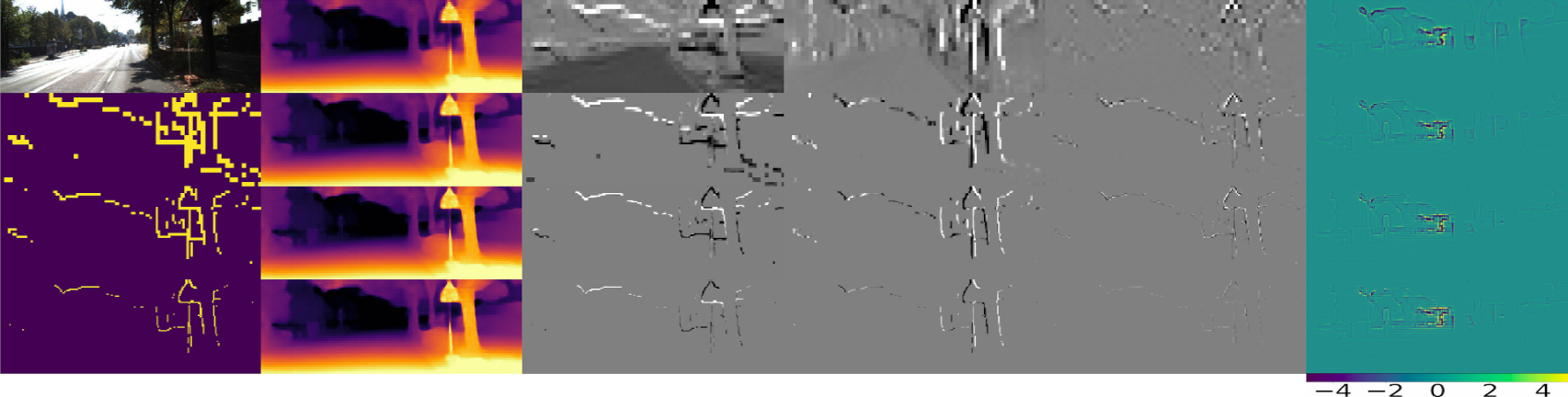}
			\end{subfigure}
		\end{tabular}
	\end{center}
	\vspace{-10pt}
	\caption{\textbf{Qualitative results of predicted wavelets coefficients of depth maps on the KITTI dataset (2/3)}. Our predicted wavelets have two desirable properties: they are sparse, allowing for efficient computation, and they are accurately located around depth edges without needing to supervise them. Results are obtained with $\eta=0.05$. For each block of results, each row shows coefficients and depth maps obtained at scale \textbf{J} in the decoder from lowest to highest scale (decreasing \textbf{J}), as well as the (signed-)error between Depth$_\mathbf{J}$ and the Depth map obtained with dense wavelet coefficients at all scales. Error is displayed within range $[-5m, 5m]$.}
	\label{fig:qualitative_wavelets_kitti2}
\end{figure*}

\begin{figure*}[!ht]
	\centering
	\setlength\lineskip{0pt}
	\setlength\tabcolsep{0pt} 
	\def\arraystretch{0}
	\begin{center}
		\begin{tabular}{>{\centering\arraybackslash}m{0.99\textwidth}
			}
			
			\begin{tabular}{>{\centering\arraybackslash}m{0.165\textwidth}
					>{\centering\arraybackslash}m{0.165\textwidth}
					>{\centering\arraybackslash}m{0.165\textwidth}
					>{\centering\arraybackslash}m{0.165\textwidth}
					>{\centering\arraybackslash}m{0.165\textwidth}
					>{\centering\arraybackslash}m{0.165\textwidth}}
				RGB / Mask$[\mathbf{J}]$ & Depth$_\mathbf{J}$ & \lhi{\textbf{J}} & \hli{\textbf{J}} & \hhi{\textbf{J}} & Error$[\mathbf{J}]$\\
			\end{tabular}\\

			\begin{subfigure}[b]{0.99\textwidth}	
				\centering
				\includegraphics[width=\textwidth]{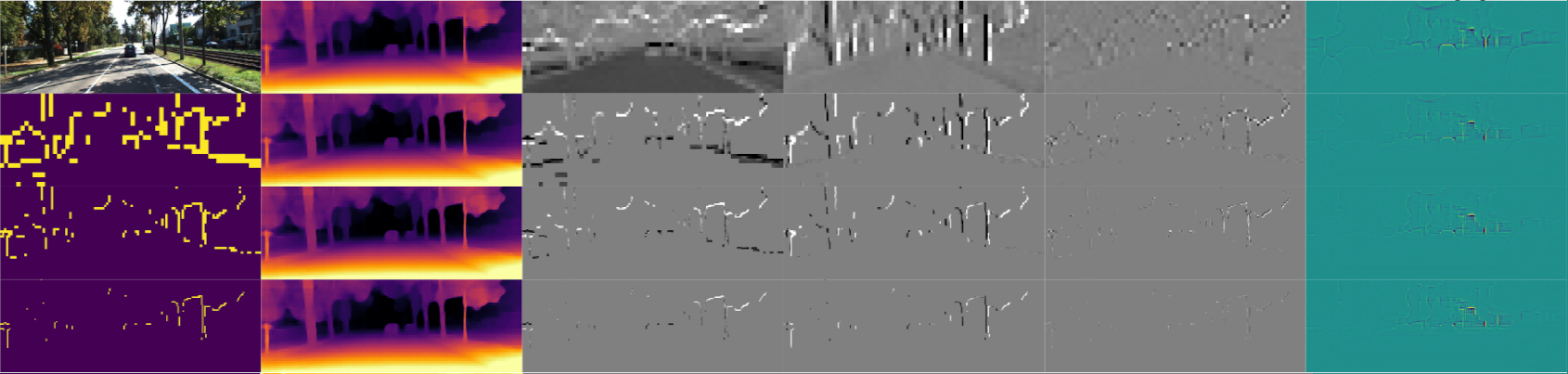}
			\end{subfigure}\\
			
			\vspace{3pt}
			\vspace{3pt}\\

			\begin{subfigure}[b]{0.99\textwidth}	
				\centering
				\includegraphics[width=\textwidth]{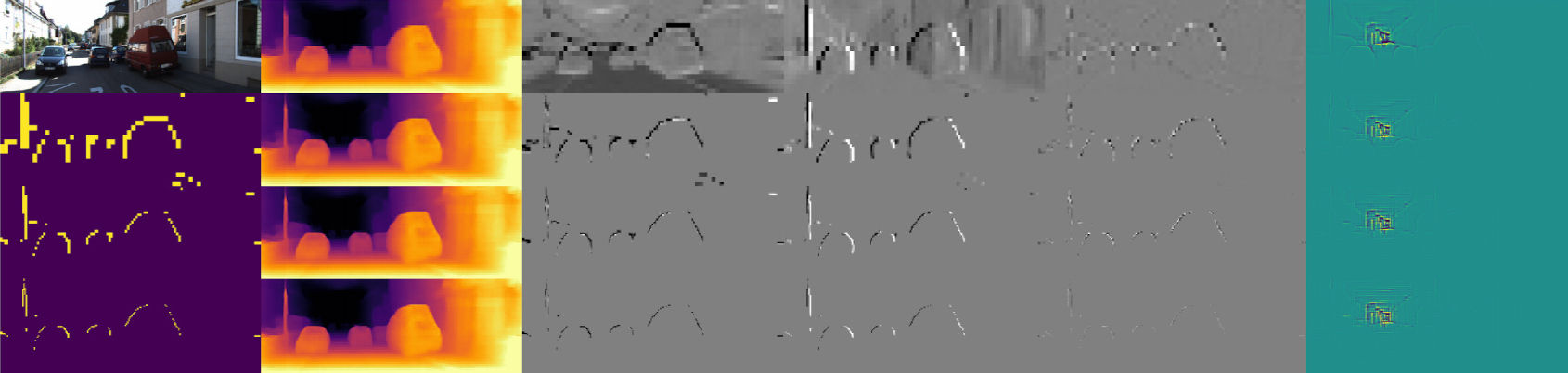}
			\end{subfigure}\\
			
			\vspace{3pt}
			\vspace{3pt}\\

			\begin{subfigure}[b]{0.99\textwidth}	
				\centering
				\includegraphics[width=\textwidth]{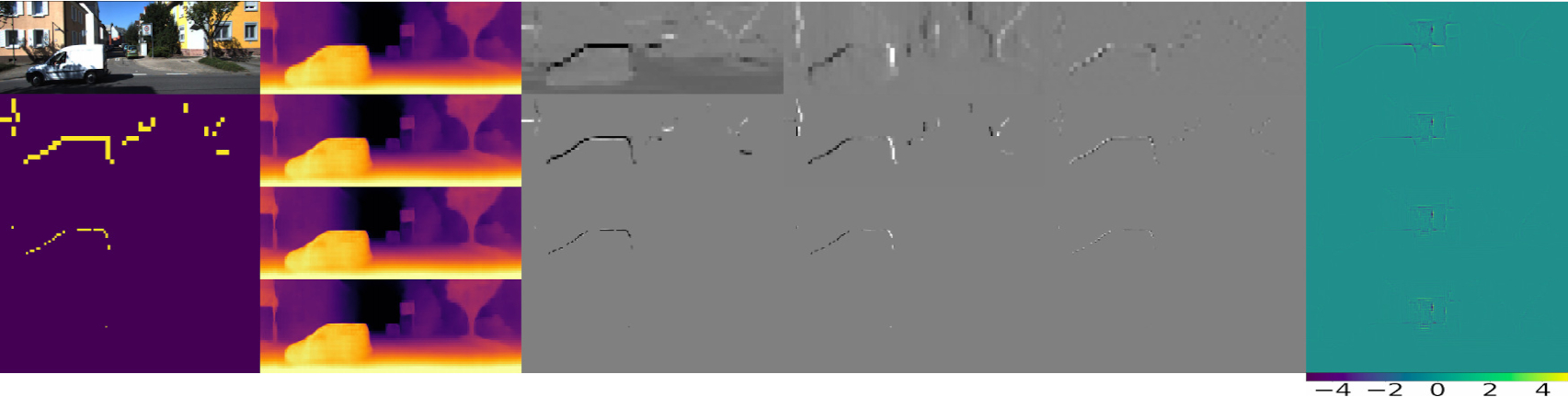}
			\end{subfigure}
		\end{tabular}
	\end{center}
	\vspace{-10pt}
	\caption{\textbf{Qualitative results of predicted wavelets coefficients of depth maps on the KITTI dataset (3/3)}. Our predicted wavelets have two desirable properties: they are sparse, allowing for efficient computation, and they are accurately located around depth edges without needing to supervise them. Results are obtained with $\eta=0.05$. For each block of results, each row shows coefficients and depth maps obtained at scale \textbf{J} in the decoder from lowest to highest scale (decreasing \textbf{J}), as well as the (signed-)error between Depth$_\mathbf{J}$ and the Depth map obtained with dense wavelet coefficients at all scales. Error is displayed within range $[-5m, 5m]$.}
	\label{fig:qualitative_wavelets_kitti3}
\end{figure*}

A popular approach to complexity and memory footprint reduction is channel pruning, which aims at removing some of the unnecessary channel in convolutional layers. Note that our wavelet enabled sparse convolutions are complementary with channel pruning, as can be seen in Figure~\ref{fig:dense_vs_sparse_outconv}. While channel pruning can, in practice, greatly reduce both complexity and memory footprint, it requires heavy hyper-parameter search that we therefore choose to leave for future work.


\subsection{Input resolution}\label{ssec:low_res}
Finally, one important factor that makes FastDepth efficient is that it is trained with $224\times224$ inputs, against our $640\times480$ input.
While our method is best designed for higher-resolution regime where sparsity of wavelets is stronger, we still show that our method achieves decent results even at low-resolution, and report our scores in Table~\ref{tab:nyu_224}.

\end{document}